\documentclass{article}

\PassOptionsToPackage{numbers, compress}{natbib}
 \usepackage[preprint]{neurips_2026}

\usepackage[utf8]{inputenc} 
\usepackage[T1]{fontenc}    
\usepackage{hyperref}       
\usepackage{url}            
\usepackage{booktabs}       
\usepackage{amsfonts}       
\usepackage{nicefrac}       
\usepackage{microtype}      
\usepackage[table]{xcolor}         

\usepackage{amsmath,amsfonts,bm}

\def\eqref#1{equation~\ref{#1}}

\def\1{\bm{1}}

\DeclareMathAlphabet{\mathsfit}{\encodingdefault}{\sfdefault}{m}{sl}
\SetMathAlphabet{\mathsfit}{bold}{\encodingdefault}{\sfdefault}{bx}{n}

\usepackage{makecell, graphicx, romannum, amssymb, amsmath, ccaption, setspace, algorithm, algpseudocode, enumitem, pifont, etoolbox, dsfont, fontawesome5, longtable, array, arydshln, bbding, multicol, multirow, caption, wrapfig, fvextra, titletoc, hhline}

\newfixedcaption{\outfigcaption}{figure}
\newfixedcaption{\outtabcaption}{table}

\usepackage{tcolorbox,titletoc}
\newcommand{\authcount}[1]{}
\usepackage{xstring,catchfile}

\title{\textit{GRASP}: Learning to Ground Social Reasoning in Multi-Person Non-Verbal Interactions}

\author{
Junho Kim$^{1}$~~~~
Xu Cao$^{1}$~~~~
Houze Yang$^{1}$~~~~
Bikram Boote$^{1}$~~~~
Ana Jojic$^{1}$\AND
Fiona Ryan$^{2}$~~~~
Bolin Lai$^{3}$~~~~
Sangmin Lee$^{4}$\thanks{Corresponding authors.}~~~~
James M. Rehg$^{1}$\footnotemark[1]
\\[4mm]
$^{1}$University of Illinois Urbana-Champaign~~~~
$^{2}$Georgia Institute of Technology\\
$^{3}$Amazon AGI~~~~~~~~
$^{4}$Korea University
}

\begin{document}

\pagenumbering{arabic}
\maketitle

\begin{abstract}
    Understanding social interactions requires reasoning over subtle non-verbal cues, yet current multimodal large language models (MLLMs) often fail to identify \textit{who interacts with whom} in multi-person videos. We introduce \textit{GRASP}, a large-scale social reasoning dataset that connects high-level social QA with fine-grained gaze and deictic gesture events. \textit{GRASP} contains 290K question--answer pairs over 46K videos totaling 749 hours, organized by a 16-category taxonomy spanning gaze, gesture, and joint gaze--gesture reasoning, together with \textit{GRASP-Bench} for evaluation. Unlike prior resources that focus on either isolated cues or high-level social QA, \textit{GRASP} builds questions from identity-consistent gaze trajectories, deictic gestures, and their joint compositions into social events. Moreover, we propose \textit{Social Grounding Reward} (\textit{SGR}), a learning signal that uses these social events to encourage models to reason about the participants involved in each interaction. Experiments show that \textit{SGR} improves performance on \textit{GRASP-Bench} while maintaining zero-shot performance on related social video QA benchmarks.
\end{abstract}

\section{Introduction}
\label{sec:intro}
In everyday social interactions, the meaning of a conversation is often shaped and influenced in subtle but important ways by non-verbal cues such as gaze and gesture~\cite{hall2019nonverbal, gagnon2021reasoning, lee2024towards}. A brief glance toward a social partner, pointing at an object of interest, or a moment of mutual eye contact can signal who is being addressed, what is being referenced, and how participants relate to one another. Interpreting such signals~\cite{lee2024modeling, cao2025socialgesture, wei2024nonverbal} is inherently challenging, as it requires reasoning about complex interactions between individuals from ambiguous visual cues. Humans naturally resolve this complexity by integrating multimodal information~\cite{frith2012mechanisms,kendrick2023turn} to infer social meaning and dynamics~\cite{fan2019understanding, li2025towards}. This capacity for social understanding underpins collaborative human behavior~\cite{tomasello1986joint} and it is key capability that can enable AI systems are to interact effectively with people, from embodied agents~\cite{wu2023tidybot} to conversational systems~\cite{shum2018eliza}.

Prior research has made significant progress toward perceiving individual social cues as low-level signals, with a focus on gaze and gesture modalities. Gaze target estimation methods~\cite{chong2020detecting, recasens2015they, ryan2025gaze} can predict where a person is looking and gesture recognition models~\cite{cao2025socialgesture, kapitanov2024hagrid, liu2022ld} can classify hand movements into predefined categories. These works operate at the level of perceptual primitives, answering \textit{what} cue is present. In contrast, this work is the first systematic attempt to address the question of  \textit{how} gaze and gesture should be interpreted in a social context, where meaning arises from the context in which a behavior is produced.

\begin{figure*}[t]
\centering
\includegraphics[width=1.0\textwidth]{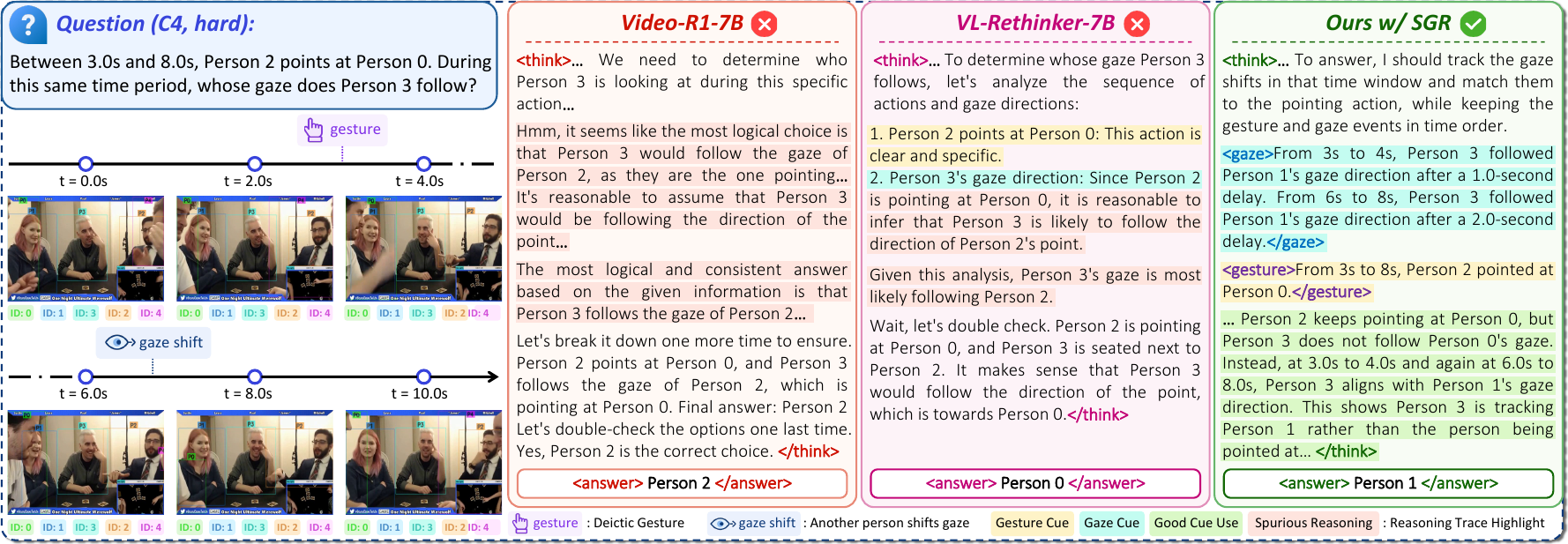}
\vspace*{-0.5cm}
\caption{Example from \textit{GRASP}. Multi-person social reasoning requires grounding subtle non-verbal cues in the correct participants over time. Existing MLLMs~\cite{feng2025video,wang2025vl} often take spurious scene-level shortcuts, whereas ours leverage evidence-aware supervision to reason from the relevant social event.}
\vspace*{-0.4cm}
\label{fig:1}
\end{figure*}

The key hypothesis underlying this work is that modern MLLMs~\cite{liu2023visual, dai2023instructblip, liu2023improved}, which integrate visual perception with strong reasoning capabilities, can be utilized to integrate nonverbal cues and infer their meaning in the context of social understanding. Several works have introduced new social understanding tasks such as multi-party interaction reasoning~\cite{kang2025can, kong2025siv}, speaker target identification~\cite{lee2024modeling, li2025towards, ouyang2025multi}, and social video QA~\cite{mathur2025social,niu2025r}. However, current MLLMs have only limited ability to capture social interactions as grounded visual events: (\lowercase\expandafter{\romannumeral1}) while they can recognize people and actions in the visual scene, they often fail to organize them into temporally grounded social events among participants, making it difficult to identify the relevant cue, participants, and timing of an interaction, \textit{e.g.,} who is looking at, gesturing toward, or responding to whom. (\lowercase\expandafter{\romannumeral2}) Their reasoning often does not attend to the fine-grained visual evidence that determines these relations, relying instead on linguistic or scene-level priors as shown in Fig.~\ref{fig:1}. Effective social interaction understanding requires grounding reasoning in non-verbal cues such as gaze direction, deictic gestures, and their joint coordination across participants, but large-scale supervision that jointly connects these fine-grained cues with high-level social reasoning remains scarce.

In this paper, we close this gap by introducing \textit{G}rounded \textit{R}easoning \textit{A}nd \textit{S}ocial \textit{P}erception (\textit{GRASP}), the first large-scale dataset for grounded social understanding that jointly supports both high-level social reasoning and fine-grained gaze--gesture grounding in multi-person videos. Existing social datasets provide fragmented supervision, focusing on either a single fine-grained cue such as gaze~\cite{recasens2015they, chong2020detecting, ilaslan2023gazevqa, peng2025eye} or gesture~\cite{cao2025socialgesture}, or high-level social QA tasks~\cite{zadeh2019social}, rather than jointly supervising how multiple non-verbal cues ground social reasoning. Building such concurrent supervision manually is prohibitively expensive, requiring temporally precise, identity-consistent annotations of multi-person gaze--gesture dynamics. For scalability, we develop a bottom-up pipeline that extracts structured social interactions from video by detecting gaze and deictic gestures across individuals, organizing them into social events (\textit{e.g.,} mutual gaze, joint attention, gaze following), and modeling their joint compositions where gaze and gesture together convey meaning. \textit{GRASP} contains 290K QA instances organized under a 16-category taxonomy, grounded in 245K interaction events across 46K video clips totaling 749 hours. In addition, we provide \textit{GRASP-Bench}, a curated benchmark of 1K test items.

An additional key question that we address in this work is how such structured social events can provide effective learning signals for MLLMs. Existing RL-based policy optimization approaches (\textit{e.g.,} GRPO~\cite{shao2024deepseekmath,liu2025understanding}) encourage models to generate reasoning before answering, but they do not explicitly tie this reasoning to the relevant social cues. In social scenarios, as exemplified in Fig.~\ref{fig:1}, this can lead models either to produce spurious reasoning traces, consistent with overthinking behavior~\cite{su2025between,wu2025more,chen2026think}, or to arrive at answers through linguistic and scene-level shortcuts. Accordingly, we introduce \textit{S}ocial \textit{G}rounding \textit{R}eward, a learning signal that verifies whether the model's reasoning references the correct participants involved in the underlying social events. Rather than relying on complex reward functions, we provide a direct signal \textit{``Did the model attend to the right person?''} that shifts supervision toward evidence-aware reasoning.

To demonstrate the utility of \textit{GRASP}, we post-train MLLMs with our data and policy, achieving competitive performance on \textit{GRASP-Bench}. Beyond in-domain evaluation, our model also generalizes to three social reasoning benchmarks~\cite{lee2024modeling, li2025towards, lei2020tvqa+}, indicating that social grounding learned from \textit{GRASP} transfers across task formulations. Collectively, these findings demonstrate that large-scale participant-level social data enables MLLMs to move beyond primitive perception and reason about social interactions from non-verbal cues.

Our contribution is three-fold:

\begin{itemize}[leftmargin=1.2em, itemsep=0.2em, topsep=0.2em]
\item We construct \textit{GRASP} and \textit{GRASP-Bench} for training and evaluating MLLMs on grounded social reasoning over multi-person gaze and gesture interactions, using a bottom-up pipeline that scales participant-level social event annotation.
\item We propose \textit{Social Grounding Reward}, a simple yet effective RL signal that directly verifies whether a model's reasoning attends to the correct participants involved in the underlying social events, teaching MLLMs to ground social reasoning in the relevant non-verbal evidence.
\item We demonstrate that our framework achieves strong performance on \textit{GRASP-Bench}, generalizes to other social reasoning benchmarks, and provides detailed analyses of where and why social grounding is most beneficial.
\end{itemize}

\section{Related Works}
\paragraph{Gaze and Gesture in Human Social Behavior.}
Non-verbal cues, particularly gaze and gesture, are fundamental to human social interaction, often indicating intention and the targets of utterances~\cite{lee2024towards,li2018eye}. Interpreting these behaviors has driven significant research in computer vision across two parallel tracks. In gaze understanding, several works have addressed the task of gaze target estimation, leveraging benchmarks such as GazeFollow~\cite{recasens2015they} and VideoAttentionTarget~\cite{chong2020detecting}. Approaches have evolved from fusion-based methods that process depth, pose, and head crops separately~\cite{recasens2015they, jin2022depth, fang2021dual, bao2022escnet, gupta2022modular}, to decoding methods leveraging visual foundation model representations~\cite{song2024vitgaze, ryan2025gaze}. A related task area is identifying social gaze dynamics such as mutual gaze and joint attention~\cite{marin2011here, fan2018inferring, fan2019understanding, gupta2024mtgs}. While some recent methods for gaze target estimation aim to integrate VLM priors into gaze pipelines~\cite{gupta2024exploring,mathew2025gazevlm,pani2025gaze}, specialized models for gaze target estimation remain the dominant approach. In contrast to these gaze-specific models, our work demonstrates how to equip MLLMs with the ability to not only estimate gaze, but jointly leverage it as a cue for higher level social understanding.

Concurrently, gesture recognition has been extensively explored through diverse datasets encompassing RGB-D videos (LD-ConGR~\cite{liu2022ld}), high-resolution images (HaGRID~\cite{kapitanov2024hagrid}), egocentric views (EgoGesture~\cite{zhang2018egogesture}), and multi-person social contexts (SocialGesture~\cite{cao2025socialgesture}). Modeling efforts in this domain have similarly advanced toward leveraging foundation models, including MLLMs and VLMs~\cite{li2025gestura,bossen2025can}. Despite extensive progress in these individual domains, a critical gap remains in jointly modeling gaze and gesture to infer social intention~\cite{cao2025toward}, an integration essential for a holistic understanding of social behavior.

\paragraph{Multi-modal Large Language Models in Social AI.}
Recent advances in MLLMs have demonstrated remarkable video understanding across diverse domains, including multimedia analysis, and human action recognition~\cite{kim2024salova,zhang2025videollama,feng2025video,yan2025videochat,zhao2025humanomni}. Leading proprietary models, such as the Gemini 3~\cite{gemini31pro} and GPT-5~\cite{gpt54} series, alongside open-source alternatives like Qwen-VL series~\cite{bai2025qwen3}, leverage extensive context windows to process complex video inputs. While these models have set new records on challenging video question-answering benchmarks~\cite{fu2025video,fu2026video,hu2025video}, a critical limitation remains: many rely disproportionately on their base language models. This reliance often leads to commonsense biases in social contexts, where models may provide accurate answers based on linguistic patterns even without accessing the visual data~\cite{li2025mimeqa,lee2024towards}. While recent datasets~\cite{zadeh2019social,lai2023werewolf,lee2024modeling,mathur2025social,kong2025siv} have attempted to address this, most existing benchmarks focus on broad social reasoning, overlooking fine-grained interpersonal cues such as gaze shifts and subtle social gestures~\cite{li2025towards,li2026omni,tahboub2025socialfusion,xie2026socialomni,thumu2026social}. In this paper, we bridge this gap by introducing a novel benchmark and a novel post-training strategy designed to enhance the fine-grained social intelligence of MLLMs.

\section{\textit{GRASP} Dataset: Grounded Reasoning And Social Perception}
\label{sec:data}
To enable social reasoning grounded in non-verbal cues, we first curate video sources spanning a broad range of multi-person social interaction types, including multi-party TV dialogue~\cite{lei2020tvqa+,wang2025friends}, social intelligence reasoning~\cite{zadeh2019social}, multi-speaker conversation~\cite{nguyen2025see}, and embodied multi-person behavior~\cite{mclean2025embody}. We further include social deduction gameplay~\cite{lai2023werewolf,cao2025socialgesture}, which we augment with additional online video sources, as such contents contain dense multi-person interactions with rich social signals. Our dataset comprises 46K videos, from which we construct structured representations of social interactions by extracting person-consistent gaze trajectories and deictic gestures and integrating them into higher-order social events. All extracted events are anchored to track person identities, providing structured supervision for both QA generation and our grounding reward. We illustrate the overall collection process in Fig.~\ref{fig:2}, and provide details in Appendix~\ref{appendix:data_detail}.

\subsection{Data Collection Pipeline}
\paragraph{Multi-person Identification \& Gaze Annotation.}
Given a raw video $V$, we first split it into shorter segments $\{V_s\}_{s=1}^{S}$ at scene cuts using PySceneDetect to preserve coherent social interactions. Within each segment, our goal is to obtain \textit{person-consistent} gaze trajectories in multi-party scenes, so that every gaze observation can be retained with a stable identity across time. To do so, we design a multi-stage process that jointly tracks person identities and aligns gaze observations across frames. 

We leverage an off-the-shelf detection model, SAM 3~\cite{carion2025sam} with the text prompt \textit{``people''} to obtain temporally consistent bounding boxes $\mathcal{B}^t_{\text{person}} = \{b_i^t\}_{i \in \mathcal{P}}$, where $\mathcal{P}$ denotes the set of tracked person IDs. While the model provides stable identity tracking, it cannot predict precise facial localization for the subsequent gaze estimation process. Accordingly, we detect face location $\mathcal{B}^t_{\text{face}} = \{f_j^t\}_{j=1}^{M_t}$ using RetinaFace~\cite{deng2020retinaface}, where $M_t$ denotes the number of detected faces at time $t$, and associate each face with a tracked identity using Hungarian matching~\cite{kuhn1955hungarian} between face boxes and the person boxes.

For each person $i \in \mathcal{P}$ at time $t$, we then use the matched face center $c_i^t$ as a point prompt together with the video frame, and feed them to GazeAnywhere~\cite{cao2026gaze} to estimate a normalized gaze point $g_i^t = (x_i^t, y_i^t) \in [0,1]^2$. This yields person-consistent gaze trajectories $\{g_i^t\}_{t=1}^T$. The resulting triplet $\{(i, b_i^t, g_i^t)\}_{t=1}^T$, where $i \in \mathcal{P}$, preserves person identity, spatial location, and gaze target over time, which are then used to define higher-order social events.

\begin{figure*}[t]
\centering
\includegraphics[width=1.0\textwidth]{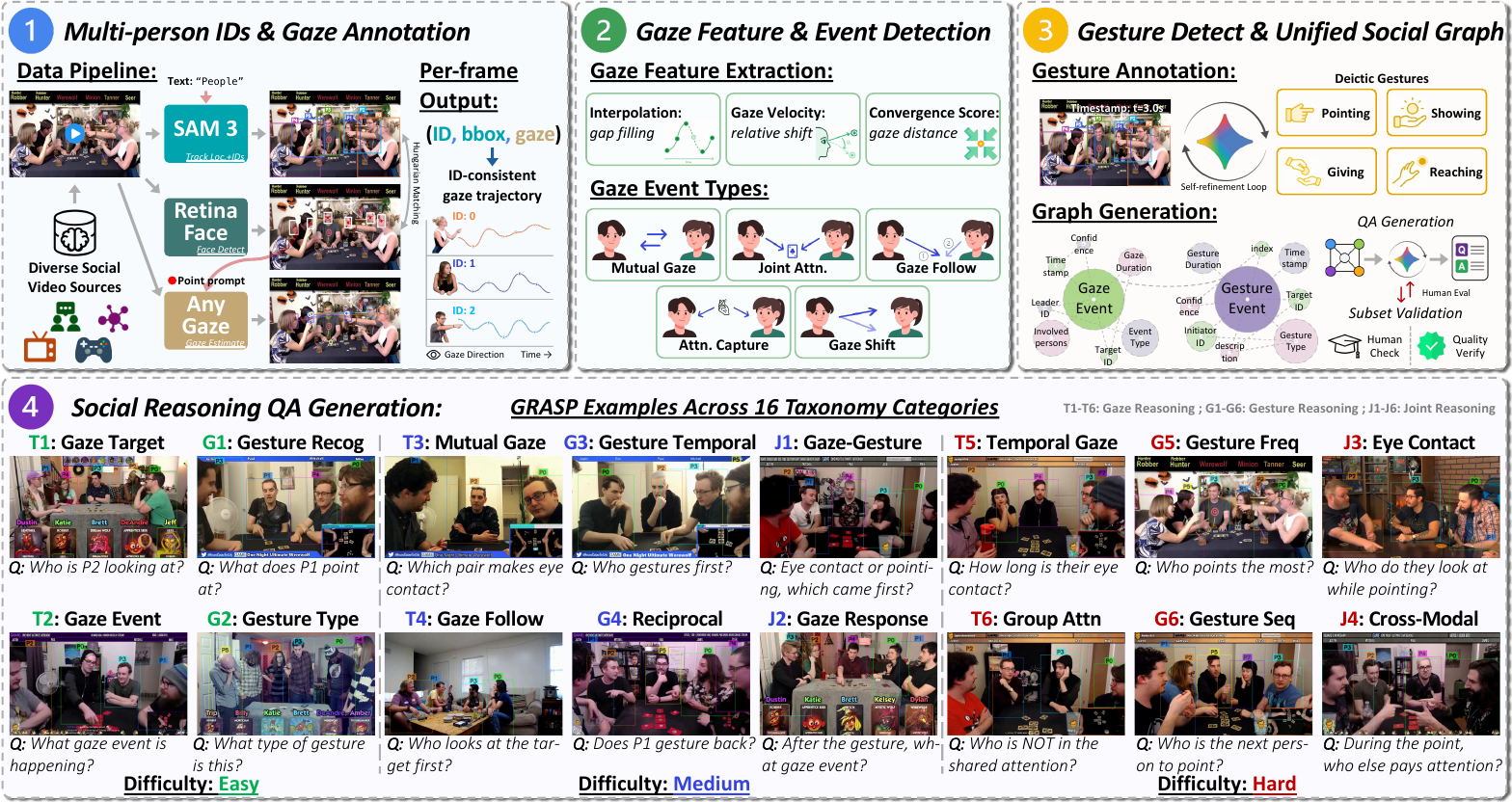}
\vspace*{-0.5cm}
\caption{Overview of the \textit{GRASP} construction pipeline and QA examples. We convert multi-person videos into person-consistent gaze and gesture events, compose them into structured social QA pairs, and apply subset validation with human feedback for quality control.}
\vspace*{-0.2cm}
\label{fig:2}
\end{figure*}

\paragraph{Gaze Feature Extraction \& Event Detection.}
Raw per-frame gaze observations $\{g_i^t\}$ are inherently noisy due to missing detections and camera motion. To obtain robust temporal features, we first apply a simple interpolation scheme within each segment to handle short gaps. We then compute a relative gaze velocity by measuring gaze motion relative to the face center. For each person $i$ at time $t$, we define their gaze direction as $d_i^t = g_i^t - c_i^t$, and estimate velocity as $\|\partial_t d_i^t\|$. To capture group-level attention dynamics, we compute a \textit{convergence score} that measures how closely gaze points cluster at each time step. Specifically, we compute the median distance between individual gaze points $g_i^t$ and their centroid, and map it to a normalized score $s^t = \exp(-\alpha\cdot\text{median}_i \|g_i^t - \bar{g}^t\|)$, where $s^t \in [0,1]$. 

From these features we define five social gaze events, grounded in established taxonomies of gaze behavior~\cite{moore2014joint, jording2018social}: \textit{mutual gaze} (bidirectional gaze between two participants), \textit{joint attention} (multiple participants attending to a common target), \textit{gaze following} (one participant shifting gaze to another’s prior target), \textit{attention capture} (simultaneous gaze shifts across multiple participants), and \textit{sudden gaze shift} (per-person saccade), resulting in 169K gaze events across the corpus (details in Appendix.~\ref{appendix:data_detail}).

\paragraph{Gesture Detection \& Unified Social Graph.}
Robust detection of communicative hand gestures in untrimmed, multi-person video is challenging due to temporal ambiguity, identity confusion, and the lack of precise grounding signals. Following prior gesture taxonomies in social interaction~\cite{mcneill1992hand, cao2025socialgesture}, we focus on deictic gestures and prompt the model to identify four types, namely \textit{pointing}, \textit{showing}, \textit{giving}, and \textit{reaching}. We leverage Gemini~\cite{gemini31pro} as an annotator, augmented with structured visual cues. For each video, we render person-tracked bounding boxes $\mathcal{B}^t_{\text{person}}$ together with their corresponding IDs $i \in \mathcal{P}$ and per-frame timestamps directly onto each frame. This provides explicit spatio-temporal grounding, allowing the model to directly read timestamp from the frame and casting temporal localization into an OCR-like problem, thereby improving the reliability of temporal grounding.

We further employ a self-refinement loop in which low-confidence predictions are iteratively regenerated until convergence, yielding 76K reliable gesture instances with person-level initiator and target identities. We then merge gaze events and gesture annotations into a unified social graph for each video. Each node represents a detected social event with structured metadata, including event type, participating identities, and temporal boundaries, along with modality-specific attributes such as gaze convergence or gesture targets. Joint-modal interactions are captured by linking gaze and gesture events that co-occur within a temporal window, enabling reasoning over joint gaze–gesture patterns. The resulting graph contains 245K events across 46K videos and serves as the foundation for both our QA taxonomy and the grounding supervision used during training.

\subsection{Social Reasoning QA Taxonomy}

\label{sec:qa-taxonomy}
Our dataset is designed to probe not only the \textit{perception} of individual cues, but also the ability to \textit{reason} about social dynamics arising from multi-person interactions. Accordingly, we carefully define a 16-category QA taxonomy with increasing levels of difficulty (examples in Appendix.~\ref{appendix:qualitative_examples}).

\begin{itemize}[leftmargin=1.2em, itemsep=0.2em, topsep=0.2em]
\item \textbf{Gaze reasoning (T1--T6)} spans tasks from perception to inference, including gaze target identification (T1), event classification (T2), mutual gaze recognition (T3), gaze following (T4), temporal reasoning (T5), and group attention dynamics (T6). While T1--T2 focus on identifying \textit{who looks where}, T5--T6 require reasoning over temporal ordering and multi-person attention dynamics.

\item \textbf{Gesture reasoning (G1--G6)} covers gesture recognition (G1), type classification (G2), temporal reasoning (G3), reciprocal patterns (G4), frequency (G5), and sequence chains (G6), with higher-level categories requiring multi-step interaction reasoning.

\item \textbf{Gaze--gesture joint reasoning (J1--J4)} requires integrating gaze and gesture signals to reason over co-occurring non-verbal cues, including temporal alignment (J1), gaze response to gesture (J2), eye contact during interaction (J3), and cross-modal person dynamics (J4).
\end{itemize}

\begin{wrapfigure}{t!}{0.42\textwidth}
\vspace{-4mm}
\centering
\includegraphics[width=0.95\linewidth]{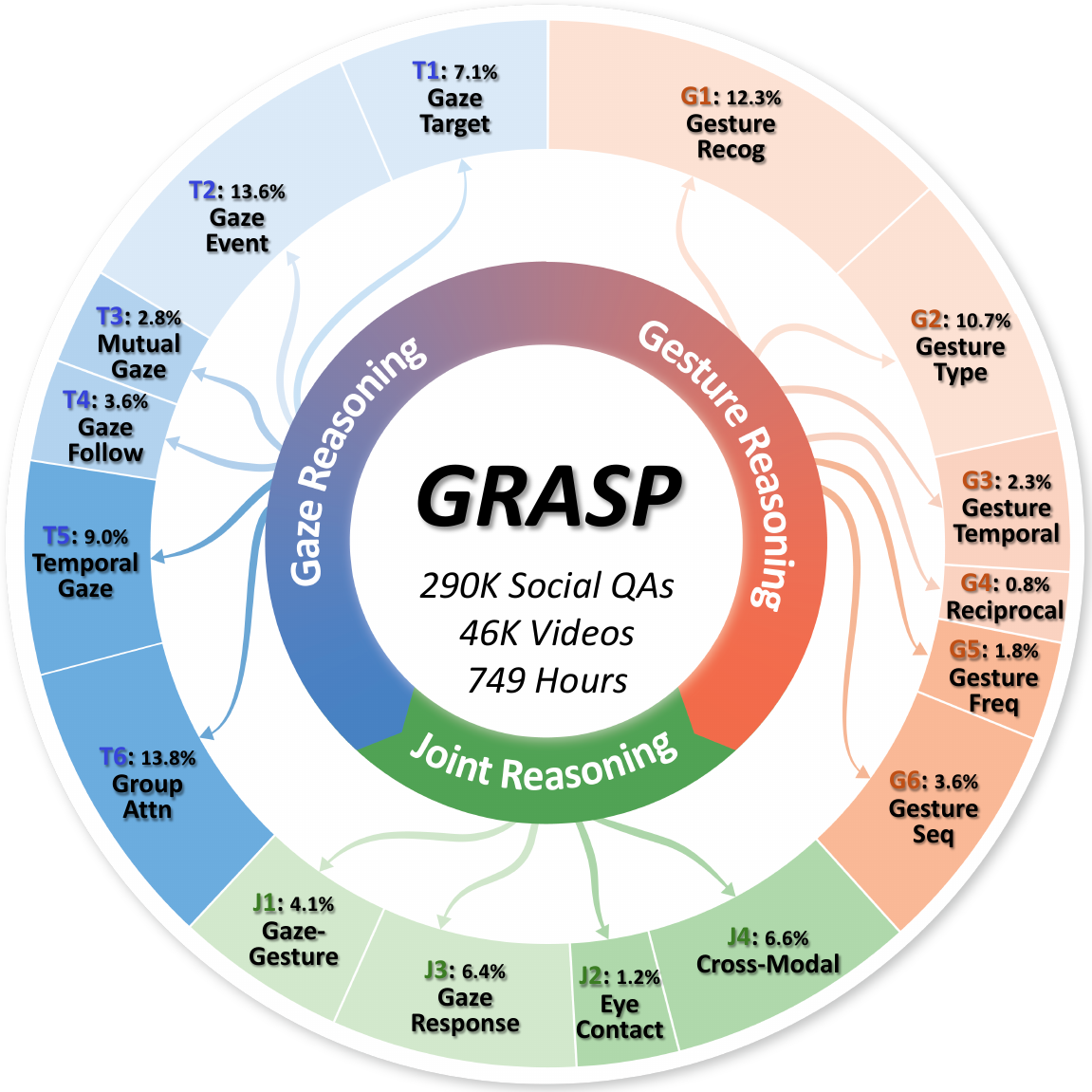}
\vspace{-1mm}
\caption{\textit{GRASP} taxonomy and statistics.}
\vspace{-5mm}
\label{fig:chart}
\end{wrapfigure}

\paragraph{Social Reasoning QA Generation.}
QA pairs are generated from structured event metadata derived from gaze and gesture interactions using a closed-source model~\cite{gemini31pro}. Each question is constructed by querying key attributes such as participant identities, temporal intervals, and interaction types, ensuring that answers are directly verifiable without exhaustive manual annotation. To provide sufficient temporal context, each QA is anchored around a target event and augmented with a short temporal window from its surrounding frames. We design this process to produce diverse questions spanning both perception and higher-order reasoning over multi-person interactions. To ensure data reliability, we further conduct human validation on a subset of generated QA pairs, where sampled QA instances are checked for event consistency, answer verifiability, and ambiguity. The assessment is used to refine the generation and filtering criteria, providing an additional quality-control loop before producing the final dataset of 290K QA pairs grounded in structured gaze and gesture events. We provide examples in Fig.~\ref{fig:2} and statistics in Fig.~\ref{fig:chart} and Appendix.~\ref{appendix:qualitative_examples}. Please see additional details in Appendix~\ref{appendix:stats_detail}.

\section{Learning to Ground Social Reasoning}
\label{sec:method}

\paragraph{Preliminaries.} GRPO~\cite{shao2024deepseekmath} provides an effective framework for training MLLMs with reasoning capabilities~\cite{feng2025video, chen2025scaling, yan2025videochat}. Given a video-question pair query $q$, the policy $\pi_\theta$ generates $K$ reasoning trajectories $\{y_1, \dots, y_K\}$. The policy is optimized by maximizing the following objective:
\begin{equation}
\max_{\pi_\theta} \; \mathbb{E}_{y \sim \pi_{\theta_{\text{old}}}} \left[
\sum_{i=1}^{K} 
\frac{\pi_\theta(y_i \mid q)}{\pi_{\theta_{\text{old}}}(y_i \mid q)} \cdot \hat{A}_i 
\;-\;
\beta \, D_{\mathrm{KL}}(\pi_\theta \,\|\, \pi_{\text{ref}})
\right],
\end{equation}
where $\pi_{\theta_{\text{old}}}$ is the policy from the previous iteration, $\pi_{\text{ref}}$ is a fixed reference (\textit{e.g.,} SFT model), and $\beta$ controls the KL regularization. The advantage for each trajectory is computed relative to the group:
\begin{equation}
\hat{A}_i = \frac{r_i - \mathrm{mean}(\{r_1, \dots, r_K\})}{\mathrm{std}(\{r_1, \dots, r_K\})},
\label{eq:advantage}
\end{equation}
where $r_i$ denotes the reward assigned to trajectory $y_i$. In practice, rewards are typically defined using simple verifiable signals such as answer correctness and output format consistency. Each trajectory is formatted with special tokens, where the model generates intermediate reasoning within \texttt{<think>} and outputs the final answer within \texttt{<answer>}.

\subsection{Training Strategy for Social Reasoning}
We adopt a two-stage training strategy: (\lowercase\expandafter{\romannumeral1}) warm-up the model with structured SFT responses that explicitly separate gaze and gesture evidence and (\lowercase\expandafter{\romannumeral2}) RL post-training with rewards for correctness, format consistency, structural grounding, and social grounding.
 
\paragraph{Stage 1: Structured Format Supervised Fine-Tuning.}
Applying RL directly to the base models induces a cold-start problem, as reasoning about fine-grained non-verbal cues is largely absent from existing multimodal corpora. Thus, stochastic rollouts rarely produce trajectories that reference relevant gaze or gesture events, resulting in a reward landscape that is too sparse to effectively bootstrap learning. Following prior works~\cite{feng2025video, park2025dip} that introduces a warm-up before policy optimization, we first utilize the 52K \textit{GRASP} subset for SFT, where each target response is formatted as:
\[
\texttt{<think>}\,\texttt{<gaze>}\,\mathcal{E}_{\text{gaze}}\,\texttt{</gaze>}\;\texttt{<gesture>}\,\mathcal{E}_{\text{ges}}\,\texttt{</gesture>}\,\texttt{</think>}\,\texttt{<answer>}\,a\,\texttt{</answer>},
\]
where $\mathcal{E}_{\text{gaze}}$, $\mathcal{E}_{\text{ges}}$ enumerate the temporally grounded gaze, gesture events drawn from the social graph, and $a$ is an answer for those events. This warm-up teaches models to decompose social reasoning into its non-verbal components before arriving at an answer, yielding a reference model $\pi_{\text{ref}}$ that can generate the dual-tag format and provides a stable starting point for the subsequent RL stage.

\paragraph{Stage 2: Grounded Reinforcement Post-Training.}
Starting from $\pi_{\text{ref}}$, we continue training on the \textit{GRASP} dataset with policy optimization. Given a video-question pair $q$ and a group of $K$ rollouts $\{y_i\}_{i=1}^{K}$, we define the scalar reward for trajectory $y_i$ as a weighted combination of four signals:
\begin{equation}
r_i \;=\; \lambda_{\text{acc}}\, r_i^{\text{acc}} \;+\; \lambda_{\text{fmt}}\, r_i^{\text{fmt}} \;+\; \lambda_{\text{str}}\, r_i^{\text{str}} \;+\; \lambda_{\text{gnd}}\, r_i^{\text{gnd}},
\label{eq:total_reward}
\end{equation}
where $r_i^{\text{acc}}=\mathds{1}[\hat{a}_i=a_i]$ enforces answer correctness, and $r_i^{\text{fmt}}=\mathds{1}[y_i \in \mathcal{F}]$ ensures that the output adheres to the required structure, where $\mathcal{F}$ denotes the set of valid outputs. We further include a structural term $r_i^{\text{str}}$ that preserves the use of \texttt{<gaze>} and \texttt{<gesture>} tags, and a social grounding term $r_i^{\text{gnd}}$ that encourages the model to identify the relevant participants in the interaction.

\subsection{Social Grounding Reward for Social Learning Signal}
To make the reasoning trace reflect non-verbal cues, we introduce a structural grounding term that encourages the model to express its observations through the dual-tags. This enforces a consistent structure where gaze and gesture information is explicitly articulated rather than implicitly inferred:
\begin{equation}
r_i^{\text{str}} \;=\; \mathds{1}\!\left[\,y_i \text{ contains } \langle\texttt{gaze}\rangle \;\text{or/and}\; \langle\texttt{gesture}\rangle \text{ blocks}\,\right].
\end{equation}

We then define a social grounding reward that verifies whether the reasoning refers to the correct participants of the interaction. Unlike perception tasks that rely on categorical labels, social interactions are inherently \textit{relational}, requiring the model to identify involved participants and their interactions.
Let $\mathcal{P}_i^{\text{pred}}$ denote the set of person identities mentioned within the \texttt{<gaze>} and \texttt{<gesture>} blocks of trajectory $y_i$, and let $\mathcal{P}^{\text{gt}}(q)$ denote the set of participants involved in the target social event specified in $q$. The grounding reward is defined using a weighted precision–recall formulation:
\begin{equation}
r_i^{\text{gnd}} = w_i \cdot P(\mathcal{P}_i^{\text{pred}},\mathcal{P}^{\text{gt}}(q)),
\qquad
w_i = 1 + R(\mathcal{P}_i^{\text{pred}},\mathcal{P}^{\text{gt}}(q)),
\end{equation}
where $P(\cdot)$, $R(\cdot)$ measure precision, recall over participant IDs by comparing predicted and target sets based on their overlap. By balancing coverage and selectivity, our reward penalizes both noisy mentions of all visible individuals and overly narrow predictions that miss key participants, which encourages models to identify relevant interacting individuals without introducing irrelevant ones.

\section{Experiments}
\label{sec:experiments}
 
\paragraph{Implementation Details.}
We build our framework on two baselines~\cite{bai2025qwen3, team2026qwen3}. The input videos are sampled at $2$ fps and rendered with person-tracked bounding boxes, where each individual is marked with IDs (\textit{e.g.,} \textit{P0}, \textit{P1}, \ldots) to support grounded references. Training proceeds in two stages: we first fine-tune the models on the $52$K small subset of \textit{GRASP}, followed by our policy optimization on the $238$K MCQ subset with $K\!=\!8$ rollouts with KL regularization. The composite reward is set as $(\lambda_{\text{acc}},\lambda_{\text{fmt}},\lambda_{\text{str}},\lambda_{\text{gnd}})=(1.0, 0.1, 0.05, 0.2)$. All models are trained in bfloat16 on $8\!\times\!$NVIDIA H200 GPUs with DeepSpeed ZeRO-2, and full training details are provided in the Appendix~\ref{appendix:exp}.

\definecolor{headerblue}{HTML}{ECF4FF}
\definecolor{headergray}{HTML}{F2F2F2}
\definecolor{deltagreen}{HTML}{4CAF50} 

\begin{table*}[t]
\centering
\small
\setlength{\tabcolsep}{4.4pt}
\caption{Evaluation results on \textit{GRASP-Bench}. We report accuracy for all 16 categories. Columns are grouped by reasoning type and difficulty level.}
\label{tab:grasp_bench_full}
\resizebox{1.0\linewidth}{!}{
\begin{tabular}{@{}l cc cc cc c !{\vrule} cc cc cc c !{\vrule} cc cc c !{\vrule} c@{}}
\Xhline{2\arrayrulewidth}
\multirow{3}{*}[-4pt]{Method}
& \multicolumn{7}{c !{\vrule}}{\textbf{Gaze Reasoning}}
& \multicolumn{7}{c !{\vrule}}{\textbf{Gesture Reasoning}}
& \multicolumn{5}{c !{\vrule}}{\textbf{Joint Reasoning}}
& \multirow{3}{*}[-4pt]{\textbf{Overall}} \\
\cmidrule(l{2pt}r{2pt}){2-8}
\cmidrule(l{2pt}r{2pt}){9-15}
\cmidrule(l{2pt}r{2pt}){16-20}
& \multicolumn{2}{c}{Easy} 
& \multicolumn{2}{c}{Medium} 
& \multicolumn{2}{c}{Hard} 
& \multirow{2}{*}{Avg.}
& \multicolumn{2}{c}{Easy} 
& \multicolumn{2}{c}{Medium} 
& \multicolumn{2}{c}{Hard} 
& \multirow{2}{*}{Avg.}
& \multicolumn{2}{c}{Medium} 
& \multicolumn{2}{c}{Hard} 
& \multirow{2}{*}{Avg.}
& \\
\cmidrule(l{1pt}r{1pt}){2-3} \cmidrule(l{1pt}r{1pt}){4-5} \cmidrule(l{1pt}r{1pt}){6-7}
\cmidrule(l{1pt}r{1pt}){9-10} \cmidrule(l{1pt}r{1pt}){11-12} \cmidrule(l{1pt}r{1pt}){13-14}
\cmidrule(l{1pt}r{1pt}){16-17} \cmidrule(l{1pt}r{1pt}){18-19}
& T1 & T2 & T3 & T4 & T5 & T6
& & G1 & G2 & G3 & G4 & G5 & G6
& & J1 & J2 & J3 & J4
& & \\
\hline

\rowcolor{headergray} \multicolumn{21}{l}{\textit{Proprietary Models}} \\

Claude Sonnet 4.6~\cite{claudesonnet46}
& {\color{black!50} 17.1} & {\color{black!50} 56.1} & {\color{black!50} 18.2} & {\color{black!50} 11.9} & {\color{black!50} 30.7} & {\color{black!50} 35.7} & {\color{black!50} 30.2}
& {\color{black!50} 42.6} & {\color{black!50} 64.0} & {\color{black!50} 30.6} & {\color{black!50} 39.0} & {\color{black!50} 29.0} & {\color{black!50} 54.5} & {\color{black!50} 46.5}
& {\color{black!50} 38.3} & {\color{black!50} 18.2} & {\color{black!50} 42.1} & {\color{black!50} 36.5} & {\color{black!50} 37.3}
& {\color{black!50} 37.0} \\

GPT-5.4~\cite{gpt54}
& {\color{gray} 38.6} & {\color{gray} 50.9} & {\color{gray} 26.4} & {\color{gray} 23.9} & {\color{gray} 27.3} & {\color{gray} 38.1} & {\color{gray} 34.9}
& {\color{gray} 63.1} & {\color{gray} 69.4} & {\color{gray} 55.1} & {\color{gray} 51.2} & {\color{gray} 38.7} & {\color{gray} 59.1} & {\color{gray} 60.4}
& {\color{gray} 39.1} & {\color{gray} 31.8} & {\color{gray} 40.8} & {\color{gray} 39.2} & {\color{gray} 39.0}
& {\color{gray} 43.9} \\

Gemini 3.1 Pro~\cite{gemini31pro}
& {\color{black!50} 42.9} & {\color{black!50} 54.4} & {\color{black!50} 49.1} & {\color{black!50} 40.3} & {\color{black!50} 28.4} & {\color{black!50} 29.8} & {\color{black!50} 41.8}
& {\color{black!50} 63.1} & {\color{black!50} 84.7} & {\color{black!50} 61.2} & {\color{black!50} 56.1} & {\color{black!50} 48.4} & {\color{black!50} 68.2} & {\color{black!50} 67.6}
& {\color{black!50} 42.6} & {\color{black!50} 50.0} & {\color{black!50} 44.7} & {\color{black!50} 44.6} & {\color{black!50} 44.3}
& {\color{black!50} 50.5} \\
\hline

\rowcolor{headergray} \multicolumn{21}{l}{\textit{Supervised Fine-Tuned Models}} \\
Qwen2.5-VL-7B-Instruct~\cite{bai2025qwen2}
& 30.0 & 55.3 & 27.3 & 28.4 & 27.3 & 31.0 & 34.3
& 41.0 & 48.6 & 36.7 & 34.1 & 22.6 & 45.5 & 40.7
& 33.9 & 31.8 & 31.6 & 36.5 & 33.8
& 36.2 \\

InternVL3.5-8B-Instruct~\cite{wang2025internvl3}
& 30.0 & 31.6 & 30.0 & 28.4 & 36.4 & \textbf{34.5} & 31.9
& 46.7 & 53.2 & 26.5 & 34.1 & 19.4 & 40.9 & 42.0
& 43.5 & \textbf{45.5} & 31.6 & 35.1 & 38.3
& 36.6 \\

LLaVA-OV-1.5-8B-Instruct~\cite{an2025llava}
& 24.3 & 36.8 & 33.6 & 17.9 & 30.7 & 29.8 & 30.0
& 35.2 & 82.9 & 32.7 & 22.0 & \underline{35.5} & 27.3 & 47.1
& 42.6 & 31.8 & 35.5 & 35.1 & 38.0
& 37.3 \\

Qwen3-VL-8B-Instruct~\cite{bai2025qwen3}
& 24.3 & 35.1 & 25.5 & 14.9 & 31.8 & \underline{33.3} & 28.3
& 45.1 & 86.5 & 28.6 & 41.5 & 22.6 & 40.9 & 52.7
& 33.9 & 31.8 & 35.5 & \underline{48.6} & 38.0
& 38.3 \\

Qwen3.5-9B (Instruct Mode)~\cite{team2026qwen3}
& 22.9 & 52.6 & 30.0 & 22.4 & 34.1 & 25.0 & 32.8
& \underline{51.6} & \underline{87.4} & 40.8 & \textbf{46.3} & 25.8 & \textbf{68.2} & \underline{59.0}
& 40.0 & 31.8 & 39.5 & 45.9 & 40.8
& 43.0 \\

\hline

\rowcolor{headergray} \multicolumn{10}{l}{\textit{RL Post-Training Models}} \\
VL-Rethinker-7B~\cite{wang2025vl}
& 24.3 & 36.0 & 22.7 & 25.4 & 39.8 & 27.4 & 29.6
& 36.1 & 49.5 & 38.8 & 29.3 & 25.8 & 22.7 & 38.0
& 32.2 & 27.3 & 36.8 & \textbf{50.0} & 37.6
& 34.2 \\

VideoChat-R1.5-7B~\cite{yan2025videochat}
& 32.9 & 39.5 & 25.5 & 35.8 & 30.7 & 32.1 & 32.6
& 44.3 & 29.7 & 36.7 & \underline{43.9} & 12.9 & 27.3 & 35.4
& 33.9 & 27.3 & 31.6 & 36.5 & 33.4
& 33.7 \\

Video-R1-7B~\cite{feng2025video}
& 30.0 & 51.8 & 38.2 & 28.4 & 35.2 & 31.0 & 37.1
& 42.6 & 29.7 & 36.7 & 26.8 & 25.8 & 40.9 & 34.8
& 36.5 & 22.7 & 30.3 & 47.3 & 36.6
& 36.3 \\

LongVILA-R1-7B~\cite{chen2025scaling}
& 28.6 & 29.8 & 32.7 & 20.9 & 30.7 & 28.6 & 29.1
& 29.5 & 26.1 & 28.6 & 24.4 & 32.3 & 36.4 & 28.5
& 37.4 & 36.4 & 38.2 & 24.3 & 34.1
& 30.1 \\

InternVL3.5-8B~\cite{wang2025internvl3}
& 20.0 & 27.2 & 31.8 & 19.4 & 28.4 & 23.8 & 25.9
& 46.7 & 55.9 & 32.7 & 31.7 & 16.1 & 36.4 & 42.8
& 36.5 & 31.8 & 38.2 & 37.8 & 36.9
& 33.9 \\

LLaVA-OV-1.5-8B-RL~\cite{an2025llava}
& 22.9 & 30.7 & 30.0 & 26.9 & 27.3 & 29.8 & 28.3
& 31.1 & 76.6 & 32.7 & 19.5 & 12.9 & 22.7 & 41.5
& 42.6 & \underline{40.9} & 30.3 & 28.4 & 35.5
& 34.2 \\

Qwen3-VL-8B-Thinking~\cite{bai2025qwen3}
& 31.4 & 36.0 & 23.6 & 16.4 & 30.7 & 27.4 & 28.1
& 41.0 & 74.8 & 40.8 & 34.1 & 19.4 & 40.9 & 48.4
& 40.9 & 36.4 & 39.5 & 32.4 & 38.0
& 36.9 \\

Qwen3.5-9B (Thinking Mode)~\cite{team2026qwen3}
& \underline{32.9} & 21.9 & 14.5 & 19.4 & 22.7 & \textbf{34.5} & 23.6
& 45.9 & 66.7 & 32.7 & 29.3 & 25.8 & 27.3 & 45.7
& 24.3 & 31.8 & 30.3 & 31.1 & 28.2
& 31.7 \\
\hline

\rowcolor{headerblue} \multicolumn{21}{l}{\textit{Socially-Grounded RL (Ours)}} \\
Qwen3-VL-8B + \textit{SGR}
& \textbf{37.1} & \underline{65.8} & \underline{47.3} & \underline{41.8} & \underline{43.2} & \underline{33.3} & \underline{46.3}
& 50.8 & \textbf{90.1} & 38.8 & 39.0 & 25.8 & 59.1 & 58.0
& \textbf{53.9} & 22.7 & \textbf{51.3} & 43.2 & \underline{48.1}
& \underline{50.4} \\

Qwen3.5-9B + \textit{SGR}
& \textbf{37.1} & \textbf{70.2} & \textbf{50.0} & \textbf{46.3} & \textbf{46.6} & 27.4 & \textbf{48.0}
& \textbf{54.9} & \textbf{90.1} & \textbf{59.2} & \textbf{46.3} & \textbf{41.9} & \underline{63.6} & \textbf{64.4}
& \underline{48.7} & \underline{40.9} & \underline{48.7} & 39.2 & \textbf{45.6}
& \textbf{52.6} \\
\Xhline{2\arrayrulewidth}
\end{tabular}
}
\end{table*}


\paragraph{Experimental Setting.}
We evaluate on \textit{GRASP-Bench}, a curated test set covering 16 categories, as our in-domain benchmark, and on three public social-reasoning benchmarks to assess cross-benchmark generalization: MMSI~\cite{lee2024modeling}, Online-MMSI~\cite{li2025towards}, and TVQA+~\cite{lei2020tvqa+}. MMSI probes multi-party social reasoning through three tasks, including speaker target identification (STI), pronoun coreference resolution (PCR), and mentioned-player prediction (MPP), while Online-MMSI evaluates the same tasks under a causally truncated setting that admits only past frames. TVQA+ measures spatio-temporally grounded video QA over multi-character dialogues. We compare against three groups of MLLMs: (\lowercase\expandafter{\romannumeral1}) proprietary models~\cite{claudesonnet46,gpt54,gemini31pro}, (\lowercase\expandafter{\romannumeral2}) instruction-tuned models~\cite{bai2025qwen2,wang2025internvl3,an2025llava,bai2025qwen3,team2026qwen3}, and (\lowercase\expandafter{\romannumeral3}) RL-based reasoning models~\cite{wang2025vl,yan2025videochat,feng2025video,chen2025scaling,wang2025internvl3,an2025llava,bai2025qwen3,team2026qwen3}.

\subsection{Main Results on \textit{GRASP-Bench}}
We report the main results on \textit{GRASP-Bench} in Tab.~\ref{tab:grasp_bench_full}. Overall, our socially-grounded RL consistently outperforms both supervised fine-tuned models and existing RL post-training baselines across all reasoning types. The gains are particularly evident in gaze and gesture reasoning, where accurate identification of interacting individuals is critical, as well as in cross-modal reasoning that requires integrating multiple social cues. These results indicate that explicitly grounding reasoning in non-verbal evidence leads to more reliable and coherent social understanding, rather than relying on coarse visual perception or language priors.

A notable trend is that SFT models are generally more competitive than RL baselines, suggesting that generic RL post-training for longer reasoning does not necessarily improve social understanding. Without an explicit grounding signal, additional reasoning can amplify spurious assumptions and distract models from the non-verbal cues and participant relations that determine the answer. In contrast, \textit{SGR} rewards reasoning that refers to the correct participants in the underlying event, retaining useful deliberation while suppressing unsupported shortcuts. We provide qualitative comparisons in Appendix~\ref{appendix:qualitative}, where RL baselines often produce plausible but ungrounded rationales, whereas our method reasons from the relevant people and non-verbal cues.

\subsection{Generalization to Other Social Understanding Datasets}
We evaluate cross-benchmark generalization on other social understanding datasets~\cite{lee2024modeling,li2025towards,lei2020tvqa+} in a zero-shot setting, without further task-specific fine-tuning. As shown in Tab.~\ref{tab:cross_benchmark}, \textit{SGR}-trained models outperform or remain competitive with corresponding instruction-tuned backbones and RL post-training baselines. Given that our model's post-training is performed only on \textit{GRASP}, these results indicate that (\lowercase\expandafter{\romannumeral1}) the models do not over-specialize to the \textit{GRASP-Bench} format and (\lowercase\expandafter{\romannumeral2}) \textit{SGR} training is beneficial for broader social reasoning tasks beyond \textit{GRASP}'s gaze and gesture focused questions. Furthermore, we provide qualitative reasoning-trace comparisons for these benchmarks in Appendix~\ref{appendix:qualitative_baselines}, demonstrating that \textit{SGR}-trained models effectively leverage gaze and gesture as intermediate cues on these benchmarks.

\begin{table*}[t]
\centering
\small
\caption{Cross-benchmark generalization results on MMSI, Online-MMSI, and TVQA+. MMSI and Online-MMSI are evaluated on three social reasoning tasks: STI, PCR, and MPP.}
\label{tab:cross_benchmark}
\resizebox{0.7\linewidth}{!}{
\begin{tabular}{@{}l cccc !{\vrule} cccc !{\vrule} c@{}}
\Xhline{2\arrayrulewidth}
\multirow{2}{*}[-4pt]{Method}
& \multicolumn{4}{c !{\vrule}}{\textbf{MMSI}}
& \multicolumn{4}{c !{\vrule}}{\textbf{Online-MMSI}}
& \textbf{TVQA+} \\
\cmidrule(l{2pt}r{2pt}){2-5}
\cmidrule(l{2pt}r{2pt}){6-9}
\cmidrule(l{2pt}r{2pt}){10-10}
& STI & PCR & MPP & Avg.
& STI & PCR & MPP & Avg.
& Acc. \\
\hline
\rowcolor{headergray} \multicolumn{10}{l}{\textit{Supervised Fine-Tuned Models}} \\
Qwen2.5-VL-7B-Instruct~\cite{bai2025qwen2}
& 57.4 & 64.1 & \textbf{52.3} & 57.4 
& 55.2 & 65.5 & 42.8 & 53.7 
& 70.4 \\ 
InternVL3.5-8B-Instruct~\cite{wang2025internvl3}
& 51.5 & 58.3 & 50.3 & 53.0
& 47.4 & 60.6 & \underline{44.0} & 49.9
& 67.1 \\
LLaVA-OV-1.5-8B-Instruct~\cite{an2025llava}
& 64.9 & 65.8 & \underline{50.9} & 60.2 
& 55.8 & 63.3 & 41.1 & 52.7 
& 72.5 \\
Qwen3-VL-8B-Instruct~\cite{bai2025qwen3}
& 67.5 & 63.6 & 48.7 & 59.7
& 59.3 & 66.6 & 43.9 & 55.9
& \underline{71.9} \\
Qwen3.5-9B (Instruct Mode)~\cite{team2026qwen3}
& 67.9 & \textbf{68.2} & 50.1 & \underline{61.6}
& 56.6 & 68.0 & 43.6 & 55.2
& 70.6 \\
\hline

\rowcolor{headergray} \multicolumn{10}{l}{\textit{RL Post-Training Models}} \\
InternVL3.5-8B~\cite{wang2025internvl3}
& 49.2 & 42.7 & 37.5 & 43.2
& 43.8 & 58.1 & 43.3 & 47.6
& 70.9 \\
VideoChat-R1.5-7B~\cite{yan2025videochat}
& 60.4 & 65.8 & 46.9 & 57.1
& 56.9 & \underline{68.3} & 38.5 & 53.5
& 69.9 \\
LongVILA-R1-7B~\cite{chen2025scaling}
& 50.5 & 44.6 & 30.5 & 41.7
& 40.9 & 45.7 & 28.4 & 37.8
& 59.4 \\
Video-R1-7B~\cite{feng2025video}
& 58.6 & 60.8 & 48.8 & 55.7
& 55.0 & 63.6 & 40.3 & 52.2
& 71.2 \\
LLaVA-OV-1.5-8B-RL~\cite{an2025llava}
& 63.3 & 59.8 & 44.1 & 55.5
& 52.8 & 57.6 & 38.1 & 48.9
& 68.4 \\
VL-Rethinker-7B~\cite{wang2025vl}
& 62.1 & 62.7 & 47.1 & 56.9
& 56.3 & 64.1 & 41.8 & 53.3
& 66.8 \\
Qwen3-VL-8B-Thinking~\cite{bai2025qwen3}
& \underline{70.4} & 57.0 & 48.8 & 59.0
& \underline{62.0} & 60.8 & 42.5 & 54.7
& 70.0 \\
Qwen3.5-9B (Thinking Mode)~\cite{team2026qwen3}
& 70.2 & 61.9 & 48.2 & 60.0
& \textbf{63.1} & 61.4 & 39.7 & 54.3
& 65.6 \\
\hline

\rowcolor{headerblue} \multicolumn{10}{l}{\textit{Socially-Grounded RL (Ours)}} \\
Qwen3-VL-8B + \textit{SGR}
& 67.4 & 63.6 & 49.8 & 60.1
& 60.6 & 63.9 & \textbf{45.1} & \underline{56.0}
& \textbf{73.2} \\
Qwen3.5-9B + \textit{SGR}
& \textbf{71.2} & \underline{66.9} & 49.2 & \textbf{62.1}
& 60.3 & \textbf{70.4} & 42.8 & \textbf{56.9}
& 70.9 \\
\Xhline{2\arrayrulewidth}
\end{tabular}
}
\end{table*}

\begin{table*}[t]
\centering
\begin{minipage}[t]{0.58\linewidth}
  \newcolumntype{C}[1]{>{\centering\arraybackslash}p{#1}}

\centering
\small
\caption{Ablation on reward components for social grounding and their impact on social reasoning performance.}
\label{tab:ablation_component}
\setlength{\tabcolsep}{2.5pt}
\resizebox{0.99\linewidth}{!}{
\begin{tabular}{@{}l !{\vrule} C{0.055\linewidth} C{0.055\linewidth} C{0.055\linewidth} C{0.055\linewidth} !{\vrule} ccc !{\vrule} ccc !{\vrule} c@{}}
\Xhline{2\arrayrulewidth}
\multirow{2}{*}{Method}
& \multicolumn{4}{c !{\vrule}}{\textbf{Reward Comp.}}
& \multicolumn{7}{c}{\textbf{\textit{GRASP-Bench}}} \\
\cmidrule(l{2pt}r{2pt}){2-5}
\cmidrule(l{2pt}r{2pt}){6-12}
& $r_i^{\text{acc}}$ & $r_i^{\text{fmt}}$ & $r_i^{\text{str}}$ & $r_i^{\text{gnd}}$
& Gaze R. & Gest. R. & Joint R. & Easy & Med & Hard & Avg. \\
\hline
\rowcolor{headerblue} \multicolumn{12}{l}{Baseline: \textit{Qwen3-VL-8B}} \\
Instruct (SFT)
& -- & -- & -- & --
& 28.3 & 52.7 & 38.0 & 49.9 & 28.5 & 36.0 & 38.3 \\
\textit{GRPO} (RFT)
& \Checkmark & \Checkmark & -- & --
& 34.3 & 50.8 & 41.1 & 56.1 & 32.4 & 33.9 & 41.1 \\
\textit{SGR} (RFT)
& \Checkmark & \Checkmark & \Checkmark & \Checkmark
& \textbf{46.3} & \textbf{58.0} & \textbf{48.1} & \textbf{63.1} & \textbf{45.0} & \textbf{42.1} & \textbf{50.4} \\
\hline
\rowcolor{headerblue} \multicolumn{12}{l}{Baseline: \textit{Qwen3.5-9B}} \\
Instruct (SFT)
& -- & -- & -- & --
& 32.8 & 59.0 & 40.8 & 56.6 & 34.7 & 36.8 & 43.0 \\
\textit{GRPO} (RFT)
& \Checkmark & \Checkmark & -- & --
& 34.1 & 56.6 & 48.4 & 57.7 & 36.9 & 39.5 & 44.6 \\
\textit{SGR} (RFT)
& \Checkmark & \Checkmark & \Checkmark & \Checkmark
& \textbf{48.0} & \textbf{64.4} & \textbf{45.6} & \textbf{65.5} & \textbf{49.3} & \textbf{41.9} & \textbf{52.6} \\
\Xhline{2\arrayrulewidth}
\end{tabular}
}
\end{minipage}
\hspace{0.01\linewidth}
\begin{minipage}[t]{0.39\linewidth}
  \centering
\small
\caption{Model scaling effects with \textit{SGR} on \textit{GRASP-Bench} evaluation.}
\label{tab:model_scaling}
\setlength{\tabcolsep}{1.0pt}
\resizebox{1.0\linewidth}{!}{
\begin{tabular}{@{}l c !{\vrule} ccc !{\vrule} c@{}}
\Xhline{2\arrayrulewidth}
\multirow{2}{*}{Model} & \multirow{2}{*}{\#Params}
& \multicolumn{3}{c !{\vrule}}{\textbf{\textit{GRASP-Bench}}}
& \multirow{2}{*}{\textbf{Avg.}} \\
\cmidrule(l{2pt}r{2pt}){3-5}
& & Gaze R. & Gest. R. & Joint R. & \\
\hline
\rowcolor{headerblue} \multicolumn{6}{l}{\textit{Baseline: Qwen3-VL}} \\
Q3VL + \textit{SGR} & 2B & 43.0 & 56.9 & 38.0 & 46.2 \\
Q3VL + \textit{SGR} & 4B & 43.2 & 58.5 & 44.6 & 48.3 \\
Q3VL + \textit{SGR} & 8B & 46.3 & 58.0 & 48.1 & 50.4 \\
\hline
\rowcolor{headerblue} \multicolumn{6}{l}{\textit{Baseline: Qwen3.5}} \\
Q3.5 + \textit{SGR} & 2B & 41.4 & 57.2 & 36.8 & 45.3 \\
Q3.5 + \textit{SGR} & 4B & 46.2 & 62.0 & 41.4 & 48.5 \\
Q3.5 + \textit{SGR} & 9B & 48.0 & 64.4 & 45.6 & 52.6 \\

\Xhline{2\arrayrulewidth}
\end{tabular}
} 
\end{minipage}
\end{table*}


\subsection{Ablation and Diagnostic Analysis}
\paragraph{Detailed Comparison with GRPO.}
We conduct a detailed comparison with GRPO~\cite{shao2024deepseekmath} to understand whether improvements arise from stronger optimization alone or from explicitly grounding social reasoning. While the RL method encourages models to generate reasoning before answering, it does not enforce that such reasoning is aligned with the correct social evidence. As a result, models often produce verbose but ungrounded reasoning traces that fail to identify the relevant individuals involved in social interactions. As shown in Tab.~\ref{tab:ablation_component}, incorporating GRPO on top of the baseline yields moderate improvements, indicating that reasoning supervision alone provides some benefit. However, adding our \textit{SGR} leads to substantially larger gains across all metrics, highlighting the importance of explicitly grounding reasoning in non-verbal cues. 

\begin{figure}[t]
  \centering
  \begin{minipage}[c]{0.42\linewidth}
    \centering
    \includegraphics[width=\linewidth]{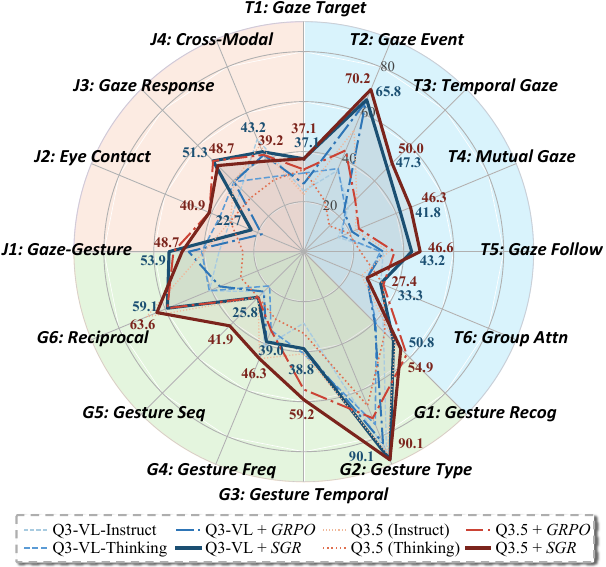}
    \captionof{figure}{Per-category comparison on \textit{GRASP-Bench} across baselines including SFT, reasoning, GRPO-tuned, and ours.}
    \label{fig:radar}
  \end{minipage}
  \hfill
  \begin{minipage}[c]{0.54\linewidth}
    \centering
    \includegraphics[width=0.95\linewidth]{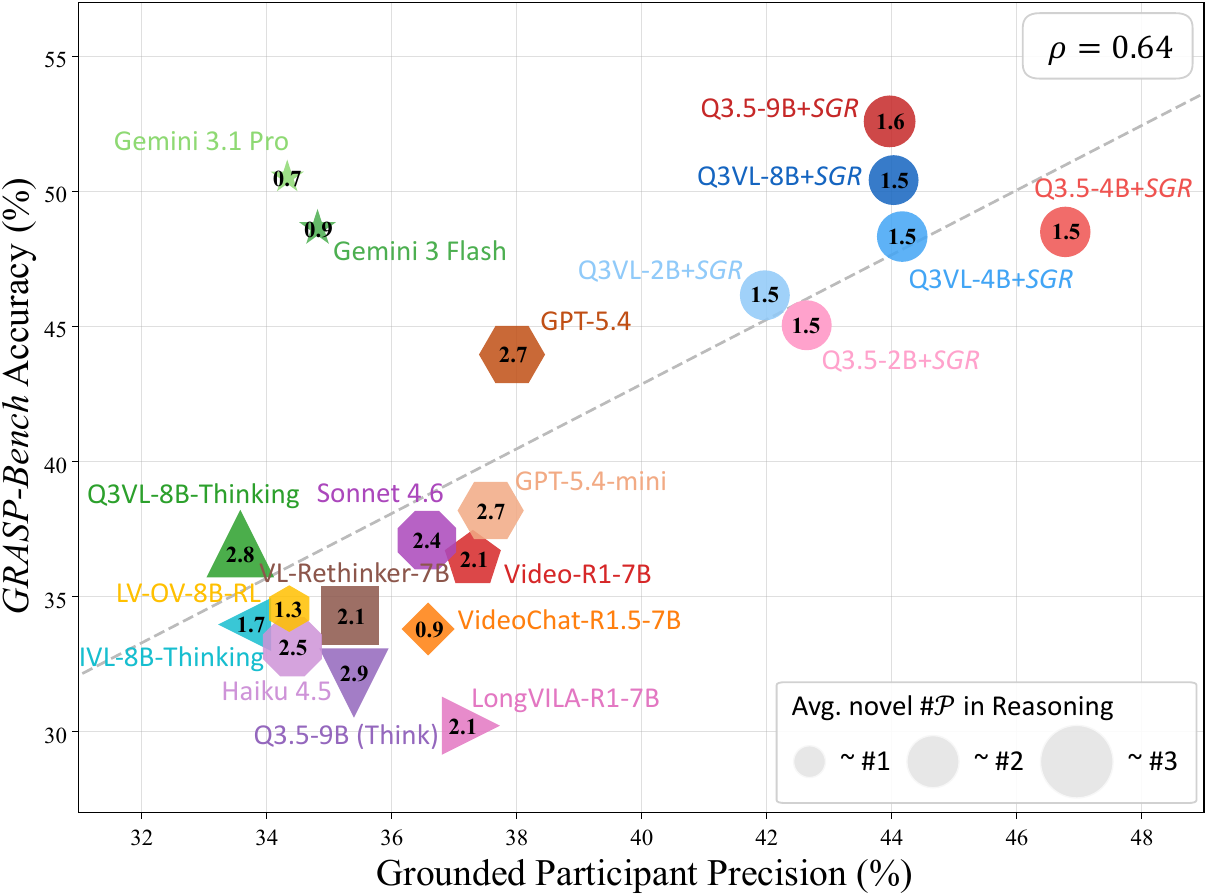}
    \captionof{figure}{Grounded participant precision\textemdash accuracy on \textit{GRASP-Bench} across various reasoning baselines. Marker size reflects the average number of novel participants mentioned in the reasoning trace.}
    \label{fig:grounding_analysis}
  \end{minipage}
\end{figure}

Fig.~\ref{fig:radar} further provides a detailed breakdown across all categories. While instruction model and zero-shot reasoning models exhibit inconsistent performance, and GRPO improves certain categories, it still struggles to reliably capture fine-grained social cues. In contrast, ours consistently improves performance, demonstrating that our grounding reasoning leads to coherent social understanding.

\paragraph{Scaling with Model Size.}
We analyze model-size scaling with \textit{SGR} in Tab.~\ref{tab:model_scaling} and compare the corresponding grounding behavior in Fig.~\ref{fig:grounding_analysis}. Across the baseline backbones~\cite{bai2025qwen3, team2026qwen3}, average accuracy improves with model size, increasing from 46.2 to 50.4 and from 45.3 to 52.6, respectively. In Fig.~\ref{fig:grounding_analysis}, the scaled \textit{SGR}-trained models remain in the high grounded-participant-precision region while achieving increasingly strong \textit{GRASP-Bench} accuracy. This shows that the scaling gains come together with participant-grounded reasoning, rather than only stronger answer prediction.

\paragraph{Grounding Precision Analysis.}
To verify whether our reward indeed encourages models to identify the correct participants, we analyze grounding precision across reasoning models. We measure the proportion of correctly mentioned participants in the reasoning trace, alongside overall accuracy. As shown in Fig.~\ref{fig:grounding_analysis}, the precision across baselines exhibits positive correlation with accuracy ($\rho = 0.64$), indicating that correctly identifying relevant individuals is closely tied to successful social reasoning. We further observe distinct differences in grounding behavior across models. Many RL baselines achieve moderate accuracy despite low grounding precision, often mentioning multiple irrelevant participants, which leads to noisy reasoning traces and higher false positives. In contrast, our method achieves higher grounding precision with fewer participant mentions, reflecting more selective and targeted reasoning. This behavior aligns with the objective of \textit{SGR}, which explicitly encourages the model to focus on the correct interacting individuals rather than enumerating all visible candidates.

\section{Discussion and Conclusion}
\paragraph{Discussion.}
We introduced \textit{GRASP} as a large-scale social reasoning dataset grounded in fine-grained non-verbal events. Constructing such supervision is inherently challenging: despite our quality-control pipeline and human validation (Appendix~\ref{app:quality_control}), the resulting annotations can still be noisy in ambiguous scenes. Moreover, we propose \textit{SGR}, a simple but effective reward signal that encourages models to ground reasoning in the participants involved in the target interaction. We also examine a possible ID-copying shortcut through ID corruption analysis (Appendix~\ref{app:id_corruption}); the observed degradation indicates that copying question IDs alone does not explain the learned behavior. Still, \textit{SGR} mainly verifies participant-level grounding, and incorporating stricter temporal alignment and richer social cues remains important future work. Broader limitations are discussed in Appendix~\ref{appendix:broader_impacts}.

\paragraph{Conclusion.}
We introduced \textit{GRASP}, a large-scale dataset for grounded social reasoning that connects high-level social QA with fine-grained gaze and gesture events in multi-person videos. \textit{GRASP} provides 290K QA pairs over 46K videos, organized by a 16-category taxonomy, together with \textit{GRASP-Bench} for social reasoning evaluation. Moreover, we proposed \textit{SGR}, a reward signal that uses structured social events to encourage models to reason about the participants involved in each interaction. Experiments show that participant-grounded supervision improves performance on \textit{GRASP-Bench} while preserving zero-shot performance on related social video QA tasks.

{
\small
\bibliography{reference}

@String(CVPR= {IEEE Conf. Comput. Vis. Pattern Recog.})

@String(ECCV= {Eur. Conf. Comput. Vis.})

@String(BMVC= {Brit. Mach. Vis. Conf.})

@String(TMM  = {IEEE Trans. Multimedia})

@String(AAAI = {AAAI})

@String(CVPRW= {IEEE Conf. Comput. Vis. Pattern Recog. Worksh.})

@String(CVPR  = {CVPR})

@String(ECCV  = {ECCV})

@String(BMVC  =	{BMVC})

@String(TMM   =	{IEEE TMM})

@String(CVPRW= {CVPRW})

@inproceedings{ilaslan2023gazevqa,   title={Gazevqa: A video question answering dataset for multiview eye-gaze task-oriented collaborations},   author={Ilaslan, Muhammet and Song, Chenan and Chen, Joya and Gao, Difei and Lei, Weixian and Xu, Qianli and Lim, Joo and Shou, Mike},   booktitle={Proceedings of the 2023 Conference on Empirical Methods in Natural Language Processing},   pages={10462--10479},   year={2023} }

@article{peng2025eye,
  title={In the eye of mllm: Benchmarking egocentric video intent understanding with gaze-guided prompting},
  author={Peng, Taiying and Hua, Jiacheng and Liu, Miao and Lu, Feng},
  journal={arXiv preprint arXiv:2509.07447},
  year={2025}
}

@article{kendrick2023turn,
  title={Turn-taking in human face-to-face interaction is multimodal: gaze direction and manual gestures aid the coordination of turn transitions},
  author={Kendrick, Kobin H and Holler, Judith and Levinson, Stephen C},
  journal={Philosophical transactions of the royal society B},
  volume={378},
  number={1875},
  pages={20210473},
  year={2023},
  publisher={The Royal Society}
}

@article{frith2012mechanisms,
  title={Mechanisms of social cognition},
  author={Frith, Chris D and Frith, Uta},
  journal={Annual review of psychology},
  volume={63},
  pages={287--313},
  year={2012},
  publisher={Annual Reviews}
}

@article{wu2023tidybot,
  title={Tidybot: Personalized robot assistance with large language models},
  author={Wu, Jimmy and Antonova, Rika and Kan, Adam and Lepert, Marion and Zeng, Andy and Song, Shuran and Bohg, Jeannette and Rusinkiewicz, Szymon and Funkhouser, Thomas},
  journal={Autonomous Robots},
  volume={47},
  number={8},
  pages={1087--1102},
  year={2023},
  publisher={Springer}
}

@article{shum2018eliza,
  title={From Eliza to XiaoIce: challenges and opportunities with social chatbots},
  author={Shum, Heung-Yeung and He, Xiao-dong and Li, Di},
  journal={Frontiers of Information Technology \& Electronic Engineering},
  volume={19},
  pages={10--26},
  year={2018},
  publisher={Springer}
}

@book{mcneill1992hand,
  title={Hand and mind: What gestures reveal about thought},
  author={McNeill, David},
  year={1992},
  publisher={University of Chicago press}
}

@inproceedings{liu2022ld,
  title={LD-ConGR: A large RGB-D video dataset for long-distance continuous gesture recognition},
  author={Liu, Dan and Zhang, Libo and Wu, Yanjun},
  booktitle={CVPR},
  pages={3304--3312},
  year={2022}
}

@inproceedings{kapitanov2024hagrid,
  title={HaGRID--HAnd Gesture Recognition Image Dataset},
  author={Kapitanov, Alexander and Kvanchiani, Karina and Nagaev, Alexander and Kraynov, Roman and Makhliarchuk, Andrei},
  booktitle={WACV},
  pages={4572--4581},
  year={2024}
}

@article{zhang2018egogesture,
  title={EgoGesture: A new dataset and benchmark for egocentric hand gesture recognition},
  author={Zhang, Yifan and Cao, Congqi and Cheng, Jian and Lu, Hanqing},
  journal={TMM},
  volume={20},
  number={5},
  pages={1038--1050},
  year={2018},
  publisher={IEEE}
}

@misc{siq2,
  author = {Alex Wilf and Leena Mathur and Sheryl Mathew and Claire Ko and Youssouf Kebe and Paul Pu Liang and Louis-Philippe Morency},
  title = {Social-IQ 2.0 Challenge: Benchmarking Multimodal Social Understanding},
  year = {2023},
  publisher = {GitHub},
  journal = {GitHub repository},
  howpublished = {\url{https://github.com/abwilf/Social-IQ-2.0-Challenge}},
}

@inproceedings{lei2018tvqa,
  title={TVQA: Localized, Compositional Video Question Answering},
  author={Lei, Jie and Yu, Licheng and Bansal, Mohit and Berg, Tamara L},
  booktitle={EMNLP},
  year={2018}
}

@article{tomasello1986joint,
  title={Joint attention and early language},
  author={Tomasello, Michael and Farrar, Michael Jeffrey},
  journal={Child development},
  pages={1454--1463},
  year={1986},
  publisher={JSTOR}
}

@article{recasens2015they,
  title={Where are they looking?},
  author={Recasens, Adria and Khosla, Aditya and Vondrick, Carl and Torralba, Antonio},
  journal={NeurIPS},
  volume={28},
  year={2015}
}

@inproceedings{chong2020detecting,
  title={Detecting attended visual targets in video},
  author={Chong, Eunji and Wang, Yongxin and Ruiz, Nataniel and Rehg, James M},
  booktitle={CVPR},
  pages={5396--5406},
  year={2020}
}

@inproceedings{fang2021dual,
  title={Dual attention guided gaze target detection in the wild},
  author={Fang, Yi and Tang, Jiapeng and Shen, Wang and Shen, Wei and Gu, Xiao and Song, Li and Zhai, Guangtao},
  booktitle={CVPR},
  pages={11390--11399},
  year={2021}
}

@article{jin2022depth,
  title={Depth-aware gaze-following via auxiliary networks for robotics},
  author={Jin, Tianlei and Yu, Qizhi and Zhu, Shiqiang and Lin, Zheyuan and Ren, Jie and Zhou, Yuanhai and Song, Wei},
  journal={Engineering Applications of Artificial Intelligence},
  volume={113},
  pages={104924},
  year={2022},
  publisher={Elsevier}
}

@inproceedings{bao2022escnet,
  title={Escnet: Gaze target detection with the understanding of 3d scenes},
  author={Bao, Jun and Liu, Buyu and Yu, Jun},
  booktitle={CVPR},
  pages={14126--14135},
  year={2022}
}

@inproceedings{gupta2022modular,
  title={A Modular Multimodal Architecture for Gaze Target Prediction: Application to Privacy-Sensitive Settings},
  author={Gupta, Anshul and Tafasca, Samy and Odobez, Jean-Marc},
  booktitle={CVPRW},
  pages={5041--5050},
  year={2022}
}

@inproceedings{fan2018inferring,
  title={Inferring shared attention in social scene videos},
  author={Fan, Lifeng and Chen, Yixin and Wei, Ping and Wang, Wenguan and Zhu, Song-Chun},
  booktitle={CVPR},
  pages={6460--6468},
  year={2018}
}

@inproceedings{marin2011here,
  title={" Here's looking at you, kid": Detecting people looking at each other in videos},
  author={Marin-Jimenez, Manuel and Zisserman, Andrew and Ferrari, Vittorio},
  booktitle={BMVC},
  year={2011},
  organization={British Machine Vision Association and Society for Pattern Recognition}
}

@inproceedings{li2018eye,
  title={In the eye of beholder: Joint learning of gaze and actions in first person video},
  author={Li, Yin and Liu, Miao and Rehg, James M},
  booktitle={ECCV},
  pages={619--635},
  year={2018}
}

@article{li2025mimeqa,
  title={MimeQA: Towards Socially-Intelligent Nonverbal Foundation Models},
  author={Li, Hengzhi and Tjandrasuwita, Megan and Fung, Yi R and Solar-Lezama, Armando and Liang, Paul Pu},
  journal={arXiv preprint arXiv:2502.16671},
  year={2025}
}

@article{hall2019nonverbal,
  title={Nonverbal communication},
  author={Hall, Judith A and Horgan, Terrence G and Murphy, Nora A},
  journal={Annual review of psychology},
  volume={70},
  number={2019},
  pages={271--294},
  year={2019},
  publisher={Annual Reviews}
}

@article{gagnon2021reasoning,
  title={Reasoning strategies explain individual differences in social reasoning.},
  author={Gagnon-St-Pierre, {\'E}milie and Doucerain, Marina M and Markovits, Henry},
  journal={Journal of Experimental Psychology: General},
  volume={150},
  number={2},
  pages={340},
  year={2021},
  publisher={American Psychological Association}
}

@article{lee2024towards,
  title={Towards social ai: A survey on understanding social interactions},
  author={Lee, Sangmin and Li, Minzhi and Lai, Bolin and Jia, Wenqi and Ryan, Fiona and Cao, Xu and Kara, Ozgur and Boote, Bikram and Shi, Weiyan and Yang, Diyi and others},
  journal={arXiv preprint arXiv:2409.15316},
  year={2024}
}

@inproceedings{wei2024nonverbal,
  title={Nonverbal interaction detection},
  author={Wei, Jianan and Zhou, Tianfei and Yang, Yi and Wang, Wenguan},
  booktitle={European Conference on Computer Vision},
  pages={277--295},
  year={2024},
  organization={Springer}
}

@inproceedings{mathur2025social,
  title={Social genome: Grounded social reasoning abilities of multimodal models},
  author={Mathur, Leena and Qian, Marian and Liang, Paul Pu and Morency, Louis-Philippe},
  booktitle={Proceedings of the 2025 Conference on Empirical Methods in Natural Language Processing},
  pages={24879--24902},
  year={2025}
}

@inproceedings{lee2024modeling,
  title={Modeling multimodal social interactions: new challenges and baselines with densely aligned representations},
  author={Lee, Sangmin and Lai, Bolin and Ryan, Fiona and Boote, Bikram and Rehg, James M},
  booktitle={Proceedings of the IEEE/CVF Conference on Computer Vision and Pattern Recognition},
  pages={14585--14595},
  year={2024}
}

@inproceedings{cao2025socialgesture,
  title={Socialgesture: Delving into multi-person gesture understanding},
  author={Cao, Xu and Virupaksha, Pranav and Jia, Wenqi and Lai, Bolin and Ryan, Fiona and Lee, Sangmin and Rehg, James M},
  booktitle={Proceedings of the Computer Vision and Pattern Recognition Conference},
  pages={19509--19519},
  year={2025}
}

@inproceedings{fan2019understanding,
  title={Understanding human gaze communication by spatio-temporal graph reasoning},
  author={Fan, Lifeng and Wang, Wenguan and Huang, Siyuan and Tang, Xinyu and Zhu, Song-Chun},
  booktitle={Proceedings of the IEEE/CVF International Conference on Computer Vision},
  pages={5724--5733},
  year={2019}
}

@article{li2025towards,
  title={Towards online multi-modal social interaction understanding},
  author={Li, Xinpeng and Deng, Shijian and Lai, Bolin and Pian, Weiguo and Rehg, James M and Tian, Yapeng},
  journal={arXiv preprint arXiv:2503.19851},
  year={2025}
}

@article{ouyang2025multi,
  title={Multi-speaker Attention Alignment for Multimodal Social Interaction},
  author={Ouyang, Liangyang and Huang, Yifei and Zhang, Mingfang and Kang, Caixin and Furuta, Ryosuke and Sato, Yoichi},
  journal={arXiv preprint arXiv:2511.17952},
  year={2025}
}

@article{niu2025r,
  title={R\^{} 3-VQA:" Read the Room" by Video Social Reasoning},
  author={Niu, Lixing and Li, Jiapeng and Yu, Xingping and Wang, Shu and Feng, Ruining and Wu, Bo and Wei, Ping and Wang, Yisen and Fan, Lifeng},
  journal={arXiv preprint arXiv:2505.04147},
  year={2025}
}

@article{kang2025can,
  title={Can MLLMs Read the Room? A Multimodal Benchmark for Assessing Deception in Multi-Party Social Interactions},
  author={Kang, Caixin and Huang, Yifei and Ouyang, Liangyang and Zhang, Mingfang and Liu, Ruicong and Sato, Yoichi},
  journal={arXiv preprint arXiv:2511.16221},
  year={2025}
}

@article{kong2025siv,
  title={Siv-bench: A video benchmark for social interaction understanding and reasoning},
  author={Kong, Fanqi and Zu, Weiqin and Chen, Xinyu and Yang, Yaodong and Zhu, Song-Chun and Feng, Xue},
  journal={arXiv preprint arXiv:2506.05425},
  year={2025}
}

@article{mathew2025gazevlm,
  title={GazeVLM: A Vision-Language Model for Multi-Task Gaze Understanding},
  author={Mathew, Athul M and Hermassi, Haithem and Khalid, Thariq and Khan, Arshad Ali},
  journal={arXiv preprint arXiv:2511.06348},
  year={2025}
}

@article{pani2025gaze,
  title={Gaze-VLM: Bridging Gaze and VLMs through Attention Regularization for Egocentric Understanding},
  author={Pani, Anupam and Yang, Yanchao},
  journal={arXiv preprint arXiv:2510.21356},
  year={2025}
}

@article{shao2024deepseekmath,
  title={Deepseekmath: Pushing the limits of mathematical reasoning in open language models},
  author={Shao, Zhihong and Wang, Peiyi and Zhu, Qihao and Xu, Runxin and Song, Junxiao and Bi, Xiao and Zhang, Haowei and Zhang, Mingchuan and Li, YK and Wu, Y and others},
  journal={arXiv preprint arXiv:2402.03300},
  year={2024}
}

@article{liu2025understanding,
  title={Understanding r1-zero-like training: A critical perspective},
  author={Liu, Zichen and Chen, Changyu and Li, Wenjun and Qi, Penghui and Pang, Tianyu and Du, Chao and Lee, Wee Sun and Lin, Min},
  journal={arXiv preprint arXiv:2503.20783},
  year={2025}
}

@inproceedings{
liu2023visual,
title={Visual Instruction Tuning},
author={Haotian Liu and Chunyuan Li and Qingyang Wu and Yong Jae Lee},
booktitle={Advances in Neural Information Processing Systems},
year={2023}
}

@article{liu2023improved,
  title={Improved baselines with visual instruction tuning},
  author={Liu, Haotian and Li, Chunyuan and Li, Yuheng and Lee, Yong Jae},
  journal={arXiv preprint arXiv:2310.03744},
  year={2023}
}

@inproceedings{
dai2023instructblip,
title={Instruct{BLIP}: Towards General-purpose Vision-Language Models with Instruction Tuning},
author={Wenliang Dai and Junnan Li and Dongxu Li and Anthony Tiong and Junqi Zhao and Weisheng Wang and Boyang Li and Pascale Fung and Steven Hoi},
booktitle={Advances in Neural Information Processing Systems},
year={2023},
}

@inproceedings{lei2020tvqa+,
  title={Tvqa+: Spatio-temporal grounding for video question answering},
  author={Lei, Jie and Yu, Licheng and Berg, Tamara and Bansal, Mohit},
  booktitle={Proceedings of the 58th annual meeting of the association for computational linguistics},
  pages={8211--8225},
  year={2020}
}

@inproceedings{zadeh2019social,
  title={Social-iq: A question answering benchmark for artificial social intelligence},
  author={Zadeh, Amir and Chan, Michael and Liang, Paul Pu and Tong, Edmund and Morency, Louis-Philippe},
  booktitle={Proceedings of the IEEE/CVF Conference on Computer Vision and Pattern Recognition},
  pages={8807--8817},
  year={2019}
}

@article{nguyen2025see,
  title={See, Hear, and Understand: Benchmarking Audiovisual Human Speech Understanding in Multimodal Large Language Models},
  author={Nguyen, Le Thien Phuc and Yu, Zhuoran and Hang, Samuel Low Yu and An, Subin and Lee, Jeongik and Ban, Yohan and Chung, SeungEun and Nguyen, Thanh-Huy and Maeng, JuWan and Lee, Soochahn and others},
  journal={arXiv preprint arXiv:2512.02231},
  year={2025}
}

@inproceedings{wang2025friends,
  title={Friends-mmc: A dataset for multi-modal multi-party conversation understanding},
  author={Wang, Yueqian and Meng, Xiaojun and Wang, Yuxuan and Liang, Jianxin and Liu, Qun and Zhao, Dongyan},
  booktitle={Proceedings of the AAAI Conference on Artificial Intelligence},
  volume={39},
  number={24},
  pages={25425--25433},
  year={2025}
}

@article{mclean2025embody,
  title={Embody 3d: A large-scale multimodal motion and behavior dataset},
  author={McLean, Claire and Meendering, Makenzie and Swartz, Tristan and Gabbay, Orri and Olsen, Alexandra and Jacobs, Rachel and Rosen, Nicholas and de Bree, Philippe and Garcia, Tony and Merrill, Gadsden and others},
  journal={arXiv preprint arXiv:2510.16258},
  year={2025}
}

@inproceedings{lai2023werewolf,
  title={Werewolf among us: Multimodal resources for modeling persuasion behaviors in social deduction games},
  author={Lai, Bolin and Zhang, Hongxin and Liu, Miao and Pariani, Aryan and Ryan, Fiona and Jia, Wenqi and Hayati, Shirley Anugrah and Rehg, James and Yang, Diyi},
  booktitle={Findings of ACL},
  pages={6570--6588},
  year={2023}
}

@inproceedings{cao2025toward,
  title={Toward Human Deictic Gesture Target Estimation},
  author={Cao, Xu and Virupaksha, Pranav and Lee, Sangmin and Lai, Bolin and Jia, Wenqi and Chen, Jintai and Rehg, James Matthew},
  booktitle={The Thirty-ninth Annual Conference on Neural Information Processing Systems},
  year={2025}
}

@article{carion2025sam,
  title={Sam 3: Segment anything with concepts},
  author={Carion, Nicolas and Gustafson, Laura and Hu, Yuan-Ting and Debnath, Shoubhik and Hu, Ronghang and Suris, Didac and Ryali, Chaitanya and Alwala, Kalyan Vasudev and Khedr, Haitham and Huang, Andrew and others},
  journal={arXiv preprint arXiv:2511.16719},
  year={2025}
}

@inproceedings{deng2020retinaface,
  title={Retinaface: Single-shot multi-level face localisation in the wild},
  author={Deng, Jiankang and Guo, Jia and Ververas, Evangelos and Kotsia, Irene and Zafeiriou, Stefanos},
  booktitle={Proceedings of the IEEE/CVF conference on computer vision and pattern recognition},
  pages={5203--5212},
  year={2020}
}

@inproceedings{cao2026gaze,
  title={Gaze Target Estimation Anywhere with Concepts},
  author={Cao, Xu and Yang, Houze and Gunda, Vipin and Zhou, Zhongyi and Xu, Tianyu and Kowdle, Adarsh and Kim, Inki and Rehg, James M},
  booktitle={Proceedings of the IEEE/CVF Conference on Computer Vision and Pattern Recognition},
  year={2026}
}

@article{chen2026think,
  title={Think Deep, Not Just Long: Measuring LLM Reasoning Effort via Deep-Thinking Tokens},
  author={Chen, Wei-Lin and Peng, Liqian and Tan, Tian and Zhao, Chao and Chen, Blake JianHang and Lin, Ziqian and Go, Alec and Meng, Yu},
  journal={arXiv preprint arXiv:2602.13517},
  year={2026}
}

@article{wu2025more,
  title={When more is less: Understanding chain-of-thought length in llms},
  author={Wu, Yuyang and Wang, Yifei and Ye, Ziyu and Du, Tianqi and Jegelka, Stefanie and Wang, Yisen},
  journal={arXiv preprint arXiv:2502.07266},
  year={2025}
}

@article{su2025between,
  title={Between underthinking and overthinking: An empirical study of reasoning length and correctness in llms},
  author={Su, Jinyan and Healey, Jennifer and Nakov, Preslav and Cardie, Claire},
  journal={arXiv preprint arXiv:2505.00127},
  year={2025}
}

@article{jording2018social,
  title={The “social gaze space”: A taxonomy for gaze-based communication in triadic interactions},
  author={Jording, Mathis and Hartz, Arne and Bente, Gary and Schulte-R{\"u}ther, Martin and Vogeley, Kai},
  journal={Frontiers in psychology},
  volume={9},
  pages={226},
  year={2018},
  publisher={Frontiers Media SA}
}

@book{moore2014joint,
  title={Joint attention: Its origins and role in development},
  author={Moore, Chris and Dunham, Philip J and Dunham, Phil},
  year={2014},
  publisher={Psychology Press}
}

@article{kuhn1955hungarian,
  title={The Hungarian method for the assignment problem},
  author={Kuhn, Harold W},
  journal={Naval research logistics quarterly},
  volume={2},
  number={1-2},
  pages={83--97},
  year={1955},
  publisher={Wiley Online Library}
}

@article{yan2025videochat,
  title={Videochat-r1. 5: Visual test-time scaling to reinforce multimodal reasoning by iterative perception},
  author={Yan, Ziang and Li, Xinhao and He, Yinan and Yue, Zhengrong and Zeng, Xiangyu and Wang, Yali and Qiao, Yu and Wang, Limin and Wang, Yi},
  journal={arXiv preprint arXiv:2509.21100},
  year={2025}
}

@article{chen2025scaling,
  title={Scaling rl to long videos},
  author={Chen, Yukang and Huang, Wei and Shi, Baifeng and Hu, Qinghao and Ye, Hanrong and Zhu, Ligeng and Liu, Zhijian and Molchanov, Pavlo and Kautz, Jan and Qi, Xiaojuan and others},
  journal={arXiv preprint arXiv:2507.07966},
  year={2025}
}

@article{feng2025video,
  title={Video-r1: Reinforcing video reasoning in mllms},
  author={Feng, Kaituo and Gong, Kaixiong and Li, Bohao and Guo, Zonghao and Wang, Yibing and Peng, Tianshuo and Wu, Junfei and Zhang, Xiaoying and Wang, Benyou and Yue, Xiangyu},
  journal={arXiv preprint arXiv:2503.21776},
  year={2025}
}

@article{wang2025vl,
  title={Vl-rethinker: Incentivizing self-reflection of vision-language models with reinforcement learning},
  author={Wang, Haozhe and Qu, Chao and Huang, Zuming and Chu, Wei and Lin, Fangzhen and Chen, Wenhu},
  journal={arXiv preprint arXiv:2504.08837},
  year={2025}
}

@article{park2025dip,
  title={Dip-r1: Deep inspection and perception with rl looking through and understanding complex scenes},
  author={Park, Sungjune and Kim, Hyunjun and Kim, Junho and Kim, Seongho and Ro, Yong Man},
  journal={arXiv preprint arXiv:2505.23179},
  year={2025}
}

@inproceedings{bossen2025can,
  title={Can vision-language models understand and interpret dynamic gestures from pedestrians? pilot datasets and exploration towards instructive nonverbal commands for cooperative autonomous vehicles},
  author={Bossen, Tonko EW and M{\o}gelmose, Andreas and Greer, Ross},
  booktitle={Proceedings of the Computer Vision and Pattern Recognition Conference},
  pages={4779--4788},
  year={2025}
}

@article{wang2025internvl3,
  title={Internvl3. 5: Advancing open-source multimodal models in versatility, reasoning, and efficiency},
  author={Wang, Weiyun and Gao, Zhangwei and Gu, Lixin and Pu, Hengjun and Cui, Long and Wei, Xingguang and Liu, Zhaoyang and Jing, Linglin and Ye, Shenglong and Shao, Jie and others},
  journal={arXiv preprint arXiv:2508.18265},
  year={2025}
}

@article{an2025llava,
  title={Llava-onevision-1.5: Fully open framework for democratized multimodal training},
  author={An, Xiang and Xie, Yin and Yang, Kaicheng and Zhang, Wenkang and Zhao, Xiuwei and Cheng, Zheng and Wang, Yirui and Xu, Songcen and Chen, Changrui and Zhu, Didi and others},
  journal={arXiv preprint arXiv:2509.23661},
  year={2025}
}

@article{bai2025qwen2,
  title={Qwen2. 5-vl technical report},
  author={Bai, Shuai and Chen, Keqin and Liu, Xuejing and Wang, Jialin and Ge, Wenbin and Song, Sibo and Dang, Kai and Wang, Peng and Wang, Shijie and Tang, Jun and others},
  journal={arXiv preprint arXiv:2502.13923},
  year={2025}
}

@article{team2026qwen3,
  title={Qwen3. 5-Omni Technical Report},
  author={Team, Qwen},
  journal={arXiv preprint arXiv:2604.15804},
  year={2026}
}

@misc{gpt54,
  title        = {GPT-5.4 Thinking System Card},
  author       = {{OpenAI}},
  year         = {2026},
  month        = {mar},
  howpublished = {\url{https://openai.com/index/gpt-5-4-thinking-system-card/}},
  note         = {Official system card}
}

@misc{claudesonnet46,
  title        = {System Card: Claude Sonnet 4.6},
  author       = {{Anthropic}},
  year         = {2026},
  month        = {feb},
  howpublished = {\url{https://www.anthropic.com/claude-haiku-4-5-system-card}},
  note         = {Official system card}
}

@misc{gemini31pro,
  title        = {Gemini 3.1 Pro Model Card},
  author       = {{Google Deepmind}},
  year         = {2026},
  month        = {feb},
  howpublished = {\url{https://deepmind.google/models/model-cards/gemini-3-1-pro/}},
  note         = {Official system card}
}

@article{gupta2024mtgs,
  title={Mtgs: A novel framework for multi-person temporal gaze following and social gaze prediction},
  author={Gupta, Anshul and Tafasca, Samy and Farkhondeh, Arya and Vuillecard, Pierre and Odobez, Jean-marc},
  journal={Advances in Neural Information Processing Systems},
  volume={37},
  pages={15646--15673},
  year={2024}
}

@inproceedings{gupta2024exploring,
  title={Exploring the zero-shot capabilities of vision-language models for improving gaze following},
  author={Gupta, Anshul and Vuillecard, Pierre and Farkhondeh, Arya and Odobez, Jean-Marc},
  booktitle={Proceedings of the IEEE/CVF Conference on Computer Vision and Pattern Recognition},
  pages={615--624},
  year={2024}
}

@article{ryan2025gaze,
  title={Gaze-LLE: Gaze Target Estimation via Large-Scale Learned Encoders},
  author={Ryan, Fiona and Bati, Ajay and Lee, Sangmin and Bolya, Daniel and Hoffman, Judy and Rehg, James M},
  booktitle={Proceedings of the IEEE/CVF Conference on Computer Vision and Pattern Recognition},
  year={2025}
}

@article{song2024vitgaze,
  title={ViTGaze: gaze following with interaction features in vision transformers},
  author={Song, Yuehao and Wang, Xinggang and Yao, Jingfeng and Liu, Wenyu and Zhang, Jinglin and Xu, Xiangmin},
  journal={Visual Intelligence},
  volume={2},
  number={1},
  pages={1--15},
  year={2024},
  publisher={Springer}
}

@inproceedings{fu2025video,
  title={Video-mme: The first-ever comprehensive evaluation benchmark of multi-modal llms in video analysis},
  author={Fu, Chaoyou and Dai, Yuhan and Luo, Yongdong and Li, Lei and Ren, Shuhuai and Zhang, Renrui and Wang, Zihan and Zhou, Chenyu and Shen, Yunhang and Zhang, Mengdan and others},
  booktitle={Proceedings of the IEEE/CVF conference on computer vision and pattern recognition},
  pages={24108--24118},
  year={2025}
}

@article{fu2026video,
  title={Video-MME-v2: Towards the Next Stage in Benchmarks for Comprehensive Video Understanding},
  author={Fu, Chaoyou and Yuan, Haozhi and Dong, Yuhao and Zhang, Yi-Fan and Shen, Yunhang and Hu, Xiaoxing and Li, Xueying and Su, Jinsen and Long, Chengwu and Xie, Xiaoyao and others},
  journal={arXiv preprint arXiv:2604.05015},
  year={2026}
}

@article{hu2025video,
  title={Video-mmmu: Evaluating knowledge acquisition from multi-discipline professional videos},
  author={Hu, Kairui and Wu, Penghao and Pu, Fanyi and Xiao, Wang and Zhang, Yuanhan and Yue, Xiang and Li, Bo and Liu, Ziwei},
  journal={arXiv preprint arXiv:2501.13826},
  year={2025}
}

@article{bai2025qwen3,
  title={Qwen3-vl technical report},
  author={Bai, Shuai and Cai, Yuxuan and Chen, Ruizhe and Chen, Keqin and Chen, Xionghui and Cheng, Zesen and Deng, Lianghao and Ding, Wei and Gao, Chang and Ge, Chunjiang and others},
  journal={arXiv preprint arXiv:2511.21631},
  year={2025}
}

@article{li2026omni,
  title={Omni-MMSI: Toward Identity-attributed Social Interaction Understanding},
  author={Li, Xinpeng and Lai, Bolin and Chen, Hardy and Deng, Shijian and Xie, Cihang and Zhou, Yuyin and Rehg, James Matthew and Tian, Yapeng},
  journal={arXiv preprint arXiv:2604.00267},
  year={2026}
}

@article{tahboub2025socialfusion,
  title={SocialFusion: Addressing Social Degradation in Pre-trained Vision-Language Models},
  author={Tahboub, Hamza and Shi, Weiyan and Hua, Gang and Jiang, Huaizu},
  journal={arXiv preprint arXiv:2512.01148},
  year={2025}
}

@article{xie2026socialomni,
  title={SocialOmni: Benchmarking Audio-Visual Social Interactivity in Omni Models},
  author={Xie, Tianyu and Huang, Jinfa and Ma, Yuexiao and Luo, Rongfang and Yang, Yan and Chen, Wang and Zeng, Yuhui and Fang, Ruize and Zou, Yixuan and Zheng, Xiawu and others},
  journal={arXiv preprint arXiv:2603.16859},
  year={2026}
}

@article{thumu2026social,
  title={Social Caption: Evaluating Social Understanding in Multimodal Models},
  author={Thumu, Bhaavanaa and Mathur, Leena and Kebe, Youssouf and Morency, Louis-Philippe},
  journal={arXiv preprint arXiv:2601.14569},
  year={2026}
}

@article{zhang2025videollama,
  title={Videollama 3: Frontier multimodal foundation models for image and video understanding},
  author={Zhang, Boqiang and Li, Kehan and Cheng, Zesen and Hu, Zhiqiang and Yuan, Yuqian and Chen, Guanzheng and Leng, Sicong and Jiang, Yuming and Zhang, Hang and Li, Xin and others},
  journal={arXiv preprint arXiv:2501.13106},
  year={2025}
}

@article{zhao2025humanomni,
  title={Humanomni: A large vision-speech language model for human-centric video understanding},
  author={Zhao, Jiaxing and Yang, Qize and Peng, Yixing and Bai, Detao and Yao, Shimin and Sun, Boyuan and Chen, Xiang and Fu, Shenghao and Wei, Xihan and Bo, Liefeng and others},
  journal={arXiv preprint arXiv:2501.15111},
  year={2025}
}

@inproceedings{
niu2026read,
title={Read the Room: Video Social Reasoning with Mental-Physical Causal Chains},
author={Lixing Niu and Jiapeng Li and Xingping Yu and Xinyi Dong and Shu Wang and Ruining Feng and Bo Wu and Ping Wei and Yisen Wang and Lifeng Fan},
booktitle={The Fourteenth International Conference on Learning Representations},
year={2026},
url={https://openreview.net/forum?id=TJilJnZjpw}
}

@misc{
wang2025gaze,
title={Gaze Following in Question Answering: A Comprehensive Benchmark for Vision-Language Models},
author={Shijing Wang and Chaoqun Cui and Yihua Cheng and Yaping Huang},
year={2025},
url={https://openreview.net/forum?id=UyiTjp0oKU}
}

@inproceedings{lin2025v,
  title={V-ALPHASOCIAL: Benchmark and Self-Reflective Chain-of-Thought Generation for Visual Social Commonsense Reasoning},
  author={Lin, Zongyu and Xu, Zhikun and Song, Xiaohan and Wan, Yixin and Yao, Xingcheng and Lin, Tsung-Han and Song, Selina and Subbaraman, Pranav and Zhou, Ben and Chang, Kai-Wei and others},
  booktitle={Findings of the Association for Computational Linguistics: ACL 2025},
  pages={19025--19047},
  year={2025}
}

@article{li2025gestura,
  title={Gestura: A LVLM-Powered System Bridging Motion and Semantics for Real-Time Free-Form Gesture Understanding},
  author={Li, Zhuoming and Liu, Aitong and Jia, Mengxi and Lu, Yubo and Zhang, Tengxiang and Sun, Changzhi and Zhang, Dell and Li, Xuelong},
  journal={Proceedings of the ACM on Interactive, Mobile, Wearable and Ubiquitous Technologies},
  volume={9},
  number={4},
  pages={1--29},
  year={2025},
  publisher={ACM New York, NY, USA}
}

@article{kim2024salova,
  title={SALOVA: Segment-Augmented Long Video Assistant for Targeted Retrieval and Routing in Long-Form Video Analysis},
  author={Kim, Junho and Kim, Hyunjun and Lee, Hosu and Ro, Yong Man},
  journal={arXiv preprint arXiv:2411.16173},
  year={2024}
}
\bibliographystyle{ieee_fullname}
}

\clearpage
\newpage
\appendix

\clearpage
\section*{Appendix Contents}

\startcontents[appendices]
\printcontents[appendices]{}{1}{}

\newpage
\section{Detailed \textit{GRASP} Construction}
\label{appendix:data_detail}
\subsection{Video Pre-processing and Person-Consistent Gaze Trajectories}
To construct temporally grounded social interactions, we first normalize raw videos into short interaction-centered clips. Long-form videos are segmented using scene-cut detection (PySceneDetect\footnote{We use AdaptiveDetector with default setup in \url{https://github.com/Breakthrough/PySceneDetect}.}). We enforce a minimum segment length of 45 seconds by merging overly short adjacent segments, and a maximum segment length of 120 seconds by splitting long segments at the nearest available scene boundary. All downstream gaze processing samples videos at 2 fps, so the temporal resolution of gaze-derived annotations is 0.5 seconds.

For each processed clip, we estimate person-consistent gaze trajectories through a multi-stage identity-aware pipeline. We first detect and track people using SAM 3~\cite{carion2025sam} across frames using a text-prompted person tracking model with the prompt ``people'', yielding temporally consistent person bounding boxes and participant identities. Since these person tracks do not provide sufficiently precise facial localization for gaze estimation, we additionally detect faces using~\cite{deng2020retinaface} in each sampled frame. Each detected face is associated with a tracked person identity by solving a Hungarian matching problem~\cite{kuhn1955hungarian} between face boxes and the upper half of the corresponding person boxes, which approximates the head region. Matches with zero overlap are rejected to avoid assigning a face to an unrelated body track.

After face-person association, we estimate the gaze target with AnyGaze~\cite{cao2026gaze} for each matched face. The gaze estimator predicts a normalized gaze point \(g_i^t=(x_i^t,y_i^t)\in[0,1]^2\) for person \(i\) at time \(t\), together with an in-frame indicator. The final per-frame representation therefore aligns each participant identity with a person box, face box, gaze point, and gaze validity flag. This identity-preserving representation is the basis for all subsequent gaze feature computation and social event detection.

\begin{table*}[ht!]
\centering
\small
\setlength{\tabcolsep}{4.5pt}
\caption{Key components used to construct person-consistent gaze trajectories.}
\vspace{-2mm}
\label{tab:grasp_gaze_components}
\resizebox{1.0\linewidth}{!}{
\begin{tabular}{@{}p{0.2\linewidth} p{0.37\linewidth} p{0.38\linewidth}@{}}
\Xhline{2\arrayrulewidth}
\rowcolor{headergray}
\textbf{Component} & \textbf{Role} & \textbf{Key setting} \\
\hline
Scene segmentation
& Split long-form videos into coherent interaction segments.
& 45--120s segments \\

Frame sampling
& Normalize the temporal resolution for downstream gaze processing.
& 2 fps; 0.5s interval \\

Person tracking
& Maintain stable participant identities across frames.
& Text prompt ``people''; confidence threshold 0.45 \\

Face detection
& Localize faces for gaze estimation.
& Per sampled frame; confidence threshold 0.54 \\

Face-person matching
& Assign each detected face to a tracked participant identity.
& Hungarian matching on the upper 50\% person region; zero-overlap matches rejected \\

Gaze estimation
& Predict where each participant is looking.
& Normalized frame coordinate \(g_i^t\in[0,1]^2\) with in-frame flag \\
\Xhline{2\arrayrulewidth}
\end{tabular}
}
\vspace{-1mm}
\end{table*}

\subsection{Gaze Feature Extraction}
Raw gaze trajectories contain missing observations caused by missed face detections, out-of-frame gaze, occlusion, and abrupt camera changes. We therefore apply a conservative interpolation scheme before computing temporal features. Missing gaps of up to 3 frames are linearly interpolated, as shown in Tab.~\ref{tab:gaze_interpolation} . Gaps of 4--10 frames are filled by carrying forward the last valid observation with decayed confidence. Longer gaps are left missing. To prevent spurious continuity across cuts or large viewpoint changes, interpolation is blocked whenever the temporal gap exceeds 3 seconds or the face displacement exceeds 30\% of the frame width.

To reduce sensitivity to camera motion, gaze velocity is computed relative to the person's face center rather than from raw gaze point displacement. Let \(c_i^t\) denote the face center of person \(i\) at time \(t\). We define the face-centered gaze direction as
\[
d_i^t = g_i^t - c_i^t,
\]
and estimate gaze velocity as:
\[
v_i^t = \frac{\lVert d_i^t - d_i^{t-1} \rVert}{\Delta t}.
\]
This formulation captures changes in where a person looks relative to their own head position, making it more robust to camera pans and global frame motion.

We also compute a group-level convergence score to identify shared attention. For all valid in-frame gaze points at time \(t\), we compute their centroid \(\bar{g}^t\) and measure the median distance from individual gaze points to this centroid. The convergence score is
\[
s^t = \exp\left(-3.0 \cdot \mathrm{median}_i \lVert g_i^t - \bar{g}^t \rVert \right).
\]
Using the median distance makes the score robust to outliers, such as one participant looking away while several others attend to a common region.

\begin{table*}[ht!]
\centering
\small
\setlength{\tabcolsep}{5.0pt}
\caption{Missing-gaze handling before temporal feature computation.}
\vspace{-2mm}
\label{tab:gaze_interpolation}
\resizebox{0.6\linewidth}{!}{
\begin{tabular}{@{}l l l@{}}
\Xhline{2\arrayrulewidth}
\rowcolor{headergray}
\textbf{Missing gap} & \textbf{Strategy} & \textbf{Assigned confidence} \\
\hline
\(\leq 3\) frames
& Linear interpolation
& \(1.0 - 0.1 \times \mathrm{gap}\) \\

4--10 frames
& Carry forward
& \(0.5 \times \exp(-0.2 \times \mathrm{gap})\) \\
\(>10\) frames

& Keep missing
& 0 \\
\Xhline{2\arrayrulewidth}
\end{tabular}
}
\vspace{-1mm}
\end{table*}

\subsection{Social Gaze Event Detection}
From the per-person velocities and group convergence features, we detect five classes of social gaze events: sudden gaze shift, joint attention, gaze following, attention capture, and mutual gaze, as summarized in Tab.~\ref{tab:gaze_event_definitions}. Each event type is defined through an operational detector over identity-consistent gaze trajectories.

\begin{table*}[ht!]
\centering
\small
\setlength{\tabcolsep}{4.2pt}
\caption{Operational definitions for the five social gaze event types used in \textit{GRASP}.}
\vspace{-2mm}
\label{tab:gaze_event_definitions}
\resizebox{1.0\linewidth}{!}{
\begin{tabular}{@{}p{0.2\linewidth} p{0.48\linewidth} p{0.3\linewidth}@{}}
\Xhline{2\arrayrulewidth}
\rowcolor{headergray}
\textbf{Event type} & \textbf{Event definition} & \textbf{Main criterion} \\
\hline
Sudden gaze shift
& One participant rapidly changes gaze direction.
& \(v_i^t > 0.7\) \\

Joint attention
& At least two participants attend to a common region.
& \(s^t \geq 0.6\), duration \(\geq 0.5\)s \\

Gaze following
& One participant looks toward another participant's previous gaze target.
& distance \(<0.03\), lag 1.0--2.0s \\

Attention capture
& At least three participants shift gaze within a short temporal window.
& \(v_i^t > 0.4\) for \(\geq 3\) persons \\

Mutual gaze
& Two participants look at each other's face regions.
& bidirectional face-box hit, duration \(\geq 1.0\)s \\
\Xhline{2\arrayrulewidth}
\end{tabular}
}
\vspace{-1mm}
\end{table*}

For sudden gaze shifts, we cluster consecutive high-velocity frames using a maximum temporal gap of 0.6 seconds and retain events whose durations fall between 0.5 and 1.5 seconds. Joint attention is detected by identifying frames with high convergence score and at least two visible faces, then clustering consecutive frames whose participant sets have at least 70\% overlap. Participants far from the convergence center are filtered out to avoid including late joiners or peripheral observers.

Gaze following is detected by comparing a participant's current gaze target against another participant's prior high-confidence measured gaze target. We only use measured gaze histories for this detector and require a temporal lag between 1.0 and 2.0 seconds. Attention capture is detected when at least three visible participants exhibit high gaze velocity within a short temporal window, indicating a simultaneous shift of attention. Mutual gaze is detected through a bidirectional hit test: person \(A\)'s gaze must fall inside person \(B\)'s face box while person \(B\)'s gaze falls inside person \(A\)'s face box. To reduce false positives, mutual gaze uses measured gaze only, not interpolated gaze, and applies a 2\% bounding-box margin. Only high-confidence gaze events are passed to the unified graph construction stage. In the final construction pipeline, downstream processing keeps gaze events with confidence at least 0.9.

\subsection{Deictic Gesture Annotation}
In addition to gaze, \textit{GRASP} models communicative hand gestures that direct attention or indicate interaction targets. We use a \texttt{Gemini 3.1 Pro}~\cite{gemini31pro} to detect four deictic gesture types: pointing, showing, giving, and reaching (Tab.~\ref{tab:gesture_definitions}). To provide explicit identity and temporal grounding, each input video is rendered with tracked person boxes, visible participant labels \(P0, P1, \ldots\), and timestamp overlays \(t=\mathrm{X.XXs}\). The model is instructed to scan the entire video, identify all visible deictic gestures, assign initiator and target IDs from the overlay, and use the displayed timestamp for start and end times. The full gesture classification prompt is shown in Fig.~\ref{fig:gesture_prompt}.

We use a strict and conservative gesture definition. A gesture is annotated only when its type and intent are clearly identifiable from hand, finger, or arm motion alone. The prompt explicitly excludes inference from gaze direction, head pose, speech, facial expression, scene context, or gaze visualization artifacts. Ambiguous hand movements, beat gestures, casual arm motion, and resting poses are skipped. We require a minimum duration of 1.0 second, allow simultaneous gestures by different participants, and return an empty set when no high-confidence deictic gesture is visible.

We further apply an iterative verification-and-regeneration step. After the initial Gemini prediction, a separate verification pass scores the predicted gesture set out of 7 points: 2 points for timeline alignment and 5 points for visual-context accuracy, including gesture type, initiator ID, target person ID, reliance on deictic hand/arm motion, and coverage of visible gesture events. Each subcriterion is scored as 1.0, 0.5, or 0.0. If the total score is at most 5/7, we regenerate the gesture prediction by providing the previous prediction and verification report as feedback. This process reduces timestamp errors, swapped participant identities, false positives from non-deictic motion, and missed salient gestures.

Each retained gesture is stored with structured fields including gesture type, initiator identity, target type, target person identity when applicable, start time, end time, and confidence. The final construction keeps gestures with confidence at least 0.85. Free-form target descriptions and reasoning text produced by Gemini are used only as contextual language for graph and QA generation. Ground-truth answers are derived only from structured, verifiable fields such as gesture type, initiator identity, target identity, and temporal boundaries.

\begin{table*}[ht!]
\centering
\small
\setlength{\tabcolsep}{4.6pt}
\caption{Gesture types annotated in \textit{GRASP} and their conservative annotation criteria.}
\vspace{-2mm}
\label{tab:gesture_definitions}
\resizebox{1.0\linewidth}{!}{
\begin{tabular}{@{}p{0.18\linewidth} p{0.72\linewidth}@{}}
\Xhline{2\arrayrulewidth}
\rowcolor{headergray}
\textbf{Gesture type} & \textbf{Annotation criterion} \\
\hline
Pointing
& Clear arm, hand, or finger extension toward a specific person or target. \\

Showing
& Intentionally orienting or presenting an object for another participant to inspect. \\

Giving
& Offering an object to another participant with transfer intent. \\

Reaching
& Directed hand extension toward an object or target with acquisition intent. \\
\Xhline{2\arrayrulewidth}
\end{tabular}
}
\vspace{-1mm}
\end{table*}

\subsection{Unified Social Graph Construction}
After extracting gaze events and gesture annotations, we merge them into a unified social graph for each video. The graph represents social interactions as temporally ordered event nodes. Each node stores the event source, event type, involved participant identities, start and end times, confidence, and modality-specific attributes such as gaze convergence statistics, leader-follower identities, gesture initiators, or gesture targets.

Before graph construction, gaze events and gestures are filtered using their respective confidence thresholds. Gaze events are retained at confidence \(\geq 0.9\), and gestures are retained at confidence \(\geq 0.85\). All timestamps are snapped to 0.5-second intervals to match the 2 fps processing resolution. Events whose temporal boundaries exceed the video duration are removed. Duplicate events are removed when they share the same type, the same participants, and an overlapping temporal window.

To support joint reasoning over coordinated non-verbal cues, we identify gaze--gesture pairs whose temporal distance is at most 3.0 seconds. These pairs become the basis for Joint Reasoning questions in categories J1--J4. During graph pruning, gaze--gesture pairs are protected before single-modality events are removed, because randomly dropping one side of a pair would destroy the evidence needed for joint reasoning. When a video contains many events, we cap the graph at 25 events, prioritizing protected gaze--gesture pairs, event-type diversity, and high-confidence events.

\begin{table*}[ht!]
\centering
\small
\setlength{\tabcolsep}{4.8pt}
\caption{Key filtering and merging parameters for unified social graph construction.}
\vspace{-2mm}
\label{tab:graph_construction_params}
\resizebox{0.92\linewidth}{!}{
\begin{tabular}{@{}p{0.28\linewidth} p{0.12\linewidth} p{0.5\linewidth}@{}}
\Xhline{2\arrayrulewidth}
\rowcolor{headergray}
\textbf{Parameter} & \textbf{Value} & \textbf{Purpose} \\
\hline
Gaze confidence threshold
& \(\geq 0.9\)
& Retain high-precision gaze events. \\

Gesture confidence threshold
& \(\geq 0.85\)
& Retain high-confidence gesture annotations. \\

Timestamp snapping
& 0.5s
& Match the 2 fps processing resolution. \\

Joint-pair proximity
& \(\leq 3.0\)s
& Link temporally adjacent gaze and gesture events. \\

Maximum events per video
& 25
& Control graph size while preserving key social evidence. \\
\Xhline{2\arrayrulewidth}
\end{tabular}
}
\vspace{-1mm}
\end{table*}

\subsection{Social Reasoning QA Generation}

\textit{GRASP} question--answer pairs are generated from the unified social graphs using Gemini 3.1 Pro~\cite{gemini31pro}. The goal is to convert structured gaze, gesture, and joint gaze--gesture events into natural-language social reasoning questions whose answers remain directly verifiable from event metadata. Each QA instance is anchored to one or more source events and includes a time range that identifies the relevant visual evidence. 

We use the 16-category taxonomy defined in \S~3.2 and Tab.~\ref{tab:grasp_taxonomy_appendix}, covering gaze reasoning (T1--T6), gesture reasoning (G1--G6), and joint reasoning (J1--J4). QA generation is adaptive to the event density of each graph. Clips with only a few gaze or gesture events generate simple perception-oriented questions, such as gaze target identification, gaze event classification, gesture recognition, and gesture type classification. Clips with richer event timelines generate harder temporal, reciprocal, frequency, sequence, group-attention, and joint reasoning questions. Joint reasoning questions are generated only when the graph contains temporally linked gaze--gesture pairs. The full QA generation prompt is shown in Fig.~\ref{fig:qa_prompt}.

All generated QA pairs are validated before inclusion. We discard malformed multiple-choice questions, questions with missing or invalid time ranges, answers containing speculative or affective language, invalid person references, and category--modality mismatches. The correct answer is derived from structured graph fields such as event type, participant IDs, initiator and target IDs, timestamps, and gaze--gesture pair links, rather than from free-form generated descriptions. Metadata such as source event IDs, time ranges, and internal reasoning fields is used for verification and filtering.

\newcommand{\darkblue}[1]{\textcolor{blue!60!black}{#1}}

\begin{table*}[tp]
    \centering
    \setlength\tabcolsep{4pt}
    \resizebox{1.0\textwidth}{!}{
    \begin{tabular}{c|l|l}
    \Xhline{1.0pt}
    \textbf{Modality} & \textbf{Subcategory} & \textbf{Example} \\
    \Xhline{1.0pt}
    \multirow{12}{*}{\textbf{Gaze}}
      & T1: Gaze Target ID            & \cellcolor{gray!5}{\textit{\darkblue{Q: At around 2.7 seconds, who is Person 0 looking at?}}} \\
      &                              & \cellcolor{gray!5}{A: Person 2} \\
    \hhline{~|-|-}
      & T2: Gaze Event Classification & \cellcolor{gray!5}{\textit{\darkblue{Q: What best describes the interaction between Person 0 and Person 1 around 2.0 seconds?}}} \\
      &                              & \cellcolor{gray!5}{A: Looking at the same thing} \\
    \hhline{~|-|-}
      & T3: Mutual Gaze Recognition   & \cellcolor{gray!5}{\textit{\darkblue{Q: How long do Person 1 and Person 2 maintain eye contact starting at 1.5 seconds?}}} \\
      &                              & \cellcolor{gray!5}{A: 3.0 seconds} \\
    \hhline{~|-|-}
      & T4: Gaze Following Detection  & \cellcolor{gray!5}{\textit{\darkblue{Q: Which pair of people makes eye contact between 2.5 and 4.5 seconds?}}} \\
      &                              & \cellcolor{gray!5}{A: Person 3 and Person 5} \\
    \hhline{~|-|-}
      & T5: Temporal Gaze Reasoning   & \cellcolor{gray!5}{\textit{\darkblue{Q: In the gaze following event between 1.0s and 3.0s, who looks at the target first?}}} \\
      &                              & \cellcolor{gray!5}{A: Person 6} \\
    \hhline{~|-|-}
      & T6: Group Attention Dynamics  & \cellcolor{gray!5}{\textit{\darkblue{Q: During the shared attention event between 2.0 and 3.0 seconds, which person is NOT part of the group?}}} \\
      &                              & \cellcolor{gray!5}{A: Person 3} \\
    \hline

    \multirow{12}{*}{\textbf{Gesture}}
      & G1: Gesture Recognition       & \cellcolor{gray!5}{\textit{\darkblue{Q: Between 3.0 and 5.0 seconds, who is Person 3 pointing at?}}} \\
      &                              & \cellcolor{gray!5}{A: Person 0} \\
    \hhline{~|-|-}
      & G2: Gesture Type              & \cellcolor{gray!5}{\textit{\darkblue{Q: What type of gesture does Person 1 perform at 3.0 seconds?}}} \\
      &                              & \cellcolor{gray!5}{A: Reaching} \\
    \hhline{~|-|-}
      & G3: Gesture Temporal          & \cellcolor{gray!5}{\textit{\darkblue{Q: Between 2.5 and 5.0 seconds, Person 10 performs two gestures. Which happens first?}}} \\
      &                              & \cellcolor{gray!5}{A: Showing} \\
    \hhline{~|-|-}
      & G4: Reciprocal Patterns       & \cellcolor{gray!5}{\textit{\darkblue{Q: What is the most common gesture type performed throughout the video clip?}}} \\
      &                              & \cellcolor{gray!5}{A: Pointing} \\
    \hhline{~|-|-}
      & G5: Gesture Frequency         & \cellcolor{gray!5}{\textit{\darkblue{Q: Person 2 gives to Person 1 between 1.5--6.5s. Who does Person 1 give to next?}}} \\
      &                              & \cellcolor{gray!5}{A: Person 0} \\
    \hhline{~|-|-}
      & G6: Sequence Chains           & \cellcolor{gray!5}{\textit{\darkblue{Q: At 2.5s, Person 3 reaches toward Person 2. Who is Person 2 simultaneously reaching toward?}}} \\
      &                              & \cellcolor{gray!5}{A: Person 1} \\
    \hline

    \multirow{8}{*}{\textbf{Joint}}
      & J1: Gaze-Gesture Alignment    & \cellcolor{gray!5}{\textit{\darkblue{Q: Which happens first for Person 0: their sudden gaze shift, or their pointing gesture towards Person 2?}}} \\
      &                              & \cellcolor{gray!5}{A: The sudden gaze shift happens first} \\
    \hhline{~|-|-}
      & J2: Gaze Response to Gesture  & \cellcolor{gray!5}{\textit{\darkblue{Q: Just as Person 0 starts pointing at a card at 2.0s, who are they making eye contact with?}}} \\
      &                              & \cellcolor{gray!5}{A: Person 1} \\
    \hhline{~|-|-}
      & J3: Eye Contact / Interaction & \cellcolor{gray!5}{\textit{\darkblue{Q: Right before Person 1 starts a giving gesture at 3.5s, which two people share attention?}}} \\
      &                              & \cellcolor{gray!5}{A: Person 0 and Person 2} \\
    \hhline{~|-|-}
      & J4: Cross-Modal Dynamics      & \cellcolor{gray!5}{\textit{\darkblue{Q: Across 1.5--30.0s, who initiates both showing gestures and joins every gaze event?}}} \\
      &                              & \cellcolor{gray!5}{A: Person 0} \\
    \Xhline{1.0pt}
    \end{tabular}
    }
    \caption{
      Task examples in our 16-category social-reasoning benchmark.
      Each row shows one representative MCQ-style question--answer pair (only the
      correct answer text is shown for brevity). Difficulty progresses Easy~$\to$~Hard along the
      rows within each modality (T1--T2 Easy, T3--T4 Medium, T5--T6 Hard;
      analogous for G/J).
    }
    \label{tab:grasp_taxonomy_appendix}
    \vspace{-0.1cm}
\end{table*}


\subsection{Quality Control and Human Validation}
\label{app:quality_control}
\textit{GRASP} uses multiple levels of quality control to reduce noise from automatic annotation. First, all gaze and gesture events are filtered by confidence before graph construction. Second, gesture annotation follows a conservative protocol that skips ambiguous gestures and avoids inferring intent from non-gesture cues. Third, graph construction removes invalid temporal boundaries, duplicate events, and malformed event records while preserving gaze--gesture pairs required for joint reasoning. A graph-level health check verifies that required structured fields such as event type, participants, start time, end time, and confidence are present across the corpus.

\begin{figure*}[ht!]
\centering
\includegraphics[width=0.99\textwidth]{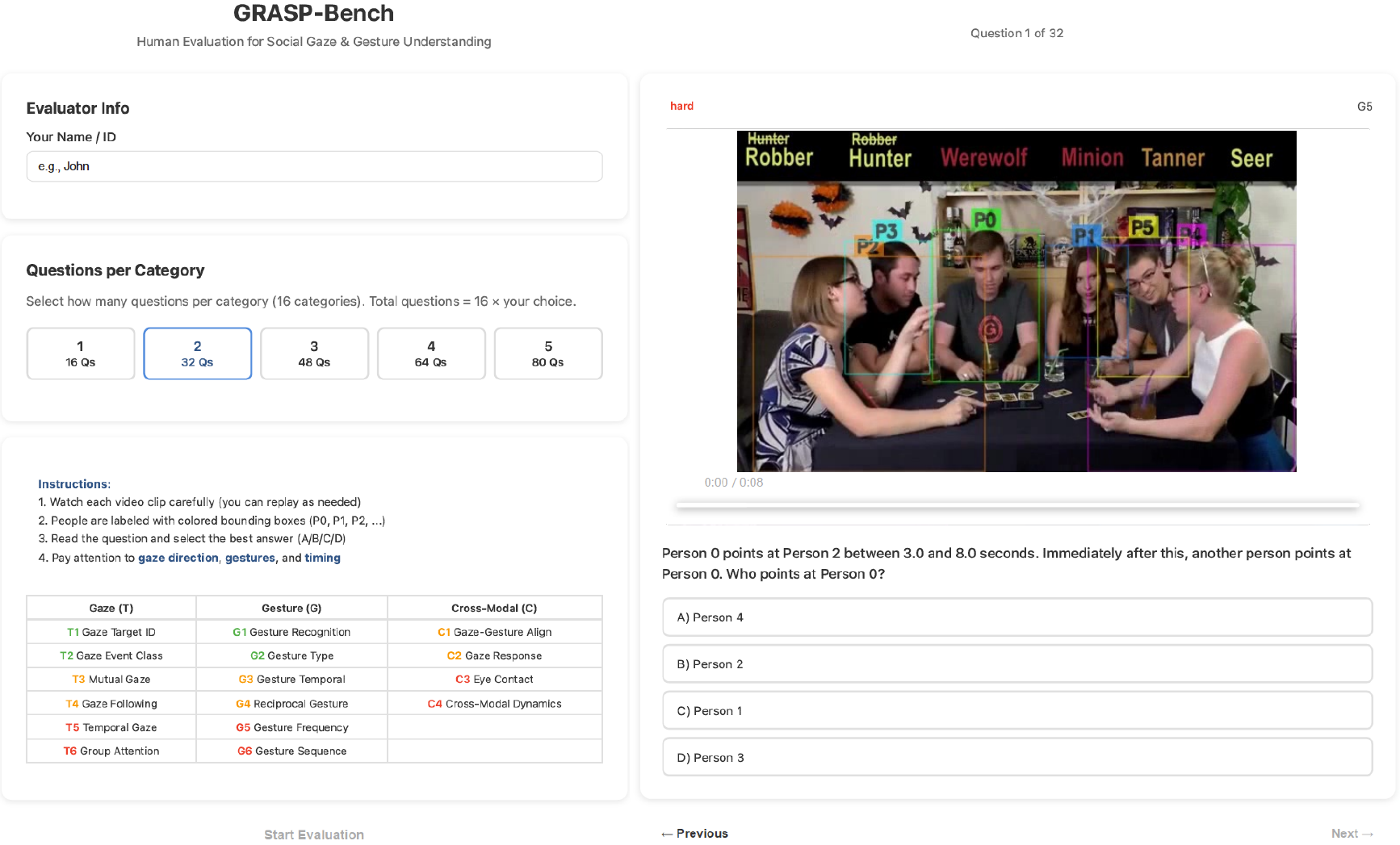}
\vspace*{-2mm}
\caption{
Human validation interface. Evaluators inspect each QA instance with the corresponding video clip, participant IDs, temporal context, category label, question, and answer options, then verify whether the QA pair is visually grounded and unambiguous.
}
\label{fig:human_validation_interface}
\end{figure*}

We additionally conduct human validation as an independent reliability audit. We randomly sample approximately 25\% of \textit{GRASP-Bench} in a category-balanced manner. Four human evaluators participate in the study. As shown in Fig.~\ref{fig:human_validation_interface}, evaluators inspect each QA instance with the corresponding video clip, participant IDs, temporal context, category label, question, and answer options. They judge whether the provided answer is visually supported, temporally grounded, and unambiguous.

Tab.~\ref{tab:human_validation_results} summarizes the validation results. The audited subset obtains an overall validity of 74.2\%, with gesture and joint reasoning showing higher validity rates of 78.9\% and 80.0\%, respectively. Gaze reasoning is comparatively more challenging, reflecting the difficulty of verifying fine-grained gaze direction, temporal ordering, and group attention in dense multi-person scenes. This validation serves as an independent sanity check on dataset reliability. The final dataset relies on structured event fields, confidence filtering, and deterministic validation rules, with human validation providing an additional estimate of QA quality.

\begin{table*}[t]
\centering
\small
\setlength{\tabcolsep}{4.6pt}
\caption{
Human validation results on \textit{GRASP-Bench}. Four evaluators audit approximately 25\% of the benchmark in a category-balanced manner. We report the percentage of QA instances judged visually supported, temporally grounded, and unambiguous.
}
\vspace{-2mm}
\label{tab:human_validation_results}
\resizebox{1.0\linewidth}{!}{
\begin{tabular}{@{}l l c !{\vrule} l l c !{\vrule} l l c@{}}
\Xhline{2\arrayrulewidth}
\multicolumn{3}{c !{\vrule}}{\textbf{Gaze Reasoning}}
& \multicolumn{3}{c !{\vrule}}{\textbf{Gesture Reasoning}}
& \multicolumn{3}{c}{\textbf{Joint Reasoning}} \\
\cmidrule(l{2pt}r{2pt}){1-3}
\cmidrule(l{2pt}r{2pt}){4-6}
\cmidrule(l{2pt}r{2pt}){7-9}
\rowcolor{headergray}
\textbf{Cat.} & \textbf{Difficulty} & \textbf{Validity (\%)}
& \textbf{Cat.} & \textbf{Difficulty} & \textbf{Validity (\%)}
& \textbf{Cat.} & \textbf{Difficulty} & \textbf{Validity (\%)} \\
\hline

T1 Gaze Target ID
& Easy
& 64.3
& G1 Gesture Recog.
& Easy
& 73.3
& J1 Gaze-Gest. Align.
& Medium
& 86.7 \\

T2 Gaze Event Class.
& Easy
& 57.1
& G2 Gesture Type
& Easy
& 86.7
& J2 Gaze Resp. to Gest.
& Medium
& 81.2 \\

T3 Mutual Gaze
& Medium
& 73.3
& G3 Gesture Temporal
& Medium
& 78.6
& J3 Eye Contact / Inter.
& Hard
& 71.4 \\

T4 Gaze Following
& Medium
& 88.2
& G4 Reciprocal Gesture
& Medium
& 93.8
& J4 Joint Person Dynamics
& Hard
& 80.0 \\

T5 Temporal Gaze
& Hard
& 42.9
& G5 Gesture Frequency
& Hard
& 80.0
& --
& --
& -- \\

T6 Group Attention
& Hard
& 62.5
& G6 Sequence Chains
& Hard
& 60.0
& --
& --
& -- \\
\hline

\rowcolor{headerblue}
\textbf{Gaze Avg.}
& --
& \textbf{65.6}
& \textbf{Gesture Avg.}
& --
& \textbf{78.9}
& \textbf{Joint Avg.}
& --
& \textbf{80.0} \\
\Xhline{2\arrayrulewidth}
\end{tabular}
}
\vspace{-1mm}
\end{table*}

\section{Detailed \textit{GRASP} Data Statistics}
\label{appendix:stats_detail}
We summarizes the scale, composition, and construction yield of \textit{GRASP}. Our video sources span diverse multi-person social settings, including multi-party TV dialogue~\cite{lei2020tvqa+,wang2025friends}, social intelligence reasoning~\cite{zadeh2019social}, multi-speaker conversation~\cite{nguyen2025see}, embodied multi-person behavior\footnote{We use only the multi-person subsets of Embody3D: \textit{multi-person conversation}, \textit{day-in-the-life}, and \textit{scenarios}.}~\cite{mclean2025embody}, and social deduction gameplay~\cite{lai2023werewolf,cao2025socialgesture}. We additionally augment the social deduction domain with online video sources, since these videos contain dense multi-person interactions with rich gaze and gesture signals. 

\begin{figure}[ht!]
    \centering
    \includegraphics[width=0.95\linewidth]{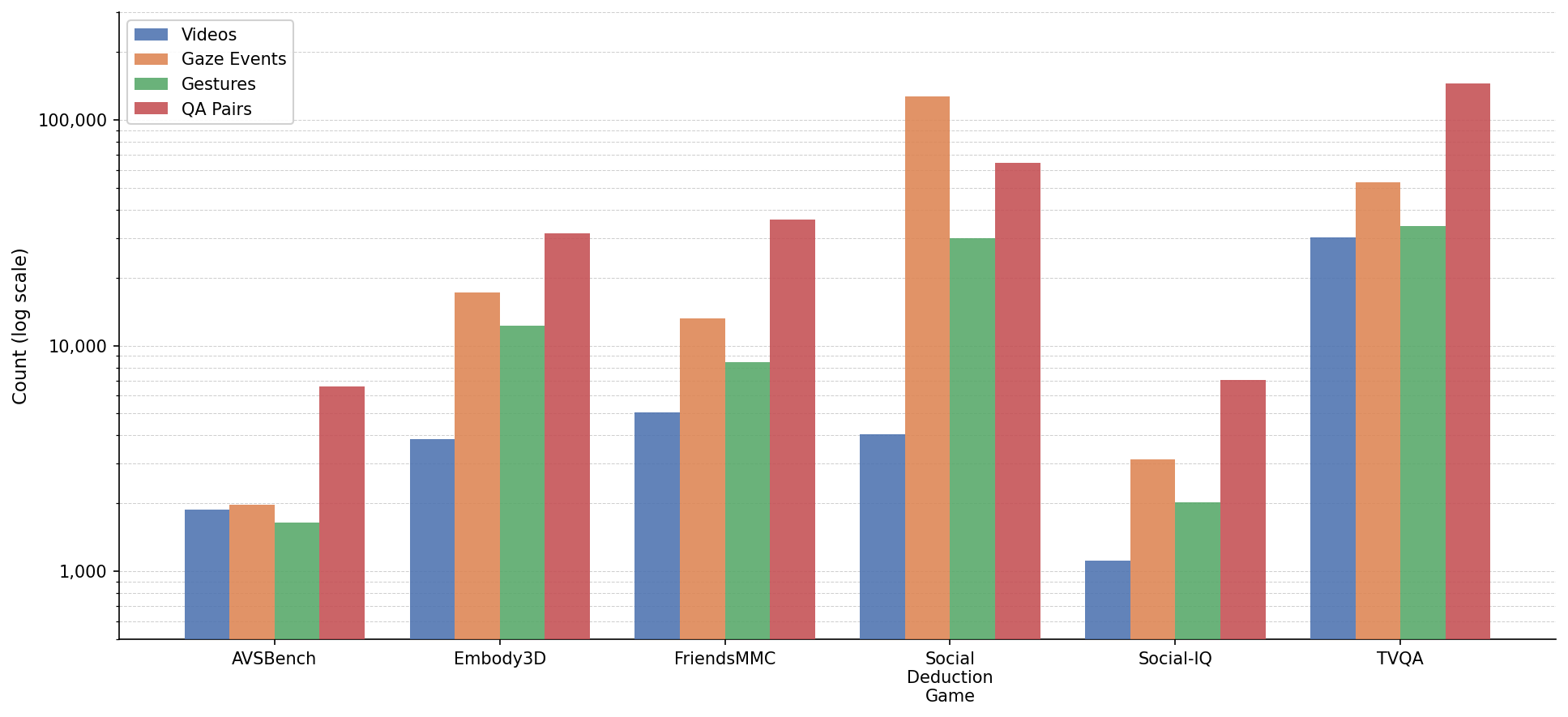}
    \caption{
    Dataset-level scale across the six source domains. We report the number of source videos, retained gaze events, detected gestures, and generated QA pairs for each dataset. The y-axis is shown in log scale to make datasets of different sizes comparable.
    }
    \label{fig:data_overview}
\end{figure}

\begin{figure}[ht!]
    \centering
    \includegraphics[width=0.72\linewidth]{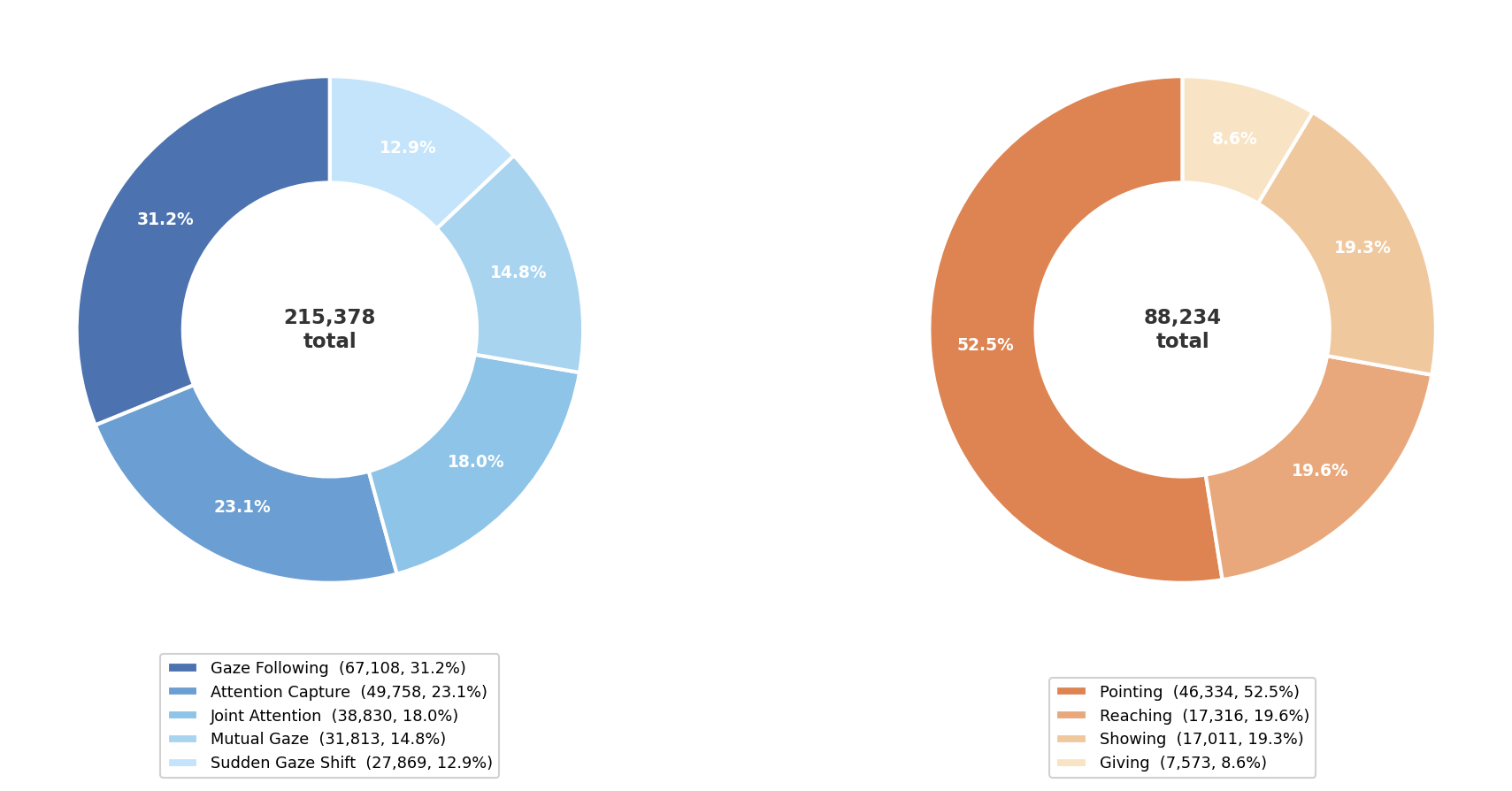}
    \caption{
    Distribution of retained gaze event types and gesture types. Gaze events are filtered at confidence $\geq 0.9$, and gestures are filtered at confidence $\geq 0.85$.
    }
    \label{fig:event_type_distribution}
\end{figure}

Across these sources, \textit{GRASP} contains 46,158 processed videos. From these videos, we construct structured social representations by extracting person-consistent gaze trajectories and deictic gestures, then integrating them into temporally ordered social graphs anchored to tracked participant identities. The pipeline detects 429,663 raw gaze events, retains 215,378 high-confidence gaze events using a confidence threshold of 0.9, and retains 88,234 gesture events using a confidence threshold of 0.85. After temporal merging and event deduplication, the pipeline produces 290,148 QA pairs across 16 categories. The final training data contains 238,296 multiple-choice QA pairs for RL training and 51,852 open-ended QA pairs for SFT training.

\begin{figure}[t!]
    \centering
    \includegraphics[width=0.9\linewidth]{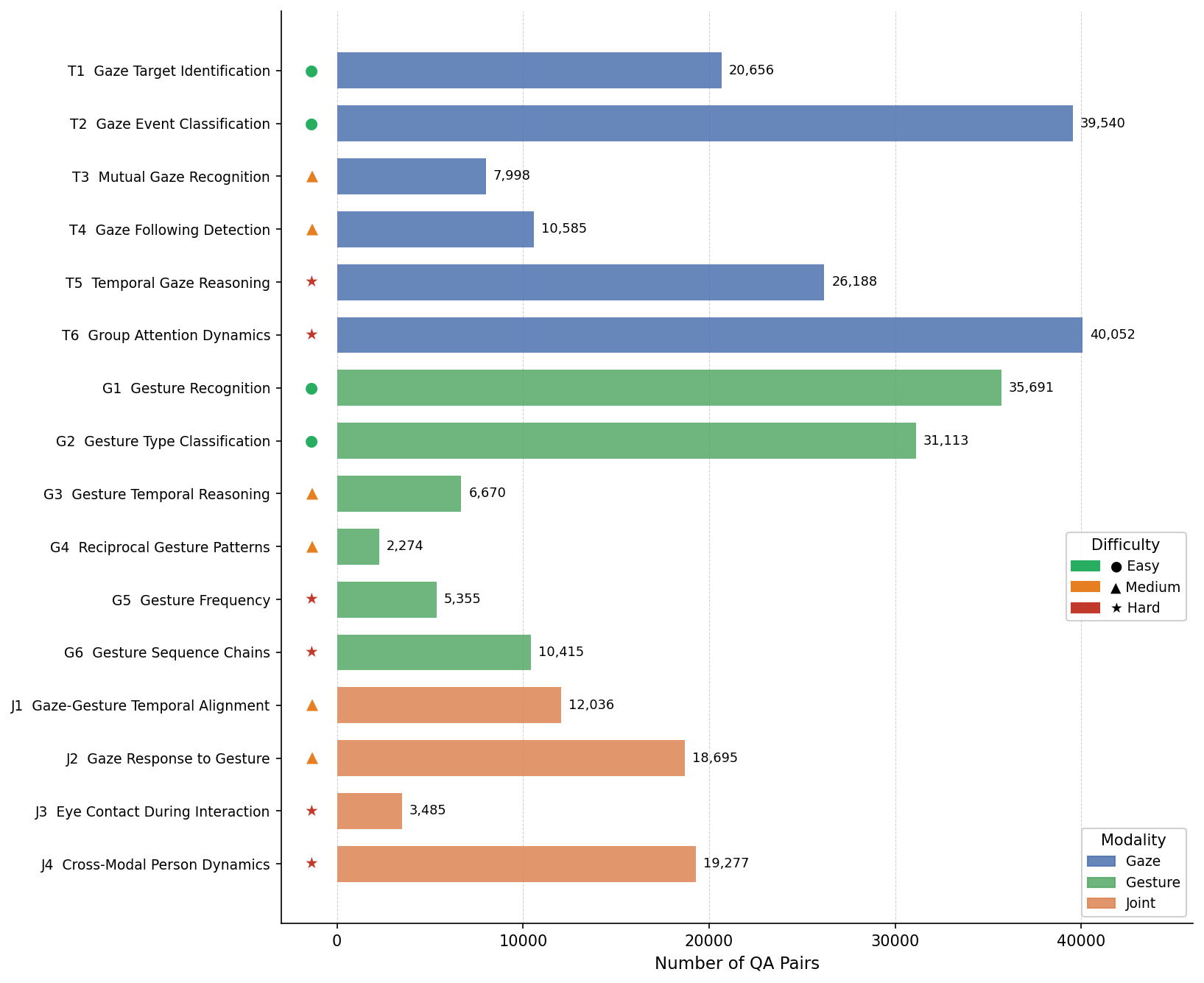}
    \caption{
    QA category distribution. The benchmark contains 16 categories: six gaze categories (T1--T6), six gesture categories (G1--G6), and four joint reasoning categories (J1--J4). Bar colors indicate modality, and markers indicate difficulty.
    }
    \label{fig:qa_category_distribution}
\end{figure}

\subsection{Per-Dataset Scale}
Fig.~\ref{fig:data_overview} shows the per-dataset scale. In particular, Social Deduction Game contains 4,044 videos but contributes 127,033 gaze events and 30,006 gestures, reflecting the dense multi-person interaction patterns in this domain.

\begin{figure}[t]
    \centering
    \includegraphics[width=0.95\linewidth]{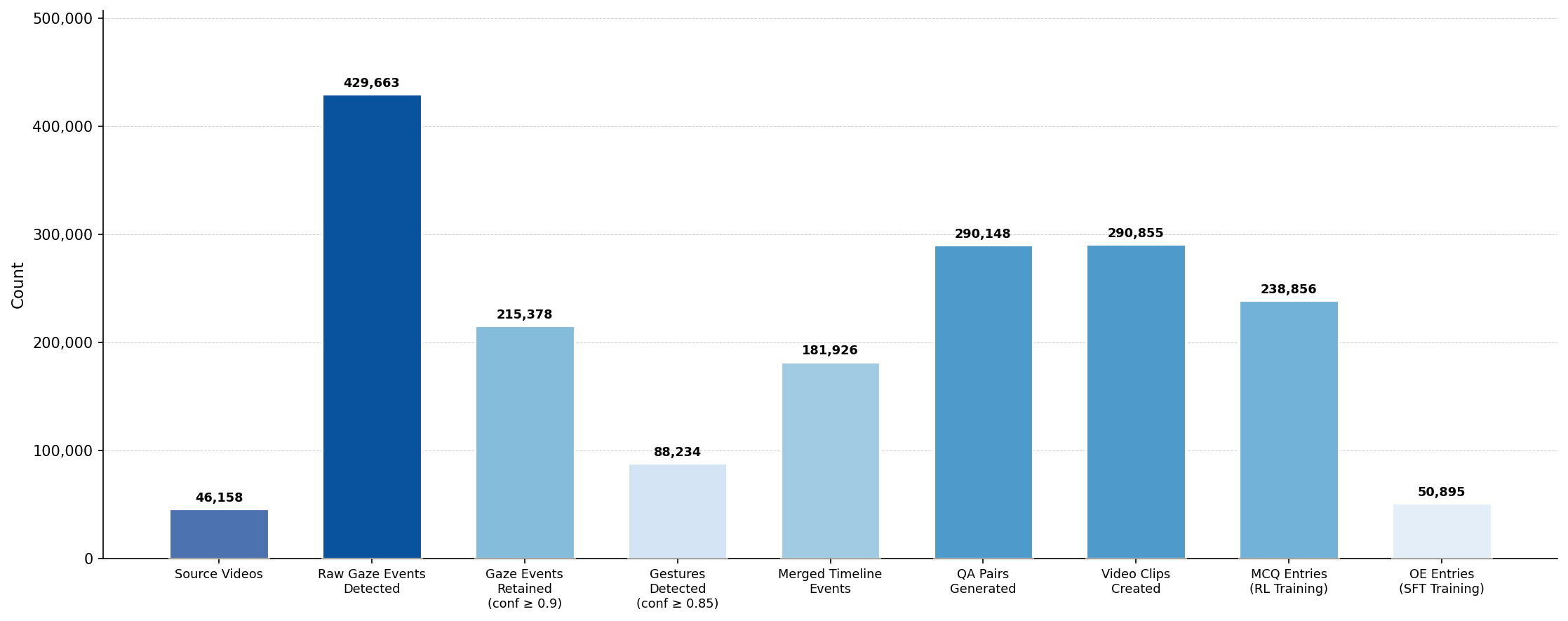}
    \caption{
    Data construction yield. The pipeline starts from source videos, detects raw gaze and gesture events, filters high-confidence gaze events, merges timeline events, and generates QA pairs and per-QA video clips for training.
    }
    \label{fig:pipeline_funnel}
\end{figure}

\begin{figure}[t]
    \centering
    \includegraphics[width=0.72\linewidth]{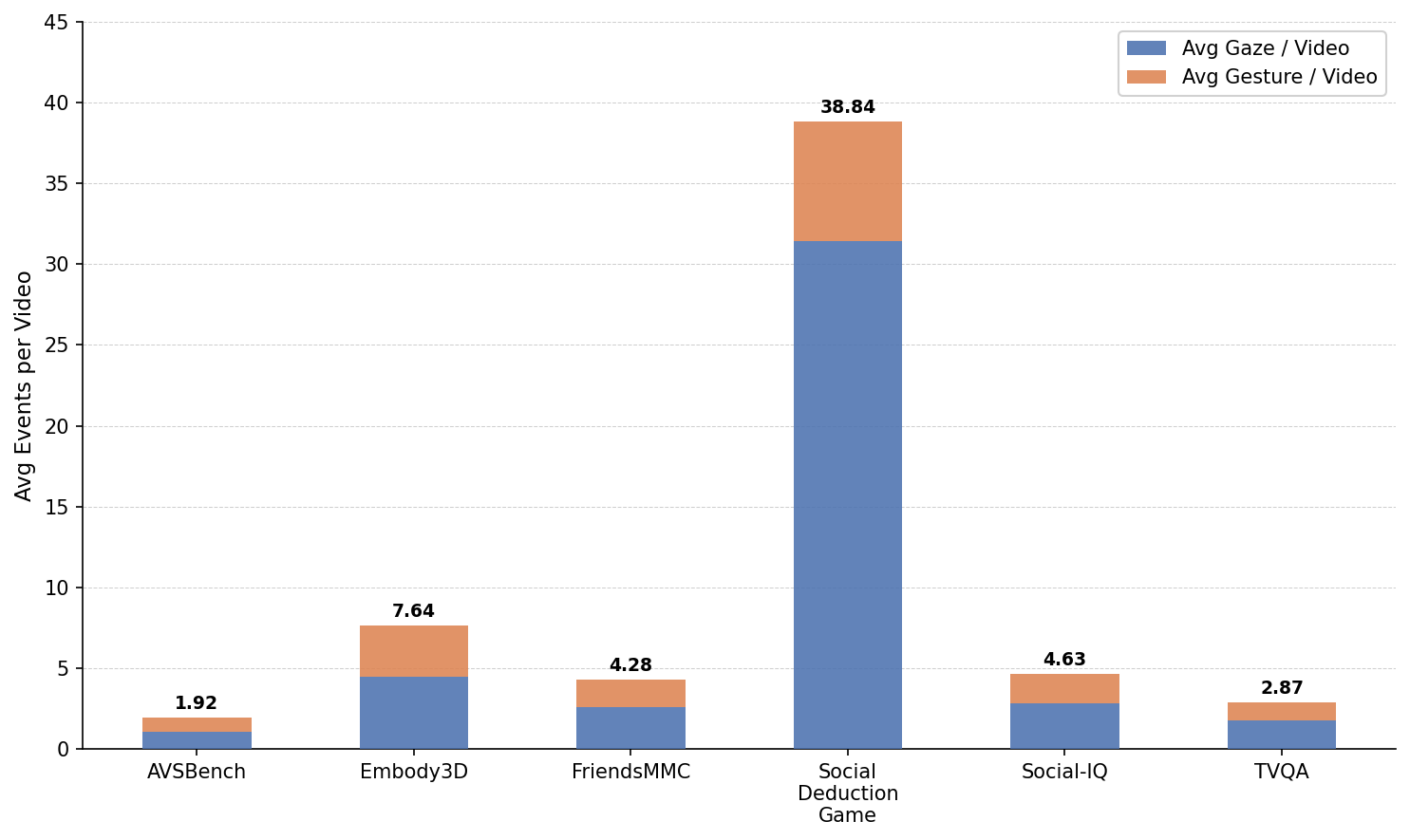}
    \caption{
    Average gaze and gesture event density per video. Social Deduction Game has the highest event density due to continuous multi-person social interaction.
    }
    \label{fig:event_density}
\end{figure}
\subsection{Gaze and Gesture Event Distributions}

As shown in Fig.~\ref{fig:event_type_distribution}, the retained gaze events cover five event types: gaze following, attention capture, joint attention, mutual gaze, and sudden gaze shift. Gaze following is the most frequent category, accounting for 31.2\% of retained gaze events, followed by attention capture at 23.1\%. Gesture annotations cover four types: pointing, reaching, showing, and giving. Pointing is the dominant gesture type, accounting for 52.5\% of all detected gestures, which is consistent with its role as a frequent cue in social and referential interaction.

\subsection{QA Category Distribution}

The QA taxonomy is organized into three groups: gaze reasoning, gesture reasoning, and joint reasoning. Gaze categories include target identification, event classification, mutual gaze recognition, gaze following detection, temporal gaze reasoning, and group attention dynamics. Gesture categories cover recognition, type classification, temporal reasoning, reciprocal patterns, frequency, and sequence chains. Joint reasoning categories require relating gaze and gesture events over time, including gaze-gesture temporal alignment, gaze response to gesture, eye contact during interaction, and cross-modal person dynamics. We show their distribution in Fig.~\ref{fig:qa_category_distribution}

\subsection{Construction Yield}
Fig.~\ref{fig:pipeline_funnel} summarizes the full construction pipeline. The confidence filter retains 215,378 of 429,663 raw gaze detections, corresponding to 50.1\% retention. The retained gaze events and detected gestures are then merged into 181,926 timeline events. Since each timeline event can support different question forms, including single-event recognition, temporal comparison, sequence reasoning, and cross-modal reasoning, the final number of QA pairs is larger than the number of merged timeline events.

\subsection{Event Density and Training Distribution}
Fig.~\ref{fig:event_density} reports the average number of retained gaze events and detected gestures per video across source domains. Event density varies substantially across domains. Social Deduction Game has the highest average density, with 31.42 gaze events and 7.42 gestures per video. Other datasets are sparser, typically containing fewer than eight total events per video. This variation is expected because the source datasets differ in interaction structure, video duration, and the frequency of visible social cues.

\begin{figure}[t]
    \centering
    \includegraphics[width=0.62\linewidth]{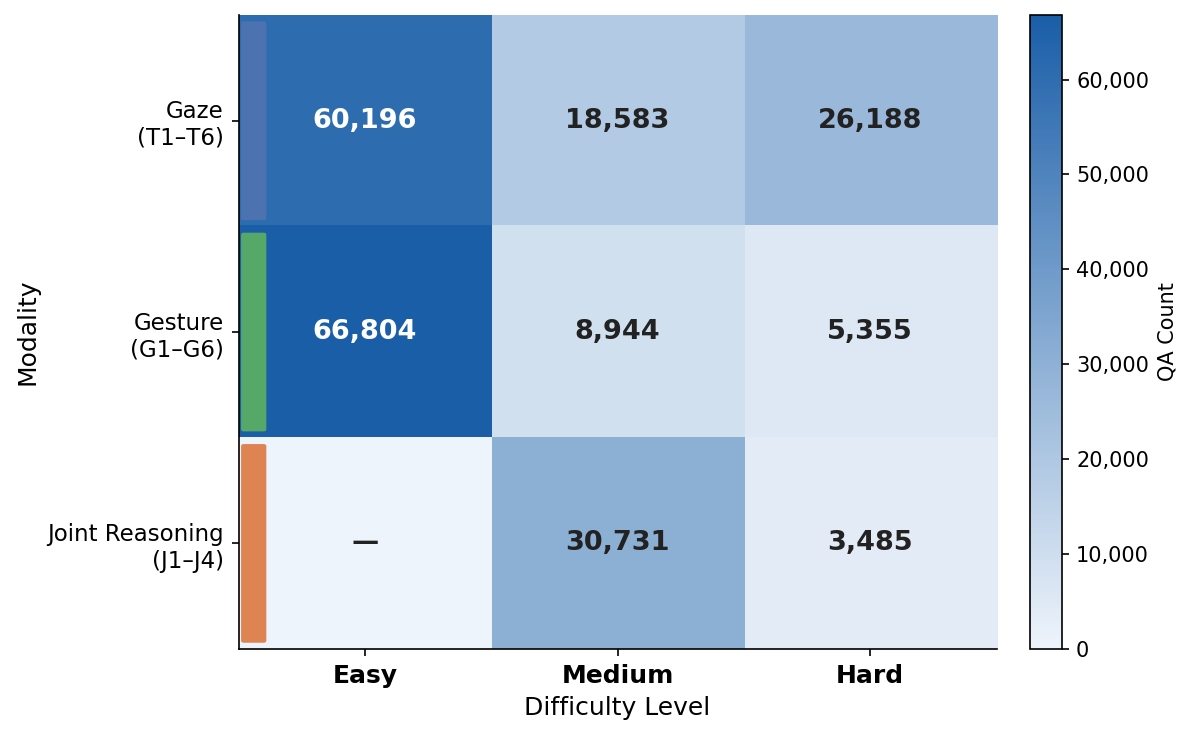}
    \caption{
    Distribution of MCQ training examples by modality and difficulty. Open-ended examples are excluded from this heatmap because they are used for SFT and are categorized as hard reasoning examples.
    }
    \label{fig:modality_difficulty}
\end{figure}

Fig.~\ref{fig:modality_difficulty} summarizes the distribution of MCQ training examples by modality and difficulty. Easy examples are mostly direct gaze or gesture recognition questions, while medium and hard examples require temporal or cross-modal reasoning. Joint reasoning does not include easy questions by design, since these examples require relating gaze and gesture events or reasoning about person-level interaction dynamics.

Overall, the constructed data provides 145,049 gaze QA pairs, 91,536 gesture QA pairs, and 53,504 joint reasoning QA pairs. In terms of format, the final data is primarily multiple-choice, with 238,296 MCQ examples and 51,852 open-ended examples. The MCQ split is used for RL training, while the open-ended split is used for SFT.

\begin{table*}[t!]
\centering
\caption{Comparison of \textit{GRASP} with existing multimodal social understanding benchmarks. Our dataset scales significantly beyond prior works while distinctly focusing on fine-grained, participant-level non-verbal grounding.}
\label{tab:social_benchmarks_comparison}
\resizebox{\linewidth}{!}{%
\begin{tabular}{@{}lllrrp{7.5cm}@{}}
\toprule
\textbf{Benchmark} & \textbf{Input} & \textbf{Grounded Cues} & \textbf{\#Samples} & \textbf{\#QAs} & \textbf{Core Focus} \\ \midrule
GazeVQA~\cite{wang2025gaze} & Image & Gaze & 122K & 410K & Re-label GazeFollow~\cite{recasens2015they}. Static gaze target understanding in single frames \\
Social-IQ 2.0~\cite{siq2} & Video & - & 1.2K & 7.5K & Unconstrained social situations, explicit human reasoning \\
SocialGesture~\cite{cao2025socialgesture} & Video & Gesture & 10K & 10K & Multi-person deictic gesture understanding \\
SIV-Bench~\cite{kong2025siv} & Video & - & 2.7K & 5.4K & Social relation theory, state reasoning, dynamics prediction \\
Social Genome~\cite{mathur2025social} & Video & - & 272 & 1.5K & Grounded multi-step reasoning traces and external knowledge \\
SocialOmni~\cite{xie2026socialomni} & Video & - & 2K & 2.2K & Omni-modal conversational interactivity (who, when, how) \\
Omni-MMSI~\cite{li2026omni} & Video & - & 7K & 5K & Re-label MMSI~\cite{lee2024modeling}. Identity-attributed interaction and multi-party speaker reference \\
V-Social~\cite{lin2025v} & Video & - & 214 & 956 & Social relation recognition and grounding in long-form videos \\
R$^3$-VQA~\cite{niu2026read} & Video & - & 316 & 4.8K & Mental state estimation and social causal reasoning \\
MimeQA~\cite{li2025mimeqa} & Video & - & 221 & 806 & Social reasoning for silent theatrical storytelling \\
\midrule
\textbf{GRASP (Ours)} & \textbf{Video} & \textbf{Gaze, Gesture} & \textbf{46K} & \textbf{290K} & \textbf{Grounded participant-level events and dynamic social tracking} \\ \bottomrule
\end{tabular}%
}
\end{table*}

\subsection{Comparison with Specialized Social VQA Benchmarks}
As summarized in Table~\ref{tab:social_benchmarks_comparison}, \textit{GRASP} occupies a unique position in the social AI landscape. Unlike single grounded cues datasets such as GazeVQA~\cite{wang2025gaze} and SocialGesture~\cite{cao2025socialgesture}, which excel in static spatial gaze target or gesture understanding and localization, \textit{GRASP} emphasizes the temporal dynamics of social events (\textit{e.g.,} joint attention and turn-taking). Furthermore, while reasoning-centric benchmarks like R$^3$-VQA~\cite{niu2026read} and Social Genome~\cite{mathur2025social} offer deep cognitive insights, their limited scale often restricts their use to evaluation rather than post-training. \textit{GRASP} provides the necessary scale (46K videos) and multi-modal grounding (gaze and gesture) to facilitate the alignment of MLLMs with nuanced human social behaviors.

\section{Additional Analyses and Experimental Details}
\label{appendix:exp}
\subsection{Training Details}
We train our models with the two-stage procedure described in \S~4. First, we perform supervised fine-tuning on the open-ended \textit{GRASP} subset to warm up the model with the structured reasoning format. We then apply GRPO-based reinforcement learning on the multiple-choice training split, using answer correctness, output-format consistency, structural grounding, and social grounding as the reward components. The detailed hyperparameters are summarized in Tab.~\ref{tab:train_hparams}.

For both SFT and RL, input videos are sampled at 2 fps and training is performed in \textit{bfloat16} with AdamW, cosine learning-rate scheduling, gradient clipping, and DeepSpeed ZeRO-2. The SFT stage is trained for two epochs with a learning rate of \(2\times10^{-6}\), while the RL stage is trained for one epoch with a learning rate of \(3\times10^{-6}\). For GRPO, we use \(K=8\) rollouts per query, a KL penalty of \(0.05\), mean-centered advantage normalization, and advantage clipping to \(\pm 5.0\). The reward weights are set to \(1.0/0.1/0.05/0.2\) for accuracy, format, structural grounding, and social grounding, respectively.

\begin{table}[h!]
\centering
\caption{Detailed training hyperparameters.}
\label{tab:train_hparams}
\small
\resizebox{0.7\linewidth}{!}{
\begin{tabular}{@{}lll@{}}
\toprule
Config                  & SFT                       & RL (GRPO) \\
\midrule
Input frames            & 2 fps                     & 2 fps \\
LR scheduler            & Cosine with warm-up       & Cosine with warm-up \\
Warmup steps            & 200                       & 256 \\
Optimizer               & AdamW ($\beta_1{=}0.9$, $\beta_2{=}0.999$) & AdamW \\
Global batch size       & 32                        & 32 \\
Learning rate           & $2 \times 10^{-6}$        & $3 \times 10^{-6}$ \\
Weight decay            & $1 \times 10^{-3}$        & $1 \times 10^{-3}$ \\
Gradient clipping       & 1.0                       & 1.0 \\
Training precision      & \textit{bfloat16}         & \textit{bfloat16} \\
DeepSpeed               & ZeRO-2                    & ZeRO-2 \\
Epochs                  & 2                         & 1 \\
Max generation length   & --                        & 1024 \\
\midrule
Number of rollouts $K$  & --                        & 8 \\
KL penalty $\beta$      & --                        & 0.05 \\
Advantage normalization & --                        & Mean-center \\
Advantage clipping      & --                        & $\pm 5.0$ \\
Reward weights          & --                        & 1.0 / 0.1 / 0.05 / 0.2 \\
\bottomrule
\end{tabular}
}
\end{table}

\subsection{Other Benchmark Details}
We use MMSI ~\cite{lee2024modeling}, Online-MMSI ~\cite{li2025towards}, and TVQA+ ~\cite{lei2020tvqa+} to test whether GRASP-Bench transfers beyond in-domain gaze/gesture reasoning to related forms of human-centered video understanding, including person-reference tracking in multi-party conversation and spatio-temporally grounded video QA.

MMSI is built from real multi-party social deduction game videos with 151 YouTube games totaling 14.8 hours and 48 Ego4D games totaling 7.3 hours. It defines three dialogue-based tasks: Speaking Target Identification (STI), Pronoun Coreference Resolution (PCR), and Mentioned Player Prediction (MPP). These tasks ask models to identify who is being addressed, referred to, or mentioned, with 4,087 STI, 3,182 PCR, and 3,832 MPP instances across the YouTube and Ego4D subsets. In this game setting, references occur in conversations shaped by accusation, defense, hidden-role claims, persuasion, and shifting group attention. Although gaze and gesture are not the labels being predicted, they are part of the visible evidence humans use to track who a speaker is addressing or discussing [TODO: CITE]. Online-MMSI uses the same source data and tasks but restricts the input to context before the target utterance, giving us a related test of person-reference tracking in a setup more realistic to active dialogue. 

TVQA+ comes from TVQA ~\cite{lei2018tvqa}, which contains 152K QA pairs from 21.8K clips across 460 hours of six TV shows. TVQA+ augments the Big Bang Theory subset with spatial grounding annotations, yielding 29,383 multiple-choice questions from 4,198 clips, 148,468 annotated images, 310,826 bounding boxes, and 2,527 visual concept categories. The benchmark asks models to answer questions from subtitles and video while localizing relevant moments and grounding visual concepts such as people and objects. The questions are not strictly social reasoning, but they are human-centered, asking who did something, what someone was holding or wearing, where someone was, what direction someone turned, or what object was involved. These questions often require interpreting people’s actions and relations to objects, space, and time, where gaze, gesture, body orientation, and movement can provide useful visual evidence.

\begin{table*}[t]
\centering
\small
\setlength{\tabcolsep}{4.2pt}
\caption{
Participant-ID corruption test on \textit{GRASP-Bench}. We consistently remap Person IDs in the question, answer options, and metadata answer text while keeping the video unchanged and preserving the correct answer letter. Normal and Corrupt denote accuracy (\%), and Drop denotes Corrupt minus Normal in percentage points.
}
\vspace{-2mm}
\label{tab:id_corruption}
\resizebox{0.75\linewidth}{!}{
\begin{tabular}{@{}l l ccc ccc@{}}
\Xhline{2\arrayrulewidth}
\multirow{2}{*}{\textbf{Cat.}} 
& \multirow{2}{*}{\textbf{Task}} 
& \multicolumn{3}{c}{\textbf{Qwen3-VL-8B + \textit{SGR}}}
& \multicolumn{3}{c}{\textbf{Qwen3.5-9B + \textit{SGR}}} \\
\cmidrule(l{2pt}r{2pt}){3-5}
\cmidrule(l{2pt}r{2pt}){6-8}
& & \textbf{Normal} & \textbf{Corrupt} & \textbf{Drop}
  & \textbf{Normal} & \textbf{Corrupt} & \textbf{Drop} \\
\hline

\rowcolor{headergray} \multicolumn{8}{l}{\textit{Gaze Reasoning}} \\
T1 & Gaze Target ID
& 37.1 & 27.1 & $-10.0$
& 37.1 & 21.4 & $-15.7$ \\

T2 & Gaze Event Class.
& 65.8 & 61.4 & $-4.4$
& 70.2 & 59.1 & $-11.1$ \\

T3 & Mutual Gaze Recog.
& 47.3 & 49.1 & $+1.8$
& 50.0 & 48.6 & $-1.4$ \\

T4 & Gaze Following
& 41.8 & 19.4 & $-22.4$
& 46.3 & 16.4 & $-29.9$ \\

T5 & Temporal Gaze
& 43.2 & 23.9 & $-19.3$
& 46.6 & 21.6 & $-25.0$ \\

T6 & Group Attention
& 33.3 & 25.0 & $-8.3$
& 27.4 & 24.7 & $-2.7$ \\

\rowcolor{gray!5}
\multicolumn{2}{l}{\(\rightarrow\) Gaze Avg.}
& 46.3 & 35.9 & $-10.4$
& 48.0 & 33.7 & $-14.3$ \\
\hline

\rowcolor{headergray} \multicolumn{8}{l}{\textit{Gesture Reasoning}} \\
G1 & Gesture Recog.
& 50.8 & 26.4 & $-24.4$
& 54.9 & 29.8 & $-25.1$ \\

G2 & Gesture Type
& 90.1 & 100.0 & $+9.9$
& 90.1 & 50.2 & $-39.9$ \\

G3 & Gesture Temporal
& 38.8 & 36.6 & $-2.2$
& 59.2 & 22.0 & $-37.2$ \\

G4 & Reciprocal Gesture
& 39.0 & 11.8 & $-27.2$
& 46.3 & 10.8 & $-35.5$ \\

G5 & Gesture Frequency
& 25.8 & 23.3 & $-2.5$
& 41.9 & 22.6 & $-19.3$ \\

G6 & Sequence Chains
& 59.1 & 63.6 & $+4.5$
& 63.6 & 50.0 & $-13.6$ \\

\rowcolor{gray!5}
\multicolumn{2}{l}{\(\rightarrow\) Gesture Avg.}
& 58.0 & 30.2 & $-27.8$
& 64.4 & 27.7 & $-36.7$ \\
\hline

\rowcolor{headergray} \multicolumn{8}{l}{\textit{Joint Reasoning}} \\
J1 & Gaze-Gest. Align.
& 53.9 & 34.9 & $-19.0$
& 48.7 & 37.6 & $-11.1$ \\

J2 & Gaze Resp. to Gest.
& 22.7 & 31.8 & $+9.1$
& 40.9 & 18.2 & $-22.7$ \\

J3 & Eye Contact / Inter.
& 51.3 & 34.2 & $-17.1$
& 48.7 & 35.5 & $-13.2$ \\

J4 & Joint Person Dynamics
& 43.2 & 13.9 & $-29.3$
& 39.2 & 27.8 & $-11.4$ \\

\rowcolor{gray!5}
\multicolumn{2}{l}{\(\rightarrow\) Joint Avg.}
& 48.1 & 29.0 & $-19.1$
& 45.6 & 33.0 & $-12.6$ \\
\hline

\rowcolor{headerblue}
\multicolumn{2}{l}{\textbf{Overall}}
& \textbf{50.4} & \textbf{32.6} & \textbf{$-17.8$}
& \textbf{52.6} & \textbf{32.0} & \textbf{$-20.6$} \\
\Xhline{2\arrayrulewidth}
\end{tabular}
}
\vspace{-1mm}
\end{table*}

\subsection{Participant-ID Corruption Analysis}
\label{app:id_corruption}
A potential concern is that \textit{SGR} could be exploited by simply repeating participant IDs that already appear in the question, rather than grounding the reasoning trace in the video. To diagnose this shortcut, we conduct a participant-ID corruption test on \textit{GRASP-Bench}. For each MCQ example, we consistently remap all Person IDs in the question, answer options, and metadata answer text, while leaving the video unchanged and preserving the correct answer letter. This keeps the textual format and answer-position prior intact, but breaks the alignment between textual Person IDs and the visual Person IDs overlaid in the video. If a model mainly relied on text-only ID shortcuts, its accuracy should remain stable under this corruption. In contrast, a visually grounded model should degrade when the answer requires mapping textual participant references to the correct people in the video.

As shown in Tab.~\ref{tab:id_corruption}, both SGR-trained models drop substantially under participant-ID corruption. Qwen3-VL-8B+\textit{SGR} drops from 50.4\% to 32.6\%, and Qwen3.5-9B+\textit{SGR} drops from 52.6\% to 32.0\%. The degradation is strongest in categories that require resolving specific participants, such as gaze following (T4), temporal gaze reasoning (T5), gesture recognition (G1), reciprocal gestures (G4), and joint person dynamics (J4). For example, T4 drops by 22.4pp and 29.9pp, G4 drops by 27.2pp and 35.5pp, and J4 drops by 29.3pp and 11.4pp for the two models, respectively.

The category-level pattern also clarifies when corruption should or should not affect performance. Categories with weaker direct dependence on participant identity, such as mutual gaze duration or some group-attention questions, show smaller or mixed changes. In contrast, categories that require tracking who looks at whom, who gestures to whom, or who participates in both gaze and gesture events consistently show large drops. This diagnostic does not rule out every possible form of reward gaming, but it directly addresses the ID-repetition shortcut: the final SGR-trained policies depend on the consistency between textual participant references and visually grounded participant identities, rather than merely reusing Person IDs from the question.

\subsection{Reasoning Length Breakdown}

\begin{figure*}[ht!]
\centering
\includegraphics[width=0.8\textwidth]{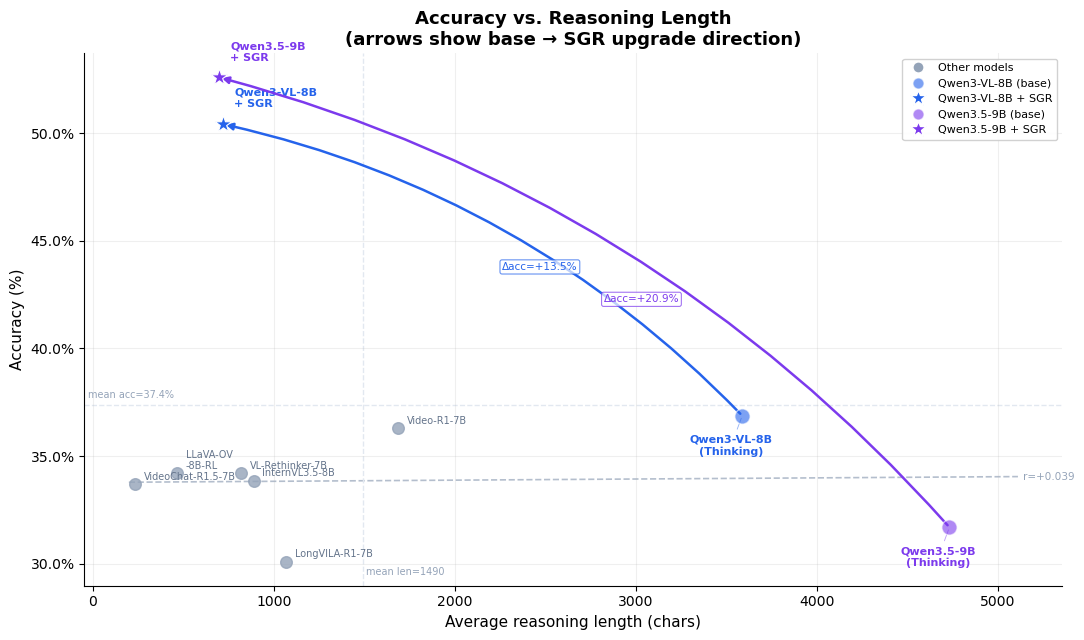}
\vspace*{-2mm}
\caption{
Accuracy compared against average reasoning length for all baselines.
}
\label{fig:length_analysis}
\end{figure*}
To understand how accuracy and reasoning length relate, we plot the accuracy/length in Fig.~\ref{fig:length_analysis} and perform Pearson correlation analysis. We find no correlation (Pearson r = + 0.039) between accuracy and average reasoning length for all baseline models for the collection of social reasoning tasks in \textit{GRASP-Bench}. This runs counter to findings for other VQA tasks where longer reasoning helps performance. Here longer reasoning is not necessarily better reasoning. Meanwhile, our method simultaneously shortens the reasoning length for the our models and improves the accuracy.

\subsection{\textit{SGR} Error Profile}
\begin{figure*}[ht!]
\centering
\includegraphics[width=0.8\textwidth]{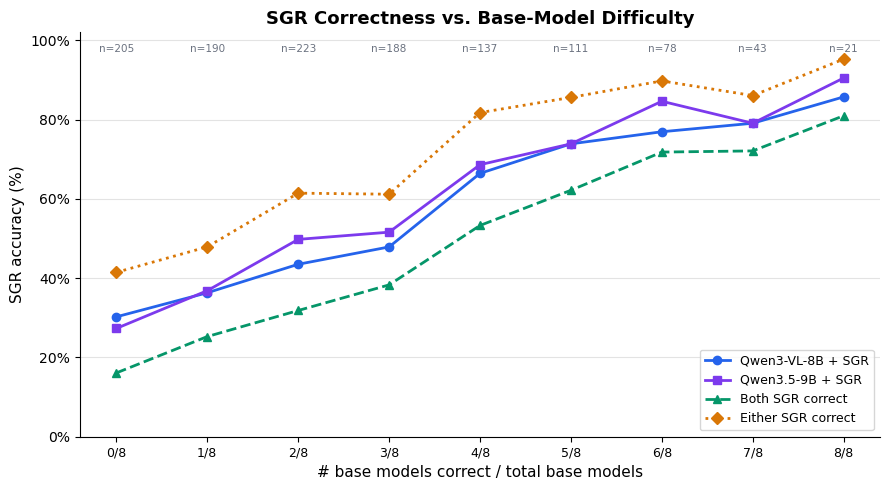}
\vspace*{-2mm}
\caption{
Error profile for \textit{SGR} compared against difficulty of the questions for the base models.
}
\label{fig:difficulty}
\end{figure*}

To understand the difficulty of the \textit{GRASP-Bench} tasks, we group cases into buckets defined by how many base models were correct, and the accuracy of the our models per bucket was computed as shown in Fig.~\ref{fig:difficulty}. This gives an estimation of how hard the \textit{GRASP} questions are for our comparison models off the shelf and allows us to evaluate the \textit{SGR} error profile. There are three primary takeaways:
\begin{itemize}[leftmargin=1.2em, itemsep=0.2em, topsep=0.2em]
    \item The \textit{SGR} performance tracks along the scale of difficulty described by the base model performance. This suggests there are other things contributing to the complexity of the benchmark tasks which are not being captured by our reward.
    \item There exist ``hard'' cases that all other models fail on where we recover using the \textit{SGR} approach (like Fig.~\ref{fig:1}), but there are also ``easier'' cases where all other models are correct but our approach fails (examples in Appendix.~\ref{appendix:qualitative_failures}).
    \item Our models separately follow similar error profiles but our two models are making different errors, creating an opening for future work to investigate the characteristics of the errors being made by each model when supported by \textit{SGR}.
\end{itemize}

\section{Qualitative Comparison}
\label{appendix:qualitative}

We provide qualitative examples to complement the main results in the paper. The examples illustrate how \textit{GRASP-Bench} questions are grounded in fine-grained gaze and gesture events, how different models reason over the same multi-person interaction, and where our method still fails.

\subsection{\textit{GRASP-Bench} Examples}
\label{appendix:qualitative_examples}
\begin{figure*}[t]
\centering
\includegraphics[width=1.0\textwidth]{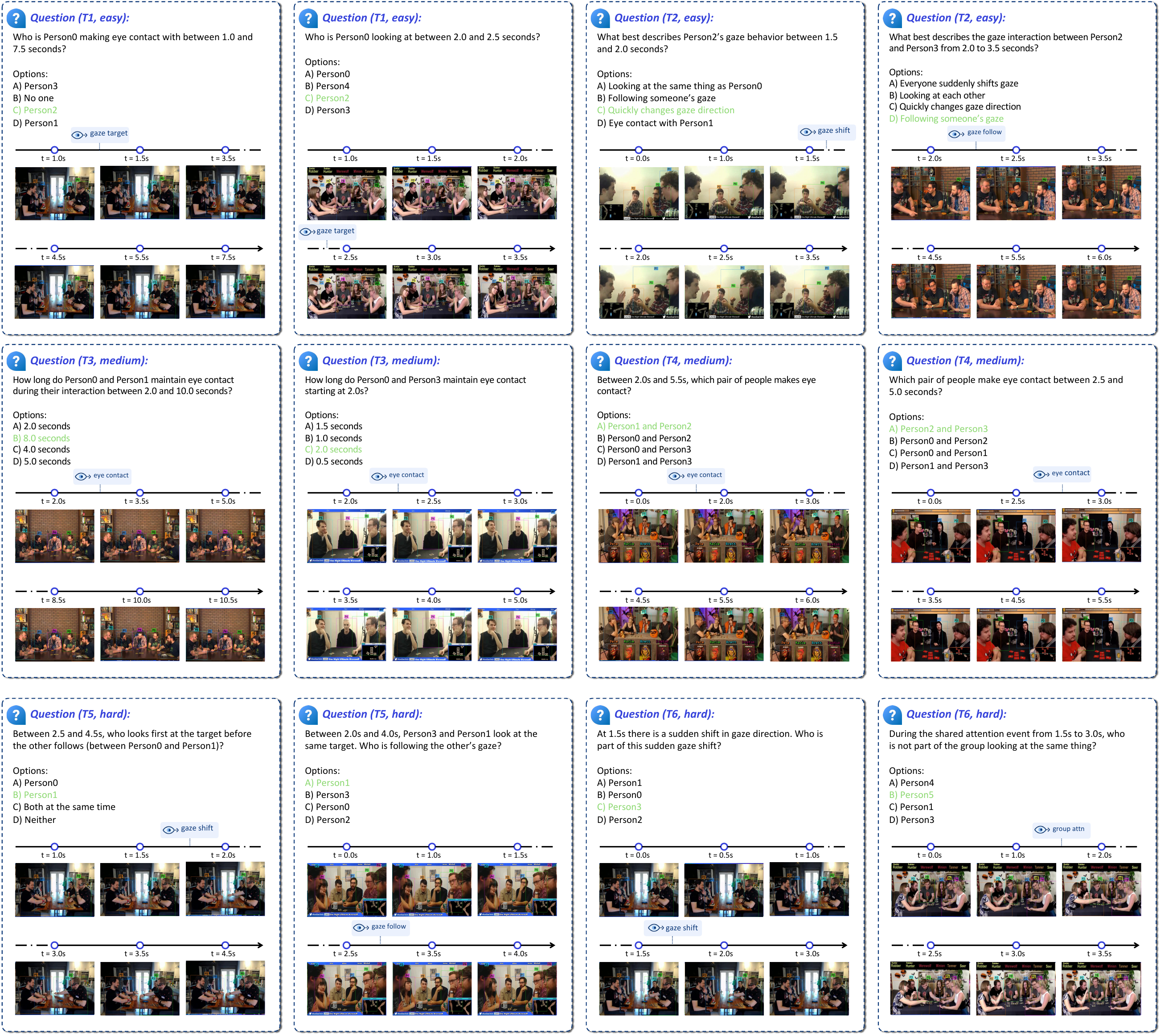}
\caption{
Qualitative \textit{GRASP-Bench} examples for gaze reasoning, covering T1--T6.
}
\label{fig:grasp_gaze}
\end{figure*}

\begin{figure*}[t]
\centering
\includegraphics[width=1.0\textwidth]{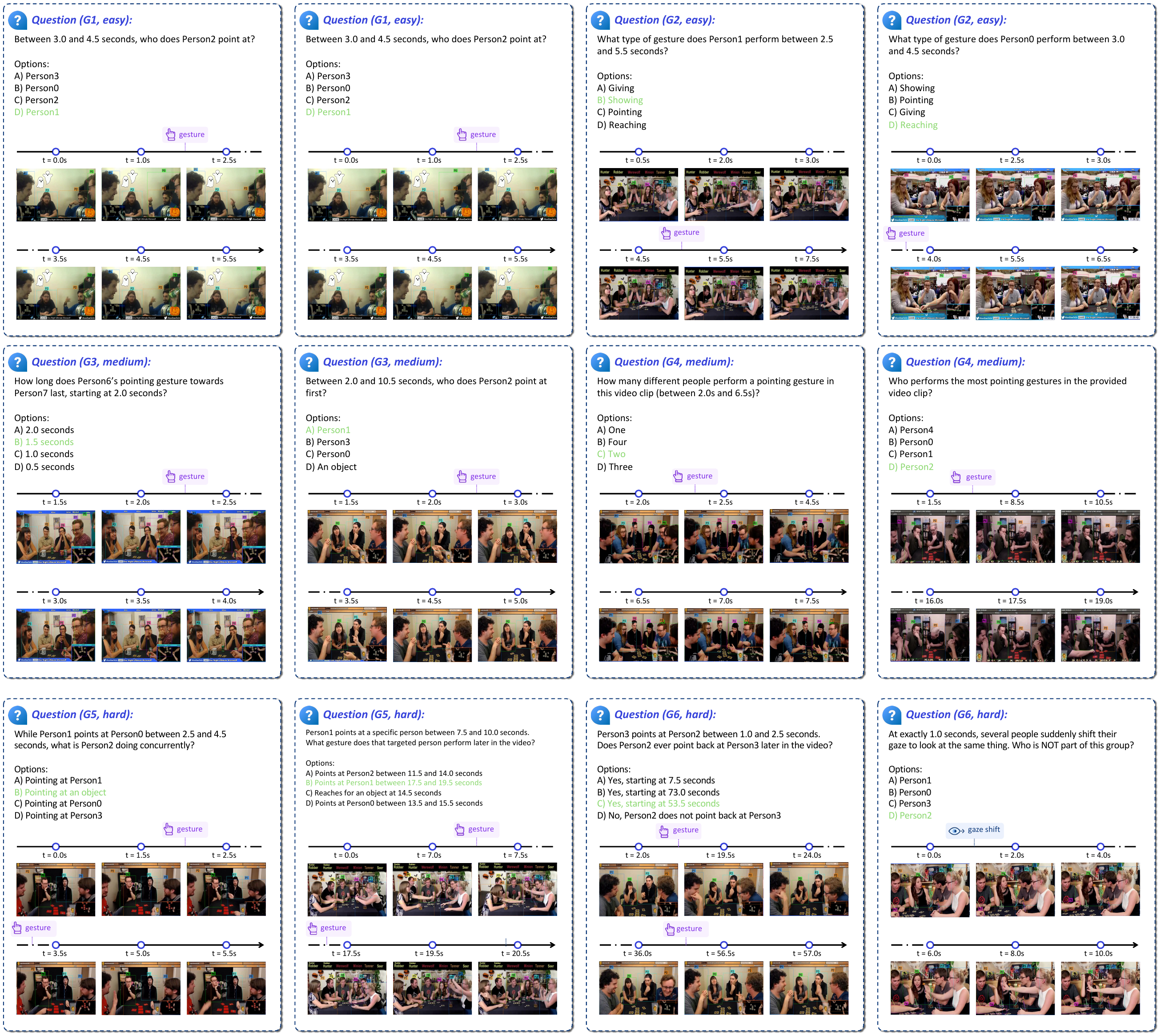}
\caption{
Qualitative \textit{GRASP-Bench} examples for gesture reasoning, covering G1--G6.
}
\label{fig:grasp_gesture}
\end{figure*}

\begin{figure*}[t]
\centering
\includegraphics[width=1.0\textwidth]{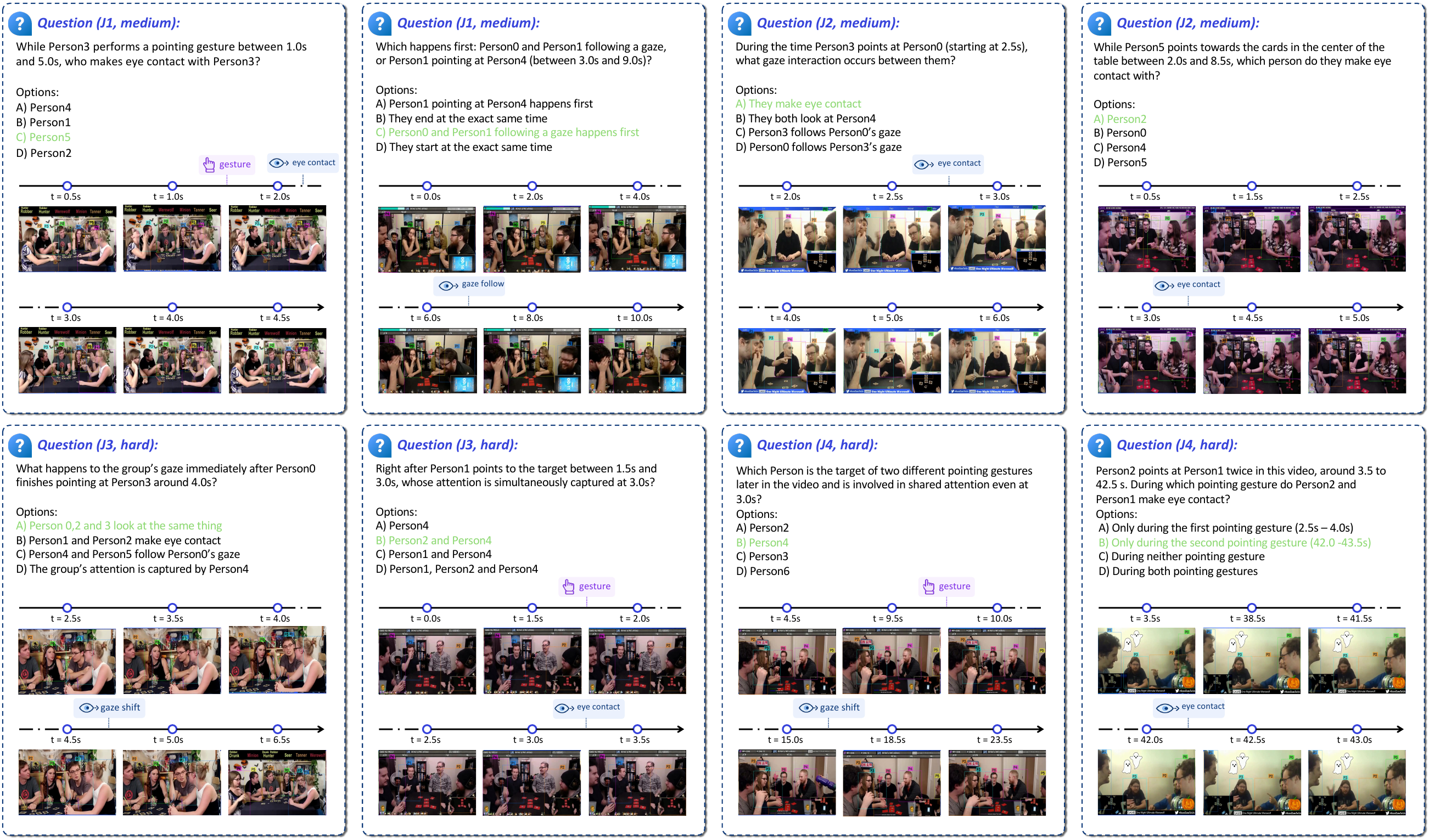}
\caption{
Qualitative \textit{GRASP-Bench} examples for joint gaze--gesture reasoning, covering J1--J4.
}
\label{fig:grasp_joint}
\end{figure*}

We present representative \textit{GRASP-Bench} examples in Fig.~\ref{fig:grasp_gaze},~\ref{fig:grasp_gesture}, and~\ref{fig:grasp_joint}. These examples cover gaze reasoning (T1--T6), gesture reasoning (G1--G6), and joint gaze--gesture reasoning (J1--J4), respectively. Each example shows the video context with tracked participant IDs, the relevant non-verbal cue, and the corresponding question--answer pair. Together, they illustrate the range of \textit{GRASP-Bench}, from direct perception questions such as gaze target or gesture type recognition to harder questions requiring temporal ordering, reciprocal interaction, and gaze--gesture alignment.

\subsection{Reasoning Trace Comparison with Baselines}
\label{appendix:qualitative_baselines}
\begin{figure*}[t]
\centering
\includegraphics[width=1.0\textwidth]{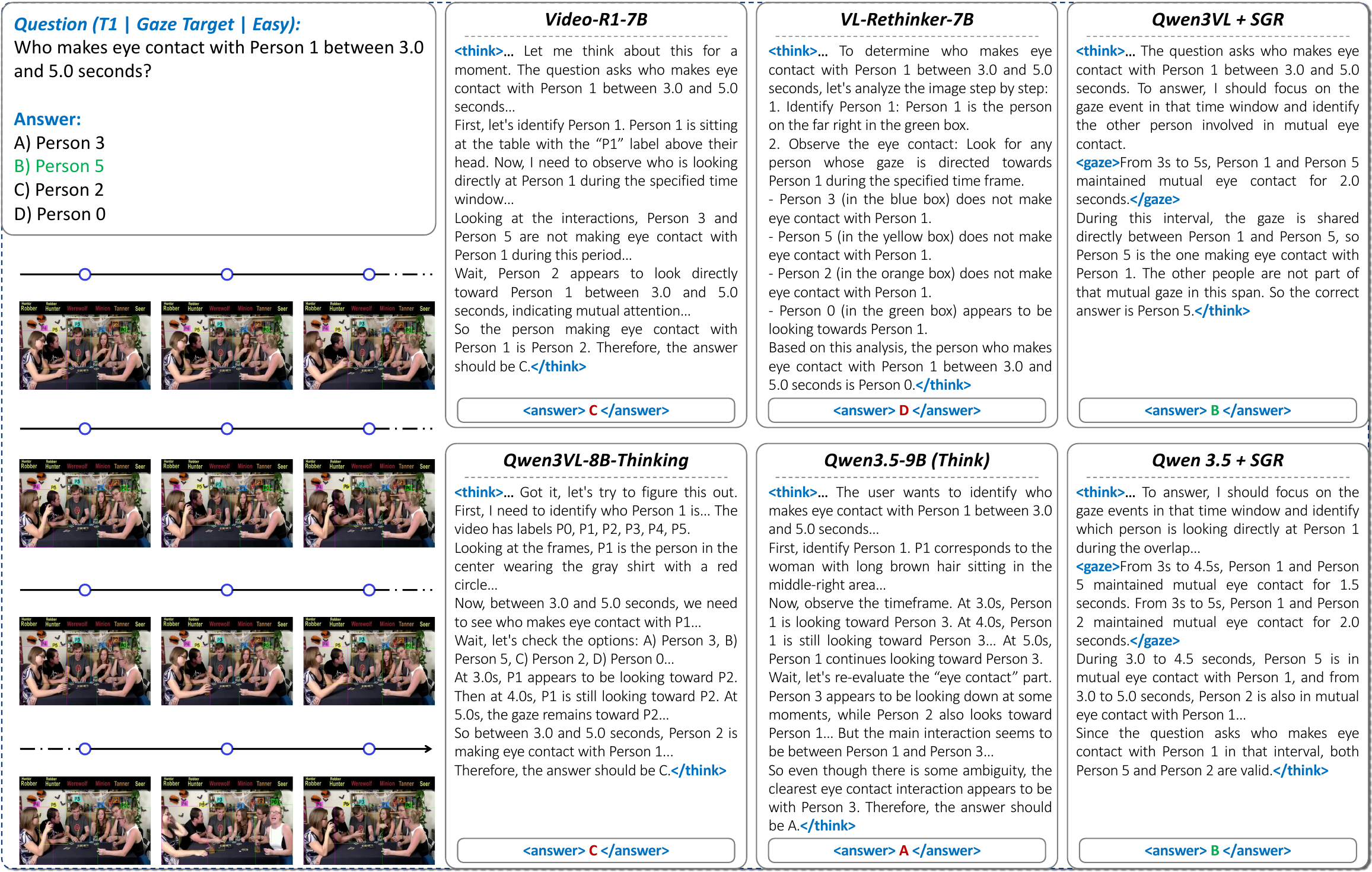}
\caption{
Qualitative comparison on an easy \textit{GRASP-Bench} example.
}
\label{fig:qual_grasp_easy}
\end{figure*}

\begin{figure*}[t]
\centering
\includegraphics[width=1.0\textwidth]{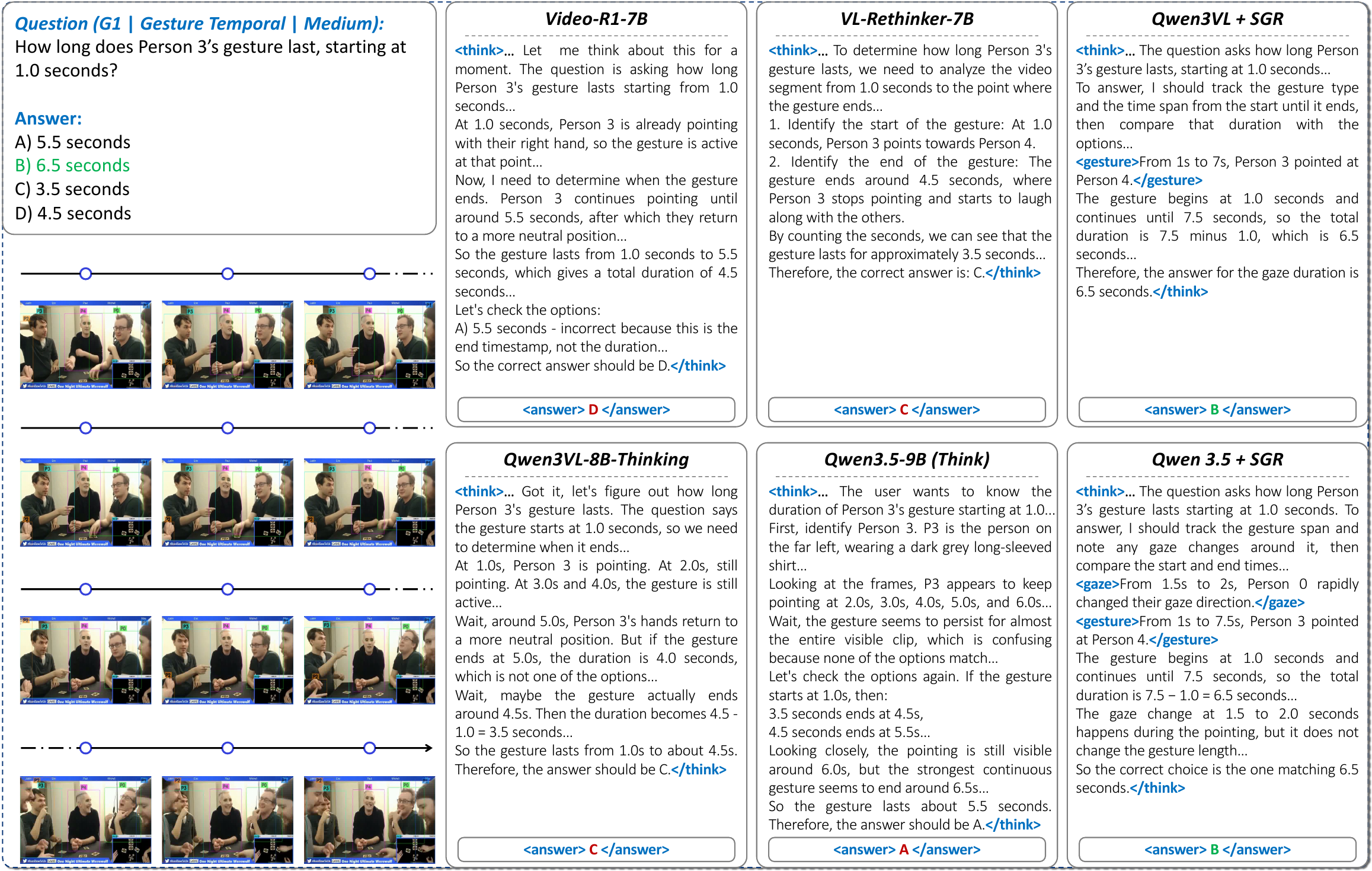}
\caption{
Qualitative comparison on a medium \textit{GRASP-Bench} example.
}
\label{fig:qual_grasp_med}
\end{figure*}

\begin{figure*}[t]
\centering
\includegraphics[width=1.0\textwidth]{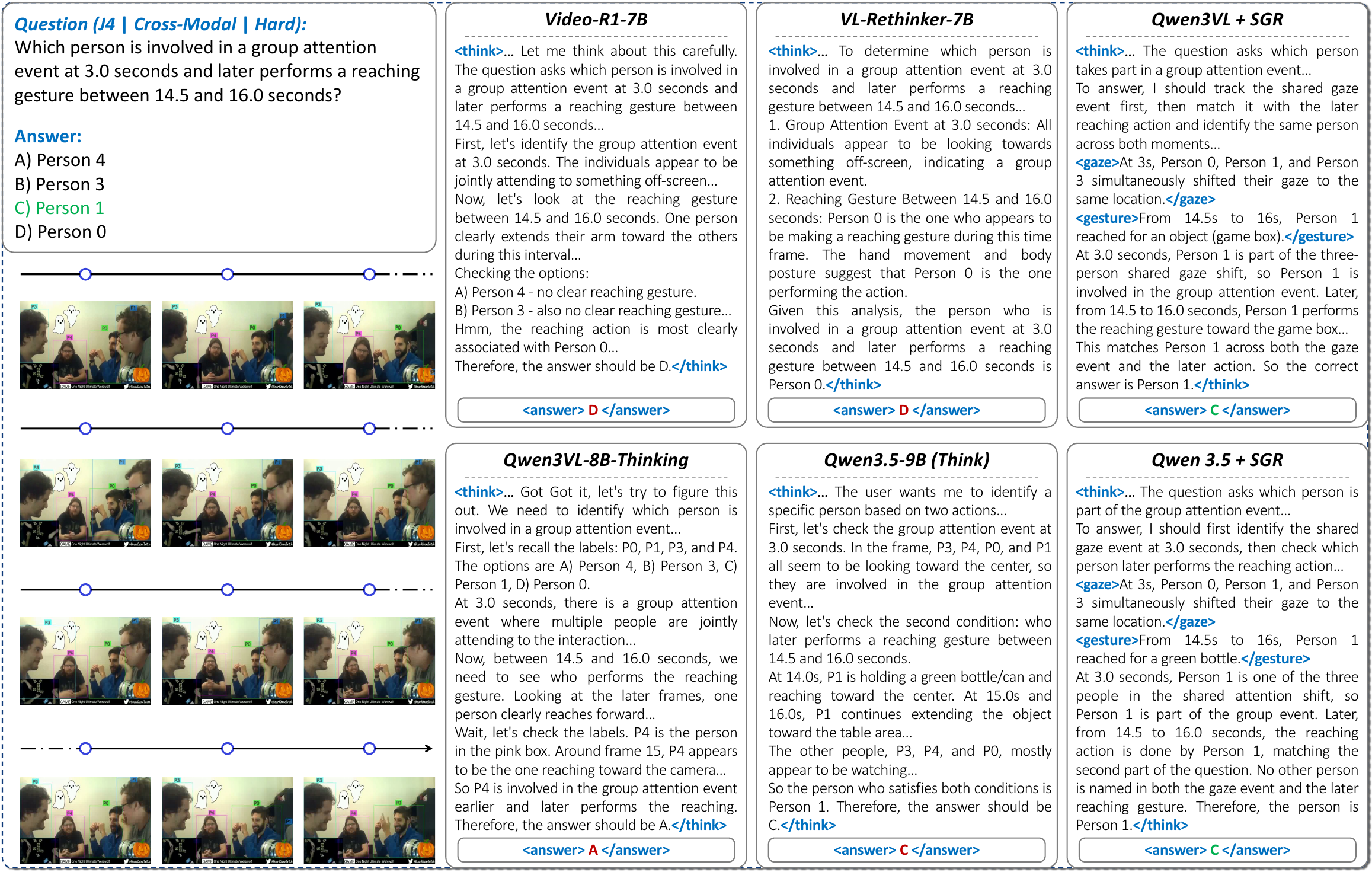}
\caption{
Qualitative comparison on a hard \textit{GRASP-Bench} example.
}
\label{fig:qual_grasp_hard}
\end{figure*}

\begin{figure*}[t]
\centering
\includegraphics[width=1.0\textwidth]{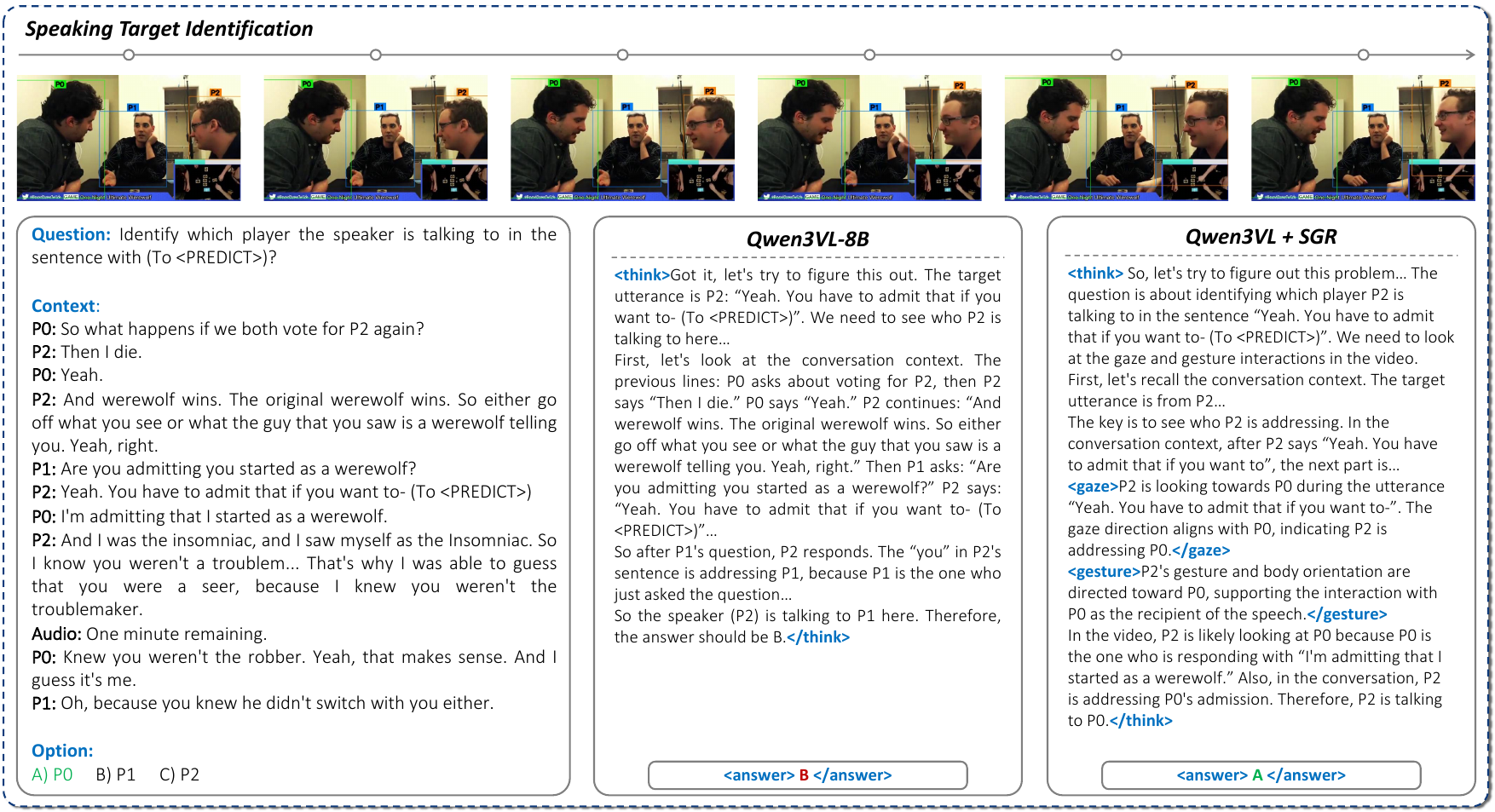}
\caption{
Qualitative comparison on MMSI speaker target identification (STI).
}
\label{fig:qual_sti}
\end{figure*}

\begin{figure*}[t]
\centering
\includegraphics[width=1.0\textwidth]{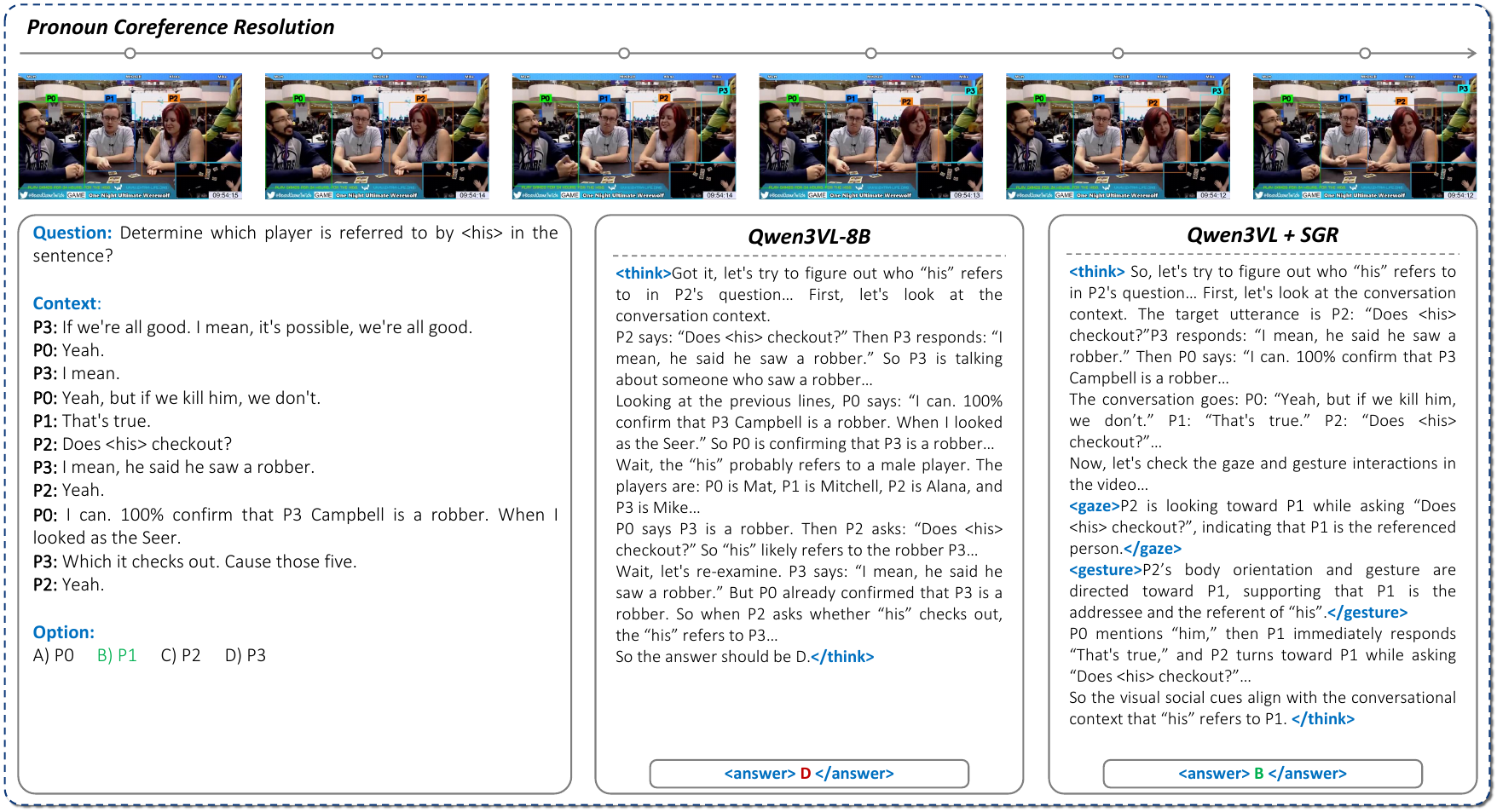}
\caption{
Qualitative comparison on MMSI pronoun coreference resolution (PCR).
}
\label{fig:qual_pcr}
\end{figure*}

\begin{figure*}[t]
\centering
\includegraphics[width=1.0\textwidth]{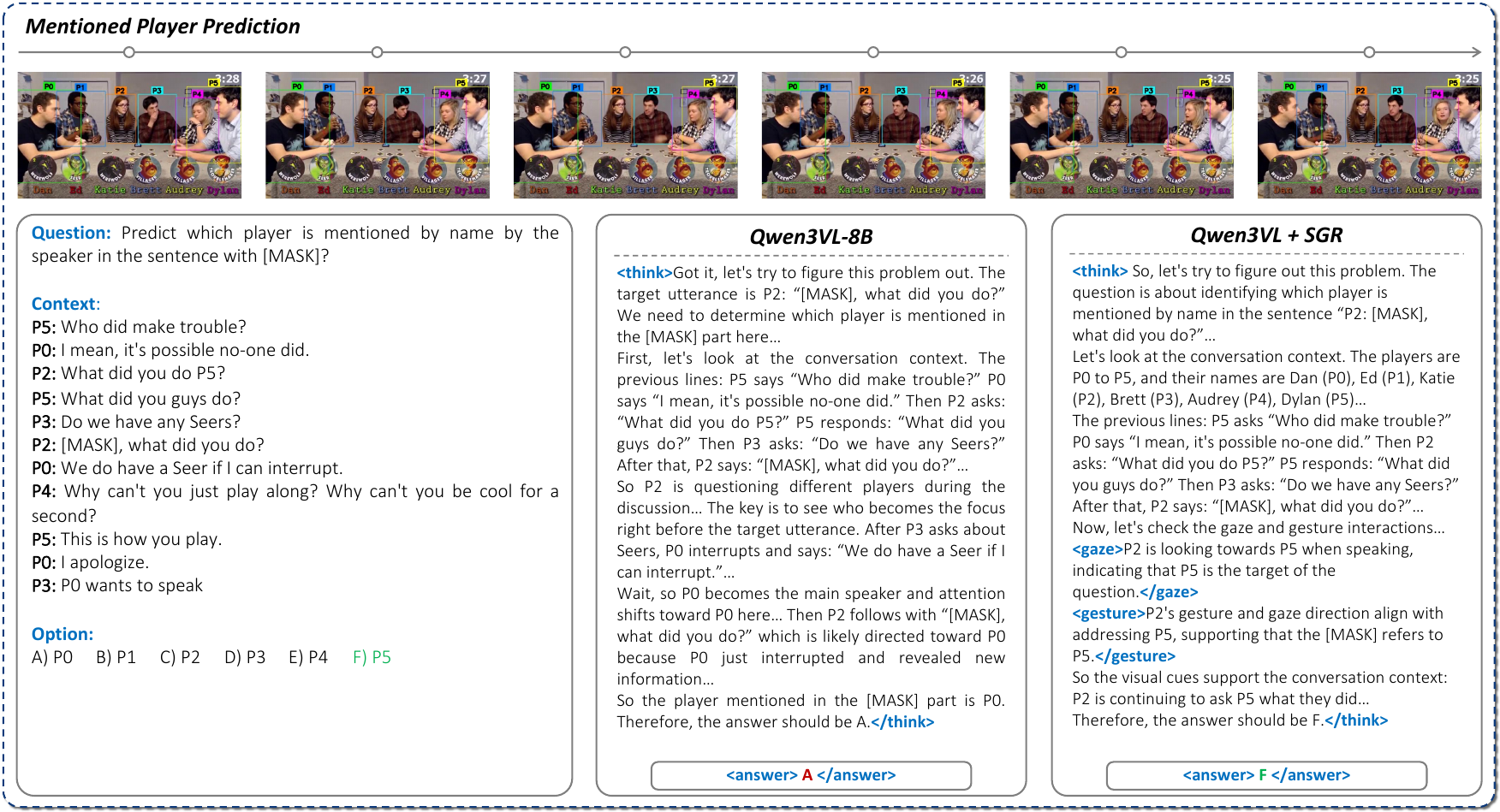}
\caption{
Qualitative comparison on MMSI mentioned-player prediction (MPP).
}
\label{fig:qual_mpp}
\end{figure*}

\begin{figure*}[t]
\centering
\includegraphics[width=1.0\textwidth]{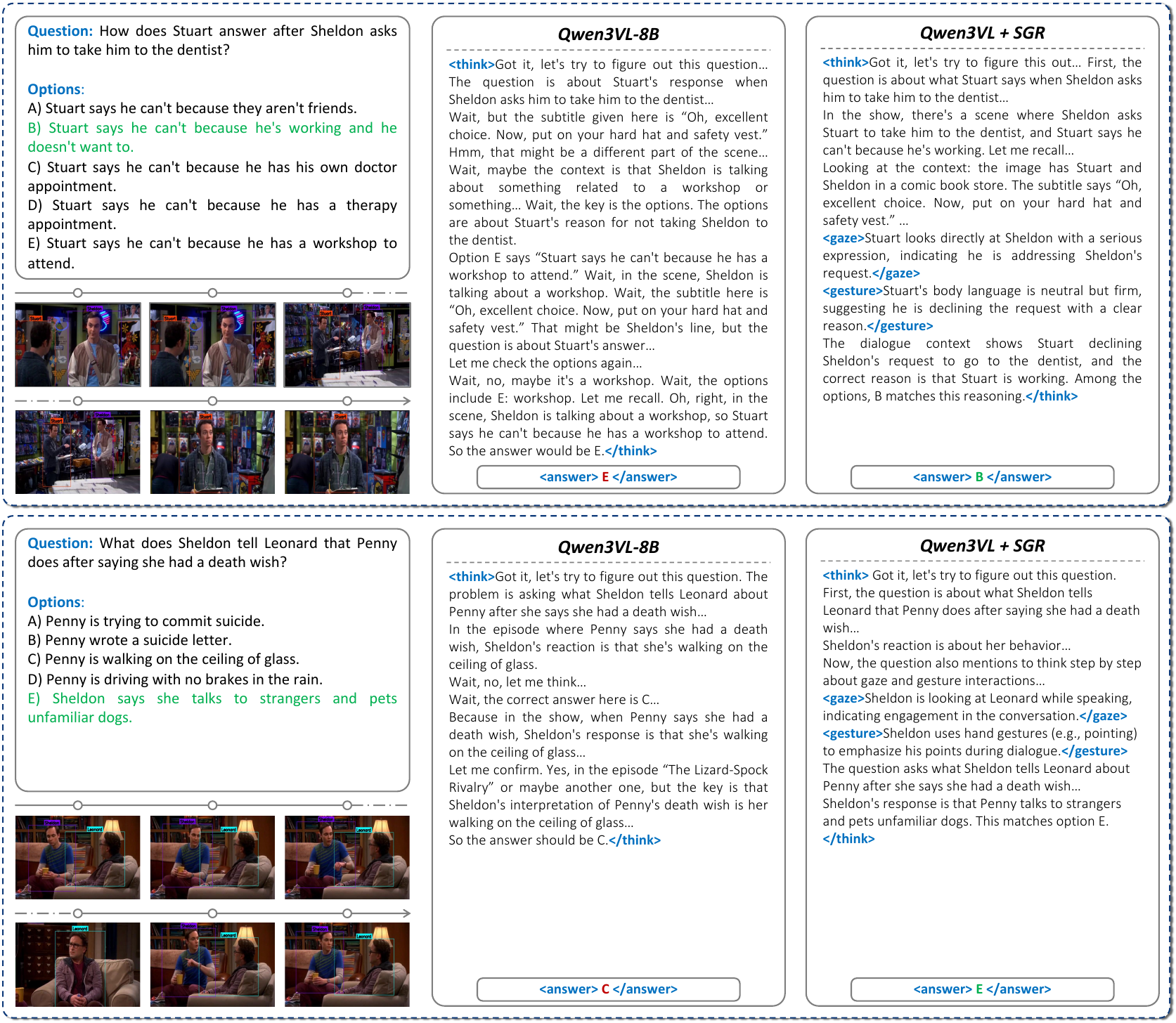}
\caption{
Qualitative comparison on TVQA+.
}
\label{fig:qual_tvqa}
\end{figure*}

We compare reasoning traces from our SGR-trained models against representative supervised, reasoning, and RL post-training baselines. Fig.~\ref{fig:qual_grasp_easy},~\ref{fig:qual_grasp_med}, and~\ref{fig:qual_grasp_hard} show \textit{GRASP-Bench} examples across easy, medium, and hard difficulty levels. In addition to \textit{GRASP-Bench}, we include examples from MMSI~\cite{lee2024modeling} and TVQA+~\cite{lei2020tvqa+}. Fig.~\ref{fig:qual_sti},~\ref{fig:qual_pcr}, and~\ref{fig:qual_mpp} show the three MMSI tasks: speaker target identification (STI), pronoun coreference resolution (PCR), and mentioned-player prediction (MPP). Fig.~\ref{fig:qual_tvqa} shows a TVQA+ example. Across these cases, baselines often rely on salient participants or dialogue priors, while our models more consistently ground their responses in the relevant participants and visual/social cues.

\subsection{Failure Cases}
\label{appendix:qualitative_failures}
\begin{figure*}[t]
\centering
\includegraphics[width=1.0\textwidth]{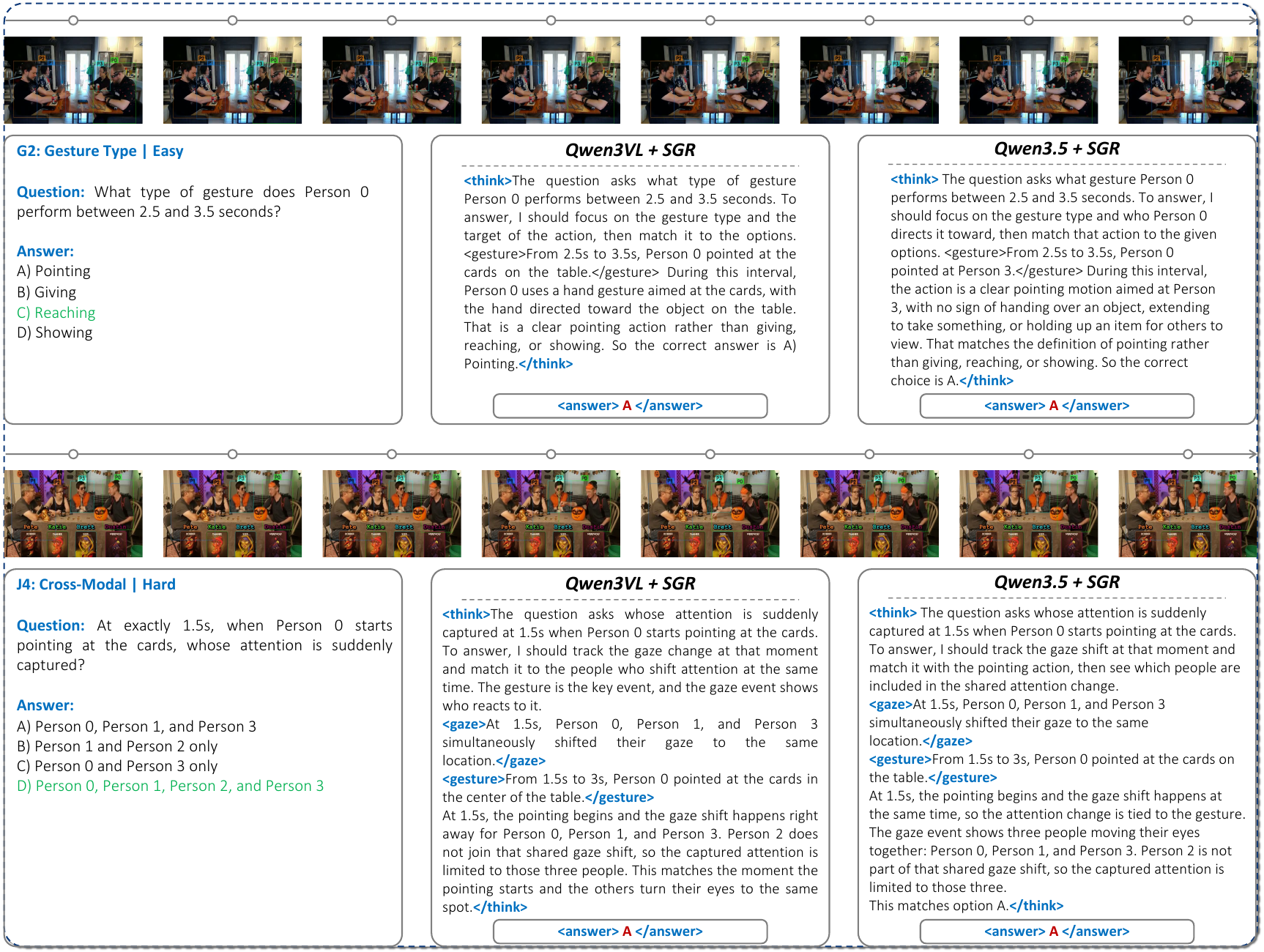}
\caption{
Failure cases on \textit{GRASP-Bench}. We show two representative errors: an ambiguous deictic gesture where reaching and pointing cues are visually close, and a crowded gaze-reasoning case where multiple gaze events occur within the target interval.
}
\label{fig:qual_fail}
\end{figure*}

Fig.~\ref{fig:qual_fail} shows representative failure cases of our method. In the first example, the gesture is visually ambiguous: the queried interval contains a point-like phase of a reaching motion, while later frames make the reaching/recoil pattern clearer. Such cases are challenging because the distinction between pointing and reaching can be subtle even under human inspection. In the second example, the target interval contains multiple gaze events from the same participant, with several people involved in nearby social activity. The model misses one relevant gaze cue and therefore grounds its answer in incomplete evidence. These examples suggest that remaining errors often arise from ambiguous gesture semantics and limited resolution of dense multi-person gaze dynamics.

\section{Broader Impacts}
\label{appendix:broader_impacts}

\textit{GRASP} and \textit{SGR} are intended to advance research on multimodal social understanding. By encouraging models to reason about who interacts with whom from gaze and deictic gestures, our work may benefit human-centered video understanding, assistive agents, collaborative robotics, and socially aware interfaces where participant-level grounding is important.

At the same time, social reasoning from video requires careful interpretation. \textit{GRASP} is constructed from publicly available or research-oriented video sources, and people are represented through anonymous participant IDs rather than real identities. The benchmark focuses on event-level visual reasoning, such as who looks at whom or who gestures toward whom, and is not designed for identifying individuals, profiling people, or inferring private mental states. Since gaze and gesture can be ambiguous, culturally variable, and sensitive to occlusion or camera viewpoint, models trained or evaluated on \textit{GRASP} should not be used for high-stakes decisions about people.

Our dataset and models may also inherit biases from the source videos and from the foundation models used in the construction pipeline. Social settings, camera styles, gestures, and interaction norms in the data may not represent all populations or contexts. In addition, automatic gaze and gesture extraction can introduce annotation noise in complex scenes. We mitigate these issues through confidence filtering, structured validation, and subset human feedback, but improving diversity and event reliability remains important future work.

Finally, \textit{GRASP} uses multimodal and language models for gesture annotation, event description, and QA generation. To reduce unsupported inferences, ground-truth answers are derived from structured fields such as participant IDs, event types, timestamps, and gaze--gesture links, rather than free-form generated text. We believe that releasing \textit{GRASP} as a documented research benchmark can help the community study socially grounded MLLMs more transparently, while making the limitations and intended scope of the dataset explicit.

\clearpage
\newpage

\begin{figure*}[t]
\centering
\begin{tcolorbox}[
    colback=gray!5,
    colframe=black!70,
    title={Gesture prompt},
    width=\textwidth,
    boxrule=0.5pt,
    arc=1mm,
    left=1mm,
    right=1mm,
    top=0.5mm,
    bottom=0.5mm
]
\tiny
\begin{Verbatim}[breaklines=true, breakanywhere=true]
Analyze this entire video and detect ALL deictic gestures.

## About This Video
- The video shows a social interaction between several people.
- P0, P1, P2... represents person IDs.
- Scan the ENTIRE video from start to end and annotate ALL deictic gestures that are defined below with the correct timestamps.
- Deictic gestures are related to HAND, FINGER, and ARM movements, and are NOT related to other motions like head pose.

## Timestamp - Visual Reference (CRITICAL)
- Look at the text overlay at the top-left of the video, which represents the current time in seconds: "t=X.XXs".
- ALWAYS use the 't' value (seconds) shown on the screen for your timestamps.

## Deictic Gesture Types

1. **pointing** - Hand gesture intentionally indicating a specific target person
  - YES: Clear arm extension with finger/hand pointing at a specific person
  - NOT the case: Talking with hands (beat/emphasis gestures), casual or rhythmic hand movements, arm resting, or reaching/manipulating gestures where the primary intent is interaction rather than indicating
  - Start: When the arm begins to lift and extend.
  - End: When the hand begins to retract after pointing.

2. **showing** - Presenting an object for others to visually inspect
  - YES: The subject intentionally presents an object by orienting, tilting, sliding, or positioning it so that its visual properties are clearly exposed to others
  - NOT the case: Simply holding or carrying an object without presenting it, object resting on table/lap without intentional presentation, or manipulating an object for use rather than display
  - Start: When the object begins to be intentionally oriented or positioned for visual inspection
  - End: When the object returns to a neutral/use position or is no longer presented

3. **giving** - Intentionally offering an object to transfer possession
  - YES: The subject intentionally offers an object to another person in a way that clearly invites transfer of possession, regardless of whether the recipient actually takes it
  - NOT the case: Simply holding or displaying an object, presenting an object for viewing (showing), or moving an object without intent to hand it over
  - Start: When the object begins to be purposefully moved or offered toward another person for transfer
  - End: When the offering action clearly ends (e.g., the object is released, pulled back, or the offer is withdrawn)

4. **reaching** - Extending the hand toward an object / person with intent to touch it
  - YES: The subject clearly expresses intent to acquire the object or touch the person through a directed hand/arm extension toward an object not in their possession, often accompanied by pre-grasp configuration (open/curved fingers) or body movement (e.g., leaning forward)
  - NOT the case: Pointing or indicating gestures, casual arm extension without acquisition intent, expressive or beat gestures, or any motion where the hand is not plausibly attempting to obtain the object
  - Start: When the hand/arm begins a purposeful reach toward an object with acquisition intent
  - End: When the reach terminates

## Rules
1. Be **strict and conservative**: Annotate a gesture ONLY when the gesture type and intent are **clearly and unambiguously identifiable** from hand/arm motion alone.
2. If a gesture's **intent or type cannot be clearly determined** from the hand/arm motion alone, do NOT annotate it.
3. Each gesture must have a clear start_time and end_time (in seconds).
4. Person IDs must exactly match the visible Person IDs (P0, P1, P2, ...).
5. If no **high-confidence** deictic gestures are visible, return an EMPTY array.

## IMPORTANT NOTE
- Focus ONLY on hand, finger, and arm movements that clearly express communicative intent.
- Do NOT infer intent from gaze, head orientation, speech, or context outside the visible gesture.
- Only annotate gestures that can be CLEARLY identified as one of the above four types: pointing, showing, giving, reaching.
- The DURATION (end_time - start_time) of the gesture should be at LEAST 1 SECOND, if not, do NOT annotate it.
- There will be multiple gesture events in the video even simultaneously. You should annotate all simultaneous gestures.
- If a previous prediction or critique report is provided, use it to reduce false positives rather than increase coverage.

## Response Format (JSON only)

{
  "gestures": [
    {
      "gesture_type": "pointing" | "showing" | "giving" | "reaching",
        "initiator_id": <number>,
        "start_time": <t value from screen>,
        "end_time": <t value from screen>,
        "target_type": "person" | "object",
        "target_person_id": <ID or null>,
        "target_description": "<brief description>",
        "confidence": <0.0 to 1.0>
        "reasoning": <why you categorized this gesture as <gesture_type>>
    }
  ]
}
\end{Verbatim}
\end{tcolorbox}
\caption{Prompt for deictic gesture annotation.}
\label{fig:gesture_prompt}
\end{figure*}

\begin{figure*}[t]
\centering
\begin{tcolorbox}[
    colback=gray!5,
    colframe=black!70,
    title={QA prompt},
    width=\textwidth,
    boxrule=0.5pt,
    arc=1mm,
    left=1mm,
    right=1mm,
    top=0.5mm,
    bottom=0.5mm
]
\tiny
\begin{Verbatim}[breaklines=true, breakanywhere=true]
You are a QA generator for a non-verbal social reasoning dataset.

You will receive structured event annotations detected in a video clip. Your job: generate question-answer pairs that the model must answer by WATCHING THE VIDEO.

=== STRICT OUTPUT RULES ===

1. ANSWER FORMAT — two types only:
   - MCQ: answer field = single letter (A/B/C/D). answer_text = full text of correct option.
   - Open-ended: answer field = answer_text = 2-4 sentence descriptive answer. Must reference specific Person IDs and timestamps from the event data.

2. ANSWER CONTENT:
   - Factual, no reasoning. NEVER include these words in answers:
     "suggesting", "indicating", "likely", "because", "implies", "seems", "probably", "emotion", "feeling"

3. QUESTION QUALITY:
   - Must require WATCHING THE VIDEO to answer. Not answerable from question text alone.
   - Use natural language: say "eye contact" not "mutual gaze event", "points at" not "performs a pointing gesture", "looking at the same thing" not "joint attention with high convergence".
   - Include specific timestamps from the event data.
   - All person references: "Person {id}" format.

4. MCQ RULES:
   - Exactly 4 options (A/B/C/D).
   - Distractors must be plausible: other person IDs in the video, other event types that didn't happen, swapped roles, nearby but incorrect timestamps.
   - Option text must be concise (<=15 words per option).

5. OUTPUT FORMAT:
   Return ONLY a JSON array. No explanation, no markdown, no commentary.
   Each element:
   {
     "category": "T1",
     "difficulty": "easy",
     "format": "mcq" | "open_ended",
     "question": "...",
     "options": ["A) ...", "B) ...", "C) ...", "D) ..."],
     "answer": "C",
     "answer_text": "Person 3",
     "source_event_ids": [4000],
     "time_range": [13.0, 15.5]
   }
   - "options" field: ONLY for mcq.
   - "source_event_ids": gaze event_id(s) or gesture index(es) used as GT.
   - "time_range": [start, end] of relevant events.
\end{Verbatim}
\end{tcolorbox}
\caption{Prompt QA generation.}
\label{fig:qa_prompt}
\end{figure*}


\end{document}